\documentclass{article}

\usepackage[colorlinks,bookmarksopen,bookmarksnumbered,citecolor=red,urlcolor=red]{hyperref}
\usepackage{amsthm}
\usepackage{mathtools}      % Loads amsmath
\usepackage{amssymb}        % provides AMS fonts and symbols ex. \mathbb{...}
\usepackage{graphicx}       % enhanced support for graphics
\usepackage{marginnote}     % Add notes
\usepackage{marvosym}       % mvat
\usepackage{overpic}        % Add text on figures
\usepackage{tabularx}
\usepackage{color}
\usepackage[linesnumbered,algoruled,boxed,lined]{algorithm2e}
\usepackage[normalem]{ulem}
\usepackage{enumitem}
\usepackage[normalem]{ulem}

\usepackage{xcolor,cancel}

\usepackage{multirow}
\usepackage{comment}
\usepackage[export]{adjustbox}% http://ctan.org/pkg/adjustbox
\usepackage{epstopdf}

\usepackage[font=small]{caption}% http://ctan.org/pkg/caption
\usepackage[letterpaper,left=1in,right=1in,top=1in,bottom=1in,
			footskip=.25in,heightrounded]{geometry}

\newif\iftwocolumn
\twocolumntrue

\newtheorem{problem}{Problem}
\newtheorem{proposition}{Proposition}%[section]
\newtheorem{lemma}{Lemma}%[section]
\newtheorem{theorem}{Theorem}%[section]
\theoremstyle{definition}
\theoremstyle{remark}
\newtheorem*{remark}{Remark}

\SetKwProg{Fn}{Function}{}{}
\SetKwComment{Comment}{$\triangleright$\ }{}

\newcommand{\D}{\mathcal{D}}

\def\ldg{G^{\,\ell}\xspace}
\def\ldgg{G^{\,\ell}\xspace}
\def\udg{G^{\,u}\xspace}
\def\udgg{G^{\,u}\xspace}
\def\toro{\texttt{TORO}\xspace}

\def\lrbm{\texttt{LRBM}\xspace} 
\def\urbm{\texttt{URBM}\xspace} 
\def\mrb{\textsc{MRB}\xspace}
\def\rb{\textsc{RB}\xspace}
\def\fvs{\textsc{MFVS}\xspace}

\def\DFSDP{\textsc{DFDP}\xspace}
\def\PQS{\textsc{PQS}\xspace}
\def\spp{\textsc{SepPlan}\xspace}
\def\vsp{\texttt{VSP}\xspace}
\def\minvs{\textsc{MinVS}\xspace}
\def\vertsep{\textsc{VS}\xspace}
\def\ilpmrb{\textsc{TB}_{\mrb}\xspace}
\def\ilptb{\textsc{TB}_{\mrb}\xspace}
\def\ilpfvs{\textsc{TB}_{\textsc{FVS}}\xspace}

\def\trlb{\textsc{TRLB}\xspace}
\def\toro{\texttt{TORO}\xspace}
\def\toroe{\texttt{TORE}\xspace}
\def\toroi{\texttt{TORI}\xspace}

\def\b#1{\textcolor{black}{#1}}

\begin{document}

\date{}
\title{Minimizing Running Buffers for Tabletop Object Rearrangement: Complexity, Fast Algorithms, and Applications}
\author{Kai Gao\quad
Si Wei Feng\quad
Baichuan Huang\quad
and Jingjin Yu}%\affilnum{1}}
\maketitle

\begin{abstract}

For rearranging objects on tabletops with overhand grasps, temporarily relocating objects to some buffer space may be necessary. 
This raises the natural question of how many simultaneous storage spaces, or ``running buffers'', are required so that certain classes of tabletop rearrangement problems are feasible.
In this work, we examine the problem for both labeled and unlabeled settings. 
On the structural side, we observe that finding the minimum number of running buffers (\mrb) can be carried out on a dependency graph abstracted from a problem instance, and show that {computing \mrb is NP-hard}. 
We then prove that under both labeled and unlabeled settings, even for 
uniform cylindrical objects, the number of required running buffers may 
grow unbounded as the number of objects to be rearranged increases. 
We further show that the bound for the unlabeled case is tight. 
On the algorithmic side, we develop effective exact algorithms 
for finding \mrb for both labeled and unlabeled tabletop rearrangement 
problems, scalable to over a hundred objects under very high object 
density. More importantly, our algorithms also compute a sequence 
witnessing the computed \mrb that can be used for solving object 
rearrangement tasks. 
Employing these algorithms, empirical evaluations reveal that random 
labeled and unlabeled instances, which more closely mimics real-world 
setups, generally have fairly small \mrb{s}. 
Using real robot experiments, we demonstrate that the running buffer
abstraction leads to state-of-the-art solutions for in-place rearrangement of 
many objects in tight, bounded workspace. 
\end{abstract}

%\section*{Paper Structure}
%\input{texs/00_paper-structure}

\section{Introduction}\label{sec:intro}
In nearly all aspects of our daily lives, be it work-related, at home, or for play, objects are to be grasped and rearranged, e.g., tidying up a messy desk, cleaning the table after dinner, or solving a jigsaw puzzle. 
Similarly, many industrial and logistics applications require repetitive rearrangements of many objects, e.g., the sorting and packaging of products on conveyors with robots, and doing so efficiently is of critical importance to boost the competitiveness of the stakeholders. 

\begin{figure}[ht]
    \centering
\begin{overpic}
[width=0.6\columnwidth]{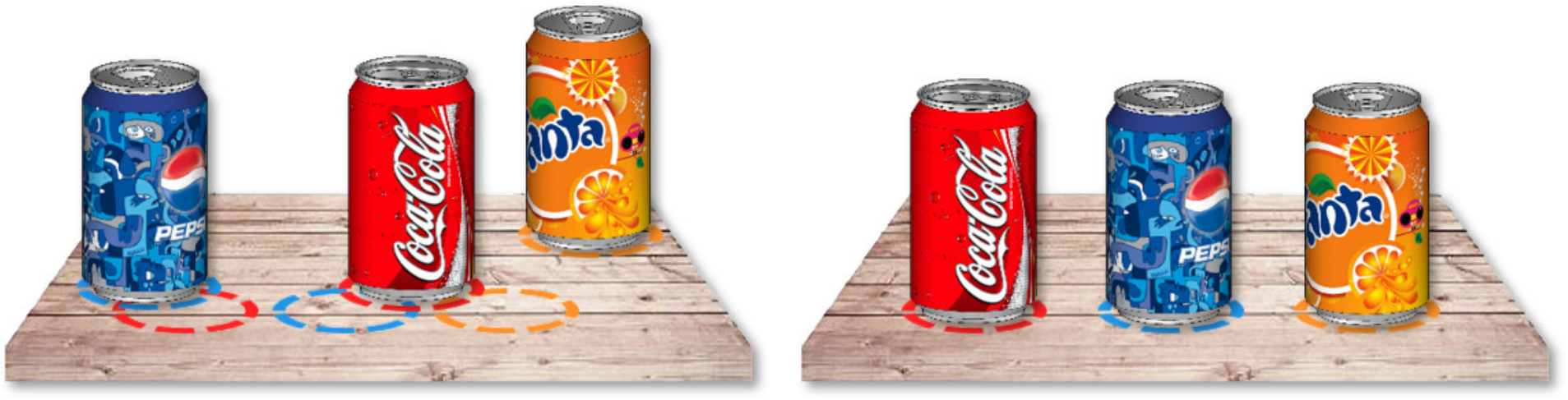}
\end{overpic}
\caption{A \toro instance where the three soda cans are to be rearranged 
from the left configuration to the right configuration.}
\label{fig:toro}
\end{figure}

However, even without the challenge of grasping, deciding the object manipulation order for optimizing a 
rearrangement task is non-trivial. To that end,  \cite{han2018complexity} examined the problem of \emph{tabletop object rearrangement with overhand grasps} (\toro), where objects may be picked up, moved around, and then placed at poses that are not in collision with other objects. 
An object that is picked up but cannot be directly placed at its goal is temporarily stored at a \emph{buffer}.
For example, for the setup given in Fig.~\ref{fig:toro}, using a single manipulator, either the Coke or the Pepsi must be moved to a buffer before the task can be completed. They show that computing a pick-n-place sequence that minimizes the use of the \emph{total number of buffers} is NP-hard and provide methods for 
computing that solution for low tens of objects.

In this study, we examine an arguably more practical objective function that minimizes the number of \emph{running buffers}  (\rb) in solving a \toro instance. 
In other words, we seek rearrangement plans that minimize the maximum number of objects stored at buffers at any given moment, assuming that each object is moved to a temporary location at most once. 
We denote this quantity as \mrb  (minimum \b{\# of} running buffers). 
The \b{\mrb} objective is important because if the required \mrb for solving a \toro instance exceeds the available buffer storage, which is limited in practice, then the instance is infeasible. 
\b{It is also shown that the objective is conducive to computing high-quality solutions for solving \toro tasks.}
Therefore, the structural results and the algorithms that we present may be used not only for computing feasible and high-quality rearrangement plans but are also invaluable as a verification tool, e.g., to verify that a certain rearrangement setup will be able to solve most tasks for which it is designed to tackle. 

Surrounding the goal of studying \toro under the \mrb objective, this work brings the following contributions:
\begin{itemize}
    \item We propose and carry out an initial study of solving \toro tasks with a focus on minimizing the number of running buffers (\mrb). The \mrb objective is a natural alternative to minimizing the total number of buffers \cite{han2018complexity}, whereas both objectives may serve as proxies for the computation of high-quality tabletop rearrangement plans. 
    \item On the structural side, in terms of computational complexity, we show that computing \mrb is NP-hard on arbitrary \emph{dependency graphs}, which encodes the combinatorial information of \toro instances. We then further show that the NP-hardness of \mrb computation extends to actual \toro instances. 
    \item Continuing on the structural analysis of the \mrb objective, we establish that for an $n$-object \toro instance where all objects are uniform cylinders, its \mrb can be lower bounded by $\Omega(\sqrt{n})$, even when all objects are \emph{unlabeled}. This implies that the  same is true for the \emph{labeled} setting. We then provide a matching algorithmic upper bound of $O(\sqrt{n})$ for the unlabeled setting. 
    \item On the algorithmic side, we have developed multiple effective methods for solving various \toro tasks optimizing \mrb, including a dynamic programming method for the labeled setting, a priority queue-based algorithm for the unlabeled setting, and a novel \emph{depth-first-search dynamic programming} routine that readily scales to instances with over a hundred objects for both settings. Furthermore, we provide methods for computing plans with the minimum number of total buffers subject to the \mrb constraints. These algorithms not only provide the optimal number of buffers but also provide a rearrangement plan that witnesses the optimal solution. 
    \item Through extensive numerical evaluations and experimental validations, we confirm the effectiveness of \mrb as a proper metric/proxy to optimize in computing high-quality solutions for \toro tasks. Specifically, \mrb-based methods significantly outperform the previous state-of-the-art methods in many highly challenging \toro setups.  
\end{itemize}

%Besides introducing running buffers and the unlabeled dependency graph, this work 
%brings forth several novel technical contributions. First, we show that 
%computing \mrb on arbitrary \emph{dependency graphs}, which encode the combinatorial 
%information of \toro instances, is NP-hard. Second, we establish that for an $n$-object 
%\toro instance, \mrb can be lower bounded by $\Omega(\sqrt{n})$ for uniform cylinders, 
%even when all objects are \emph{unlabeled}. This implies that the  same is true for 
%the \emph{labeled} setting. Then, we provide a matching algorithmic upper bound 
%$O(\sqrt{n})$ for the unlabeled setting. 
%\sw{for all instances}
%\jy{I think this is not necessary to state, since that is what the upper bound means}
%Last but not least, we develop multiple highly effective and optimal algorithms for 
%computing rearrangement plans with \mrb for \toro. 
%
%In particular, we present a dynamic programming method for the labeled setting, 
%a priority queue-based algorithm for the unlabeled setting, and a much more 
%efficient \emph{depth-first-search dynamic programming} routine that readily 
%scales to instances with over a hundred objects for both settings. Furthermore, 
%we provide methods for computing plans with the minimum number of \emph{total 
%buffers} subject to the \mrb constraints. 
%

\begin{figure}[ht!]
    \centering
    \includegraphics[width=0.6\textwidth]{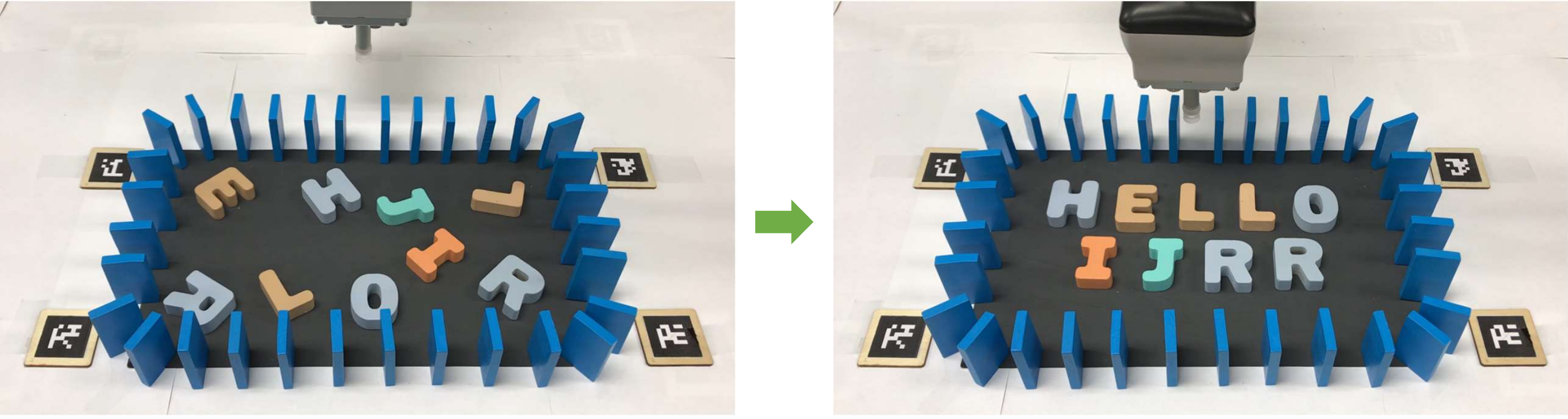}
    \caption{An in-place tabletop rearrangement instance with letter-shaped objects in a bounded workspace. The blue dominoes work as boundaries and do not fall during the execution.}
    \label{fig:helloIJRR}
\end{figure}

This paper builds on the conference version \cite{GaoFenYuRSS21}, with major additions; two of the most important ones are: (1) Many additional theoretical and algorithmic results are added; for both existing and new theorems, complete and expanded proofs are included, and (2) An application of \mrb to carry out in-place tabletop rearrangement in a bounded workspace, with real robot experiment, is added. Part of this addition is based on \cite{gao2021fast} and with new and improved experiments (e.g., Fig.~\ref{fig:helloIJRR}). 
%\jy{Add one figure showing the start and goal configuration of the "HELLO IJRR" setup.}

\textbf{Paper organization}. The rest of the paper is organized as follows. 
We provide an overview of related literature in Sec.~\ref{sec:related-works}.
In Sec.~\ref{sec:preliminaries}, we describe the \mrb focused rearrangement 
problems and discuss the associated dependency graph structure. 
In Sec.~\ref{sec:struture}, we establish some basic structural properties and show that computing \mrb is computationally intractable. 
We proceed to establish lower and upper bounds on \mrb for selected settings in Sec.~\ref{sec:bounds} and describe our proposed algorithmic solutions in Sec.~\ref{sec:algorithms}.
In Sec.~\ref{sec:application}, we present an application of the \mrb algorithms used in the extended scenario which allows internal running buffer{\color{blue} s}. 
Evaluation of simulations follows in Sec.~\ref{sec:experiments}. 
In Sec.~\ref{sec:hardware}, we set up a hardware platform to execute the 
rearrangement plans computed by our proposed algorithms.
We discuss and conclude with Sec.~\ref{sec:conclusions}.

\section{Related Work}\label{sec:related-works}
As a high utility capability, manipulation of objects in 
a bounded workspace has been extensively studied, with works devoted to 
perception/scene understanding \cite{saxena2008robotic,gualtieri2016high,
mitash2017self,xiang2017posecnn}, task/rearrangement planning 
\cite{stilman2005navigation,
treleaven2013asymptotically,havur2014geometric,
krontiris2015dealing,king2016rearrangement,
han2018complexity,lee2019efficient}, 
manipulation \cite{taylor1987sensor,goldberg1993orienting,
lynch1999dynamic,dogar2011framework,bohg2013data,dafle2014extrinsic,
boularias2015learning,chavan2015prehensile}, as well as integrated, 
holistic approaches \cite{kaelbling2011hierarchical,levine2016end,mahler2017dex,
zeng2018robotic,wells2019learning}.
As object rearrangement problems often embed within them multi-robot motion planning 
problems, rearrangement inherits the PSPACE-hard complexity \cite{hopcroft1984complexity}. 
These problems remain  NP-hard even without complex geometric constraints 
\cite{wilfong1991motion}. Considering rearrangement plan quality, e.g, minimizing the 
number of pick-n-places or the end-effector travel, is also computationally 
intractable \cite{han2018complexity}. 

For rearrangement tasks using mainly prehensile actions, the algorithmic 
studies of Navigation Among Movable Obstacles \cite{stilman2005navigation,
stilman2007manipulation} result in backtracking search methods that can 
effectively deal with monotone instances which restrict the robot to move each obstacle at most once.
Via carefully calling monotone solvers, difficult non-monotone cases can 
be solved as well \cite{krontiris2015dealing,wang2022lazy}.
\cite{han2018complexity} relates tabletop rearrangement problems 
to the Traveling Salesperson Problem \cite{papadimitriou1977euclidean}  
and the Feedback Vertex Set problem \cite{karp1972reducibility},
both of which are NP-hard. Nevertheless, integer programming 
models could quickly compute high-quality solutions for 
practical sized ($\sim20$ objects) problems. 
Focusing mainly on the unlabeled setting, bounds on the number of 
pick-n-places are provided for disk 
objects in \cite{bereg2006lifting}.
In \cite{lee2019efficient}, a complete algorithm is developed that reasons 
about object retrieval, rearranging other objects as needed, with later 
work \cite{nam2019planning} considering plan optimality and sensor 
occlusion. 
While objectives in most problems focus on the number 
of motions, \cite{halperin2020space} seeks to minimize 
the space needed to carry out a rearrangement task for discs moving inside the workspace.

Non-prehensile rearrangement has also been extensively studied\cite{ben1998practical,huang2019large}, with object
singulation as an early focus \cite{chang2012interactive,
laskey2016robot,eitel2020learning}. In this problem, a robot is tasked to separate a target object from surrounding obstacles with non-prehensile actions, e.g., pushing and poking, in order to provide room for performing grasping actions. An iterative search was employed in 
\cite{huang2019large} for accomplishing a multitude of rearrangement tasks spanning singulating, separation, and sorting of identically shaped cubes.
\cite{song2019multi} combines Monte Carlo Tree Search
with a deep policy network for separating many objects into coherent clusters
within a bounded workspace, supporting non-convex objects. 
More recently, a bi-level planner is proposed \cite{pan2020decision}, 
engaging both (non-prehensile) pushing and (prehensile) overhand grasping 
for sorting a large number of objects. 
Synergies between non-prehensile and prehensile actions have been explored 
for solving clutter removal tasks \cite{zeng2018learning,huang2020dipn} 
and more challenging object retrieval tasks \cite{huang2021visual,vieira2022persistent}
using a minimum number of pushing and grasping actions.

On the structural side, a central object that we study is the 
\emph{dependency graph} structure.  Similar dependency structures were first introduced to multi-robot path planning problems to deal with path conflicts between agents \cite{buckley1988fast,van2009centralized}.  
Subsequently, the structure was employed for reasoning about and solving challenging 
rearrangement problems \cite{krontiris2015dealing,krontiris2016efficiently,wang2020robot,gao2022toward,gao2022utility}. 
The full labeled dependency graph, as induced by a rearrangement 
instance, is first introduced and studied in \cite{han2018complexity}. This 
current work introduces the unlabeled dependency graph. 
We observe that, in the labeled setting, through the dependency graph, the 
running buffer problems naturally connect to \emph{graph layout} problems 
\cite{diaz2002survey,garey1979computers,papadimitriou1976np,garey1974some,gavril2011some, bodlaender1995approximating},
% \sw{\cite{garey1979, papadimitriou1976np, garey1974some, gavril1977, bodlaender1995approximating} is also related.} 
where an optimal linear ordering of graph vertices is sought. Graph layout 
problems find a vast number of important applications including VLSI design, scheduling 
\cite{shin2011minimizing}, and so on. For the unlabeled 
setting, the dependency graph becomes a planar one for uniform objects with a round base. 
Rearrangement can be tackled through partitioning of the dependency 
graph using a \emph{vertex separator} 
\cite{lipton1979separator,gilbert1984separator,alon1990separator,elsner1997graph}. For a survey on 
these topics, see~\cite{diaz2002survey}. 
% \sw{I feel survey on "Graph Partition" is a bit narrow, maybe survey [40] is better here?}

\section{Preliminaries}\label{sec:preliminaries}
For \b{\toro} tasks where objects assume similar geometry, there are two natural practical settings depending on whether the objects  are distinguishable, i.e., whether they are \emph{labeled} or \emph{unlabeled}. 
We describe external buffer formulations under these two distinct settings, and 
discuss the important \emph{dependency graph} structure for both. 

\subsection{Labeled \toro with External Buffers}
\begin{figure}[h]
    \centering
\begin{overpic}
[width=0.6\textwidth]{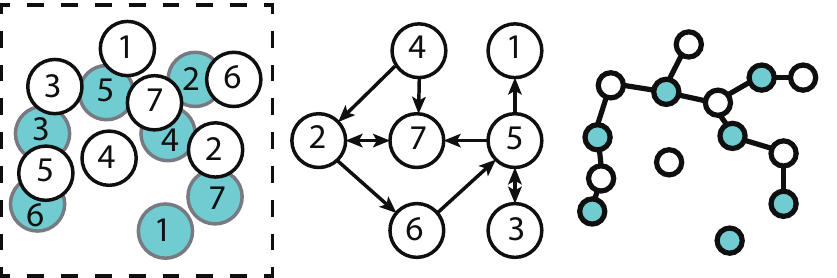}
\put(13, -4){{\small (a)}}
\put(48, -4){{\small (b)}}
\put(80, -4){{\small (c)}}
\end{overpic}
\vspace{2mm}
    %\begin{tabular}{ccc}
    %\includegraphics[width=0.15\textwidth]{../figures/example.png}
    %\includegraphics[width=0.15\textwidth]{../figures/labeled_DG_example.png}
    %\includegraphics[width=0.15\textwidth]{../figures/unlabeled_DG_example.png}
    %\end{tabular}
    \caption{A 7-object labeled instance with uniform cylinders; we 
    will use this instance as a running example. (a) The 
    unshaded discs (as projections of cylinders) represent the start arrangement
    $\mathcal{A}_1$ and the shaded discs represent the goal arrangement $\mathcal{A}_2$. 
    (b) The corresponding labeled dependency graph. (c) The corresponding 
    unlabeled dependency graph, which is bipartite and planar.}
    \label{fig:ex-prob}
\end{figure}

Consider a bounded workspace $\mathcal W \subset \mathbb R^2$ with a set of 
$n$ objects $\mathcal O = \{o_1, \ldots, o_n\}$ placed inside it. All objects 
are assumed to be \emph{generalized cylinders} with the same height. A 
\emph{feasible arrangement} of these objects is a set of poses $\mathcal A 
=\{x_1,\ldots, x_n\}, x_i \in SE(2)$ in which no two objects collide. 
Let $\mathcal A_1 = \{x_1^s, \ldots, x_n^s\}$ and $\mathcal A_2 = 
\{x_1^g, \ldots, x_n^g\}$ be two feasible arrangements, a tabletop object 
rearrangement problem 
\cite{han2018complexity}
seeks a plan using 
\emph{pick-n-place} operations that move the objects from $\mathcal A_1$ to 
$\mathcal A_2$ (see Fig.~\ref{fig:ex-prob}(a) for an 
example with 7 uniform cylinders). In each pick-n-place operation, an object 
is grasped by a robot arm, lifted above all other objects, transferred to and 
lowered at a new pose $p \in SE(2)$ where the object will not be in collision 
with other objects, and then released. A pick-n-place operation can be 
formally represented as a 3-tuple $a = (i, x', x'')$, denoting that object 
$o_i$ is moved from pose $x'$ to pose $x''$. A full rearrangement plan $P = (a_1, a_2, 
\ldots)$ is then an ordered sequence of pick-n-place operations. 

%\jy{ todo: add labels.}

Depending on $\mathcal A_1$ and $\mathcal A_2$, it may not always be 
possible to directly transfer an object $o_i$ from $x_i^s$ to $x_i^g$
in a single pick-n-place operation, because $x_i^g$ may be occupied 
by other objects. This creates \emph{dependencies} between objects. 
If object $o_i$ at pose $x^g_i$ intersects objects $o_j$ at 
pose $x^s_j$, we say $o_i$ \emph{depends} on $o_j$. This suggests that 
object $o_j$ must be moved first before $o_i$ can be placed at its goal
pose $x^g_i$. 

It is possible to have circular dependencies. As an example, for the instance given in Fig.~\ref{fig:ex-prob}(a), objects $3$ and $5$ have dependencies on each other. 
In such cases, some object(s) must be temporarily moved to an intermediate 
pose to solve the rearrangement problem. Similar to \cite{han2018complexity}, \b{for most of this paper},
we assume that \emph{external buffers} outside of the workspace are used for intermediate poses, which avoids time-consuming geometric computations 
if the intermediate poses are to be placed within $\mathcal W$. 
During the execution of a rearrangement plan, there can be multiple objects 
that are stored at buffer locations. We call the buffers that are currently in use \emph{running buffers} (\rb). %
With the introduction of buffers, there are three types of pick-n-place 
operations: 1) pick an object at its start pose and place it at a buffer,
2) pick an object at its start pose and place it at its goal pose, and 3) 
pick an object from a buffer and place it at its goal pose. Notice that buffer 
poses are not important. 
Naturally, it is desirable to be able to solve a rearrangement problem
with the least number of running buffers, giving rise to the \emph{labeled 
running buffer minimization} problem. 

\begin{problem}[Labeled Running Buffer Minimization (\lrbm)]\label{p:1} Given 
feasible arrangements $\mathcal A_1$ and $\mathcal A_2$, find a rearrangement 
plan $P$ that minimizes the maximum number of running buffers in use at any 
given time as the plan is executed. 
\end{problem}

%For reference, \toro is similar to \lrbm instance but does not 
%restrict the buffers to be running buffers. 
%\kg{does not restrict running buffers to be external buffers?}

%
In an \lrbm instance, the set of all dependencies induced by $\mathcal 
A_1$ and $\mathcal A_2$ can be represented using a directed graph $\ldgg = 
(V, A)$, where each $v_i \in V$ corresponds to object $o_i$ and there is an 
arc $v_i \to v_j$ for $1 \le i, j \le n, i \ne j$ if object $o_i$ 
depends on object $o_j$. We call $\ldgg$ a \emph{labeled dependency graph}. 
The labeled dependency graph for Fig.~\ref{fig:ex-prob}(a) is given in 
Fig.~\ref{fig:ex-prob}(b).
Based on the dependency graph $\ldgg$, we can immediately identify multiple 
circular dependencies in the graph, e.g., between objects $3$ and $5$, or among 
objects $7, 2, 6$ and $5$. The cycles form strongly connected components
of $\ldgg$, which can be effectively computed \cite{tarjan1972depth}.  
%
% It is straightforward to see that the dependency graph abstraction fully 
% captures the information needed to solve a tabletop rearrangement problem 
% with external buffers.
Since moving objects to external buffers does not create additional dependencies, we have
%\jy{Can probably add a theorem here to make this more formal.}
\begin{proposition}
$\ldgg$ fully captures the information needed to solve the tabletop rearrangement problem with external buffers moving objects from $\mathcal A_1$ to $\mathcal A_2$.
\end{proposition}

\begin{figure}[ht!]
    \centering
    \includegraphics[width=0.6\textwidth]{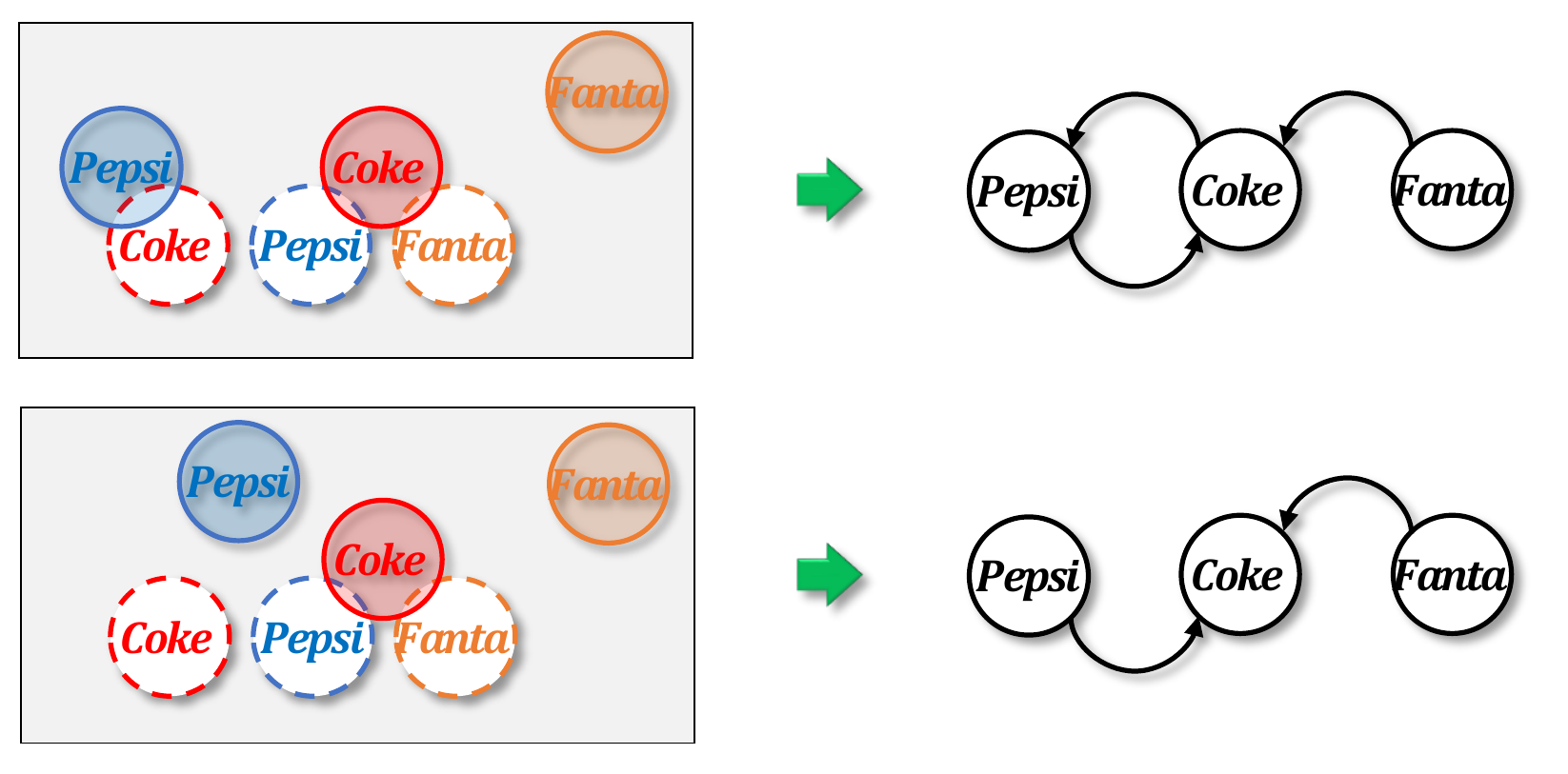}
    \caption{Two configurations of the setup given in Fig.~\ref{fig:toro} and the corresponding dependency graphs.}
    \label{fig:coke-pepsi}
\end{figure}

\b{We use two examples to illustrate the relationships between \toro, its dependency graph, and external buffer-based solutions. First, for the example given in Fig.~\ref{fig:toro}, the corresponding start/goal configurations are given in Fig.~\ref{fig:coke-pepsi}[top left]. The labeled dependency graph is given in  Fig.~\ref{fig:coke-pepsi}[top right]. Because of the existence of cyclic dependencies between Coke and Pepsi, one of these two must be temporarily moved to a buffer. Suppose that Pepsi is moved to buffer. This changes the configuration and dependency graph as shown in the second row of Fig.~\ref{fig:coke-pepsi}. The \toro instance can then be readily solved. The complete solution sequence is $\langle Pepsi \to b, Coke \to g, Pepsi \to g, Fanta \to g \rangle$, where $b$ refers to a buffer and $g$ refers to the goal of the corresponding object.}

\b{For the example given in Fig.~\ref{fig:ex-prob}[a], its labeled dependency graph Fig.~\ref{fig:ex-prob}[b] shows two cycles ($2\leftrightarrow7$ and $3\leftrightarrow 5$). One planning sequence for solving this can be derived that moves $2$ and $3$ to buffer: $\langle 1\to g, 2\to b, 3 \to b, 7 \to g, 4 \to g, 5 \to g, 6 \to g, 2 \to g, 3 \to g\rangle$. }

\subsection{Unlabeled \toro with External Buffers}
In the unlabeled setting, objects are interchangeable. That is, it does not 
matter which object goes to which goal. For example, in Fig.~\ref{fig:ex-prob}, 
object $5$ can move to the goal for object $6$. We call this version the 
\emph{unlabeled running buffer minimization} (\urbm) problem, which is intuitively 
easier. The plan for the unlabeled problem can be represented similarly as the 
labeled setting; we continue to use labels so that plans can be clearly represented
but do not require matching labels for start and goal poses.

For the unlabeled setting, \b{the dependency structure remains but appears in a different form. It is now an \emph{undirected bipartite} graph.} That is, $\udgg = (V_1\cup V_2, E)$ 
where each $v \in V_1$ (resp., $v \in V_2$) corresponds to a start (resp., goal) 
pose $p \in \mathcal A_1$ (resp., $p \in \mathcal A_2$). We denote the vertices representing the start and goal poses as \emph{start vertices} and \emph{goal 
vertices}, respectively. There is an edge between 
$v_1 \in V_1$ and $v_2 \in V_2$ if the objects at the corresponding poses 
overlap. The unlabeled dependency graph for Fig.~\ref{fig:ex-prob}(a) is illustrated in Fig.~\ref{fig:ex-prob}(c). 

\b{In practice, many \toro instances have objects with footprints that are uniform regular polygons (e.g., squares) or discs. In such settings, we can say something additional about the resulting unlabeled dependency graphs (the proofs of the two propositions below can be found in the appendix).}

\begin{proposition}\label{p:udg-polygon}
For unlabeled \toro where footprints of objects are uniform regular polygons, the maximum degree of the unlabeled dependency graph is upper bounded by 19. 
\end{proposition}

\begin{proposition}\label{p:udg}
For unlabeled \toro where footprints of objects are uniform discs, the dependency graph is a planar bipartite graph with a maximum degree of 5. 
\end{proposition}

\b{For either \lrbm or \urbm, the minimum required the number of running buffers is denoted as \mrb, which can be computed based on the corresponding dependency graph.}

\section{Structural Analysis and NP-Hardness}\label{sec:struture}
The introduction of running buffers to tabletop rearrangement problems 
induces unique and interesting structures. 
We highlight some important structural properties of \lrbm, including 
the comparison to minimizing the total number of buffers \cite{han2018complexity}, 
the solutions of \lrbm and \emph{linear arrangement} \cite{shiloach1979minimum} or 
\emph{linear ordering} \cite{adolphson1973optimal} of its dependency graph, and the hardness of computing \mrb for \toro. 

\b{As will be established in this section, computing \mrb solutions for \toro is intimately connected to finding a certain optimal linear ordering of vertices in the associated dependency graph. Because finding an optimal linear ordering is hard, it renders computing \mrb solutions hard as well. This in turns limits the algorithmic solutions that one can secure for computing \mrb solutions for \toro tasks.}

\subsection{Running Buffer versus Total Buffer}
In solving \toro, running buffers are related to but different from the total number of buffers, as studied in \cite{han2018complexity}. 
It is shown the minimum number of total buffers for solving \toro is the same as the size of the minimum \emph{feedback vertex set} (FVS) of the underlying dependency graph. 
An FVS is a set of vertices the removal of which leaves a graph 
acyclic. A \toro with an acyclic dependency graph can be solved 
without using any buffer \b{because there are no cyclic dependencies between any pairs of objects}. We denote the size of the minimum FVS as 
\fvs. 

\b{As a labeled \toro problem, the example from Fig.~\ref{fig:toro} has $\fvs = \mrb = 1$. For the example from Fig.~\ref{fig:ex-prob}(a)(b), $\fvs = 2$ (an FVS set is $\{2, 3\}$) and $\mrb = 2$.}

\begin{figure}[h]
    \centering
    \includegraphics[width=0.45\textwidth]{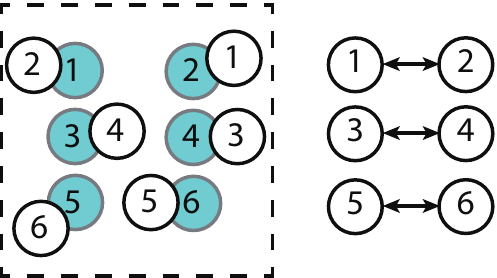}
    \caption{An instance with the labeled dependency graph formed 
by 3 copies of $2$-cycles. The \fvs is 3. On the other hand, 
the \mrb is just $1$ for the problem. \b{The example scales to have arbitrarily large \fvs with \mrb remaining at $1$.}}
    \label{fig:MFVS_vs_MRB}
\end{figure}

As an \b{extreme} example \b{illustrating the difference between \mrb and \fvs}, consider a labeled dependency graph that is formed by $n$ copies of $2$-cycles, e.g., Fig.~\ref{fig:MFVS_vs_MRB}. 
The \fvs of the instance is $n$ \b{because one vertex must be removed from each of the $n$ cycles to make the graph acyclic}. On the other hand, 
the \mrb is just $1$ \b{because each cyclic dependency can be resolved independently from other cycles using a single external buffer}. 
Therefore, whereas the total number of buffers used has more bearing on 
global solution optimality, \mrb sheds more light on \emph{feasibility}. 
Knowing the \mrb tells us whether a certain number of external 
buffers will be sufficient for solving a class of rearrangement problems. 
This is critical for practical applications where the number of 
external buffers is generally limited to be a small constant. 
For example, in solving everyday rearrangement tasks, e.g., sorting 
things or retrieving items in the fridge, a human may attempt to 
temporarily hold multiple items, sometimes awkwardly. There is clearly 
a limit to the number of items that can be held simultaneously this way. 

\begin{figure}[h!]
    \centering
    \begin{overpic}[height=1.5in]{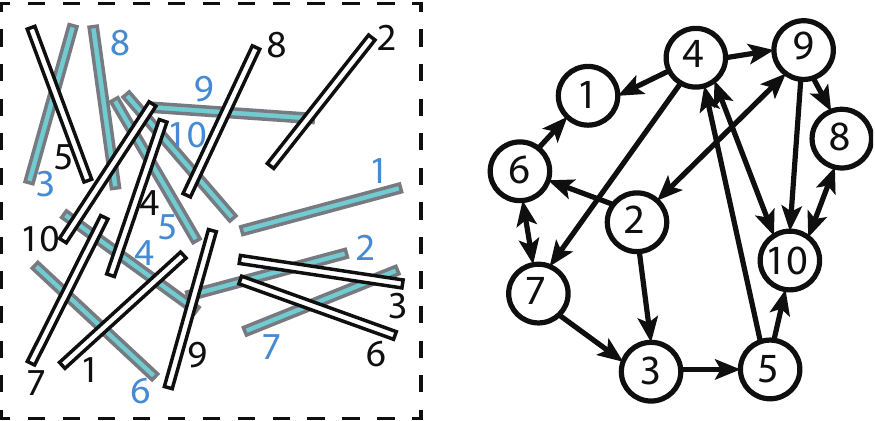}
    \end{overpic}
    \caption{An \lrbm instance with uniform thin cuboids (left) and its 
    labeled dependency graph, where the total number of buffers needed is 
    more than the size of the \fvs when the number of running buffer is 
    minimized.}
    \label{fig:ex-mrb-fvs}
\end{figure}

We give an example where the \mrb and \fvs cannot always be optimized 
simultaneously\b{, i.e., they form a Pareto front}. 
%\kg{The reviewers show confusion here. We'd better clearer demonstrate the purpose of this example. The sentence in the brackets is the original sentence.}
%[We give an example where if the \mrb is 
%optimized first, the total number of buffers used is larger than the \fvs.]
For the setup in Fig.~\ref{fig:ex-mrb-fvs}, the \fvs $\{7, 9, 10\}$ has size $3$\b{ - it can be readily checked that deleting  $\{7, 9, 10\}$ leaves the dependency graph acyclic}. Using our algorithms, the \mrb \b{can be computed to be $2$ with the witness sequence being $1, 10, 8, 4, 5, 3, 6, 7, 2, 9$, corresponding to the plan of $\langle 1 \to g, 10\to b, 8 \to g, 4 \to b, 5\to g, 3\to g, 10 \to g, 6 \to b, 7 \to g, 6 \to g, 2\to b,  9\to g, 4 \to g, 2 \to g \rangle$, where $b$ refers to a buffer and $g$ refers to the goal of the corresponding object. The RB reaches the maximum $2$ when $4$ and $6$ are moved to the buffer.}
\b{However, in this case, the total number of buffers used is $4$: $2, 4, 6$, and $10$ are moved to buffers. This turns out to be the best we can do for this example, that is,  if we are constrained by solutions with $\mrb=2$, the total number of buffers that must be used is at least $4$ instead of the \fvs size of $3$.} 

We note that \b{it is rarely the case that \fvs will be larger when the solution space is constrained to \mrb solutions}; for uniform cylinders, \b{for example}, the total number of buffers needed after first minimizing the running buffer is almost always the same as the \fvs size. 

\subsection{Running Buffers and Linear Vertex Ordering}\label{subsec:logvrb}
Given a graph with vertex set $V$, a \emph{linear ordering} of $V$ is a bijective 
function $\varphi: \{1, \ldots, |V|\} \to V$. Given a labeled dependency graph $ \ldgg(V, A)$ for an \lrbm problem, and a linear ordering $\varphi$ for $V$, we may turn $\varphi$ into a plan $P$ for the \lrbm problem by sequentially picking up objects corresponding to vertices $\varphi(1), \varphi(2), 
\ldots$ For each object that is picked up, it is moved to its goal pose if it has no 
further dependencies; otherwise, it is stored in the buffer. Objects already in the 
buffer will be moved to their goal pose at the earliest possible opportunity. 

\begin{figure}[h!]
    \centering
    \begin{overpic}
    [width=0.6\textwidth]{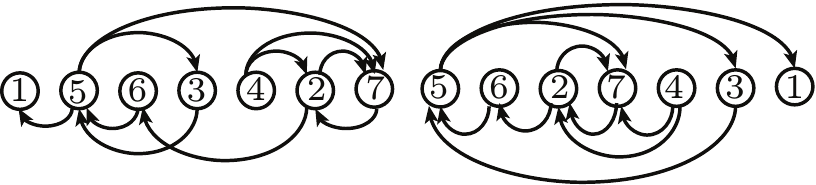}
    \put(22, -3.5){{\small (a)}}
    \put(74, -3.5){{\small (b)}}
    \end{overpic}
    \vspace{2mm}
    \caption{Two linear orderings of vertices of the labeled 
    dependency graph from Fig.~\ref{fig:ex-prob}(b) (i.e., these are different representations of the same graph). The right one minimizes \mrb.}
    \label{fig:lr}
\end{figure}

For example, given the linear ordering $1, 5, 6, 3, 4, 2, 7$ for the dependency graph from Fig.~\ref{fig:ex-prob}(b), first, $1$ can be directly moved to its 
goal. Then, $5$ is moved to the buffer because it has dependencies on $3$ and $7$ 
(but no longer on $1$). Then, $6$ can be directly moved to its goal because 
$5$ is now at a buffer location. Similarly, $3$ can be moved to its goal next. 
Then, $4$ and $2$ must be moved to the buffer, after which $7$ can be moved to its
goal directly. Finally, $2, 4$, and $5$ can be moved to their respective goals 
from the buffer. This leads to a maximum running buffer size of $3$. This 
is not optimal; an optimal sequence is $5, 6, 2, 7, 4, 3, 1$, with $\mrb = 2$. Both 
sequences are illustrated in Fig.~\ref{fig:lr}.

The discussion suggests that we may effectively view the number of running buffers as a function of a dependency graph $ \ldgg$ and a linear ordering $\varphi$. We thus define $\rb( \ldgg, \varphi)$ as the number of running buffers needed for rearranging $ \ldgg$ following the order given by $\varphi$. It is straightforward to see that $\mrb( \ldgg) = \min_{\varphi}\rb( \ldgg, \varphi)$.

\subsection{Intractability of Computing \mrb}
Since computing \fvs is NP-hard \cite{han2018complexity}, one would expect that computing \mrb for a labeled dependency graph, which can be any directed graph, is also hard. We show that this is indeed the case, by examining the interesting relationship between \mrb and the \emph{vertex separation problem} (\vsp), which is equivalent to path width, gate matrix layout, and search number problems as described in Theorem 3.1 in \cite{diaz2002survey}, resulting from a series of studies \cite{kirousis1986searching, kinnersley1992vertex, fellows1989search}. Unless $P=NP$, there cannot be an absolute approximation algorithm for any of these problems \cite{bodlaender1995approximating}. First, we describe the vertex separation problem. 
Intuitively, given an undirected graph $G = (V, E)$, \vsp seeks a linear ordering $\varphi$ 
of $V$ such that, for a vertex with order $i$, the number of vertices that come no later than 
$i$ in the ordering, with edges to vertices that come after $i$, is minimized. 
%

% \begin{problem}[VertSep]
% \end{problem}

\vspace{1mm}
\noindent\fbox{\begin{minipage}{\textwidth}
\vspace{1mm}
\noindent
\textbf{Vertex Separation (\vsp)}\\
\noindent
\textbf{Instance}: Graph $G(V,E)$ and an integer $K$.\\
\noindent
\textbf{Question}: Is there a bijective function $\varphi: \{1, \dots, n\} \to 
V$, such that for any integer $1 \le i \le n$, 
$|\{u\in V \mid \exists (u, v) \in E \ and \ \varphi(u)\leq i < \varphi(v) \}| \leq K$?
\vspace{1mm}
\end{minipage}}
\vspace{1mm}

\begin{figure}[h!]
    \centering
\begin{overpic}
[width=0.6\columnwidth]{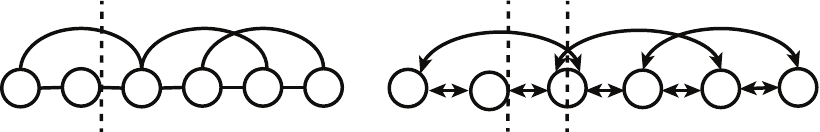}
\put(19, -3){{\small (a)}}
\put(72, -3){{\small (b)}}
\end{overpic}
\vspace{3mm}
\caption{(a) An undirected graph and a linear ordering of its vertices. (b)
A corresponding labeled dependency graph with the same vertex ordering.}
\label{fig:vsp}
\end{figure}

As an example, in Fig.~\ref{fig:vsp}(a), with the given linear ordering, at the second 
vertex, both the first and the second vertices have edges crossing the vertical separator, 
yielding a crossing number of $2$. Given a graph $G$ and a linear ordering $\varphi$, we 
define $\vertsep(G, \varphi) :=
\max_i | \{u\in V \mid \exists (u, v) \in E  \ and \  \varphi(u)\leq i < \varphi(v) 
\} |$, \vsp seeks $\varphi$ that minimizes $\vertsep(G, \varphi)$. 
Let $\minvs(G)$, the vertex separation number of graph $G$, be the minimum $K$ for
which a \vsp instance has a yes answer, then $\minvs(G) = \min_\varphi \vertsep(G, \varphi)$.
Now, given an undirected
graph $G$ and a labeled dependency graph $\ldgg$ obtained from $G$ by replacing each 
edge of $G$ with two directed edges in opposite directions, we observe that there 
are clear similarities between $\vertsep(G, \varphi)$ and $\rb(\ldgg, \varphi)$, which 
is characterized by the following lemma. 

\begin{lemma}\label{l:vs-rb}
$\vertsep(G, \varphi) \leq \rb(\ldgg, \varphi) \leq \vertsep(G, \varphi) + 1$.
\end{lemma}
\begin{proof}
Fixing a linear ordering  $\varphi$, it is clear that $\vertsep(G, \varphi) \le 
\rb(\ldgg, \varphi)$, since the vertices on the left side of a separator with edges 
crossing the separator for $G$ correspond to the objects that must be stored 
at buffer locations. For example, in Fig.~\ref{fig:vsp}(a), past the second vertex
from the left, both the first and the second vertices have edges crossing the 
vertical ``separator''. In the corresponding dependency graph shown in 
Fig.~\ref{fig:vsp}(b), objects corresponding to both vertices must be moved 
to the external buffer. 
%
%Intuitively, this is the case because, for any vertex separator
%
On the other hand, we have $\rb(\ldgg, \varphi) \leq \vertsep(G, \varphi) + 1$ 
because as we move across a vertex in the linear ordering, the corresponding 
object may need to be moved to a buffer location temporarily. 
For example, as the third vertex from the left in Fig.~\ref{fig:vsp}(a) is 
passed, the vertex separator drops from $2$ to $1$, but for dealing with 
the corresponding dependency graph in Fig.~\ref{fig:vsp}(b), the object 
corresponding to the third vertex from the left must be moved to the 
buffer before the first and the second objects stored in the buffer can be 
placed at their goals. 
% \qed
\end{proof}

\begin{theorem}\label{t:mrb-hard}
Computing \mrb, even with an absolute approximation, for a labeled dependency graph is NP-hard. 
\end{theorem}
\begin{proof}
Given an undirected graph $G$, we reduce from approximating \vsp within a 
constant to approximating \mrb within a constant for a dependency 
graph  $\ldgg$ from $G$ constructed as stated before, replacing each edge in $G$ as a 
bidirectional dependency. 

Unless $P=NP$, \vsp does not have absolute approximation in polynomial time. 
Henceforth, if $\mrb(\ldgg, \varphi)$  can be approximated within $\alpha$ in polynomial time, 
which means 
for graph $G$, we can find a $\varphi^*$ in polynomial time such that $\rb(\ldgg, \varphi^*) 
\leq  \mrb(\ldgg) + \alpha$, we then have $\vertsep(G, \varphi^*)\leq \rb(\ldgg, \varphi^*)  
\leq \alpha + \mrb(\ldgg) \leq \minvs(G)+\alpha+1$, which shows vertex separation can have 
an absolute approximation, implying $P=NP$.
% \qed
\end{proof}

\begin{comment}
\begin{itemize}
    \item Hardness result for dependency graphs - relation to vertex separation. 
\end{itemize}
\end{comment}

%\begin{figure}[ht]
%\begin{center}
%\begin{overpic}[width={\iftwocolumn 3.5in \else 5in \fi},tics=5]
%{../figures/problem-eps-converted-to.pdf}
%\put(7, 3){{\small $(x_L, 0)$}}
%\put(83, 33){{\small $(x_R, y_T)$}}
%\put(29, -2){{\small $(x_A, 0)$}}
%\put(45.5,-2){{\small $(0, 0)$}}
%\put(15,28){{\small moving direction}}
%\end{overpic}
%\end{center}
%\caption{\label{fig:conveyor}Illustration of a conveyor workspace where 
%the base of the robot arm is located at $(X_A,0)$. The end-effector %picks 
%up objects within a region $\W$ with a lower left corner of $(x_L, 0)$ 
%and an upper right corner of $(x_R, y_T)$, and drops off objects at 
%the drop-off location $(0, 0)$.}
%\end{figure}

\b{With some additional effort, we can show that computing \mrb solutions for certain \toro instances is NP-hard.}

\b{\begin{theorem}\label{t:mrb-toro-hard}
Computing \mrb for \toro is NP-hard. 
\end{theorem}
\begin{proof}
By \cite{monien1988min}, \vsp remains NP-complete for planar graphs with a maximum degree of three. 
From Lemma~\ref{l:vs-rb} and Theorem~\ref{t:mrb-hard} and the corresponding proofs, if we can show that we can convert an arbitrary planar graph with maximum degree three into a corresponding \toro instance, then we are done. That is, all we need to show is that, if Fig.~\ref{fig:vsp}(a) is a planar graph with maximum degree three, we can construct a \toro instance for which the dependency graph is Fig.~\ref{fig:vsp}(b). \\
We proceed to show something stronger: instead of showing the above for planar graphs with a maximum degree of three, we will do so for all planar graphs. By \cite{istvan1948straight}, all planar graphs can be drawn in the plane without edge crossings using only straight-line edges. Given an arbitrary planar graph and one of its straight-line-edge non-crossing embedding in the plane, we show how we can convert the embedding to a corresponding \toro instance.
\begin{figure}[h]
    \centering
\begin{overpic}
[width=0.6\columnwidth]{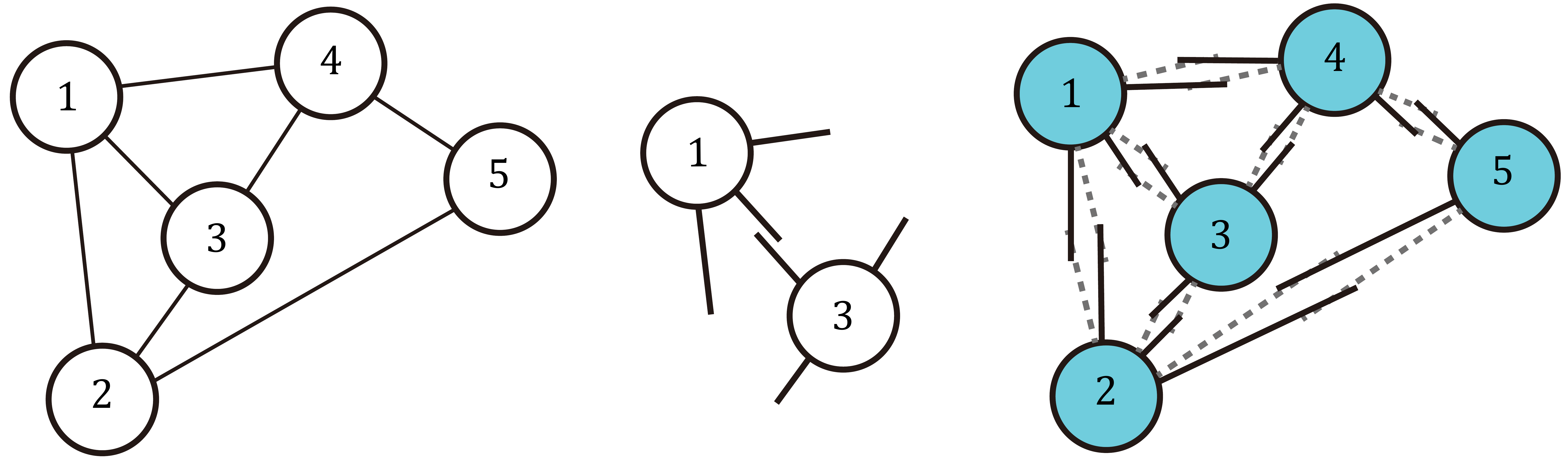}
\put(14, -3){{\small (a)}}
\put(45, -3){{\small (b)}}
\put(82, -3){{\small (c)}}
\end{overpic}
\vspace{3mm}
\caption{(a) An undirected planar graph with five vertices (b)
``Object gadgets'' for two of the vertices. (c) The corresponding \toro instance.}
\label{fig:mrb-toro-hard}
\end{figure} 
\linebreak\indent The construction process is explained using Fig.~\ref{fig:mrb-toro-hard} as an illustration, where Fig.~\ref{fig:mrb-toro-hard}(a) shows a straight-line-edge non-crossing embedding of planar graph. We construct \emph{object gadgets} based on the embedding's geometric structure as follows. For each node of the graph, we take the node and a bit more than half of each of the edges coming out of it. For example, for nodes $1$ and $3$, the corresponding objects are given in Fig.~\ref{fig:mrb-toro-hard}(b). We now construct a \toro instance with a dependency graph corresponding to replacing edges of Fig.~\ref{fig:mrb-toro-hard}(a) with bidirectional edges. To do so, we place each object gadget as they appear in Fig.~\ref{fig:mrb-toro-hard}(a). For the goal configuration, we rotate each object \emph{counterclockwise} by some small angle $\varepsilon$ around the center of the corresponding node. For the start configuration, we perform a similar rotation but in the \emph{clockwise} direction. The process yields Fig.~\ref{fig:mrb-toro-hard}(c) for  Fig.~\ref{fig:mrb-toro-hard}(a). 
It is straightforward to check that the construction converts each edge $(i, j)$ in the planar graph into a mutual dependency between two objects $i$ and $j$. For example, there is a cyclic dependency between object gadgets $1$ and $3$ in Fig.~\ref{fig:mrb-toro-hard}(c).
% \qed
\end{proof}
}

\section{Lower and Upper Bounds on \mrb}\label{sec:bounds}
\b{After connecting computing \mrb to vertex linear ordering and proving the computational intractability,} we proceed to establish \b{quantitative} bounds on \mrb, i.e., what is the lowest possible \mrb for \lrbm and \urbm, and what is the best that we can do to lower \mrb? 

\subsection{Intrinsic \mrb Lower Bounds} 
\begin{figure}[h]
    \centering    \includegraphics[width=0.2\textwidth]{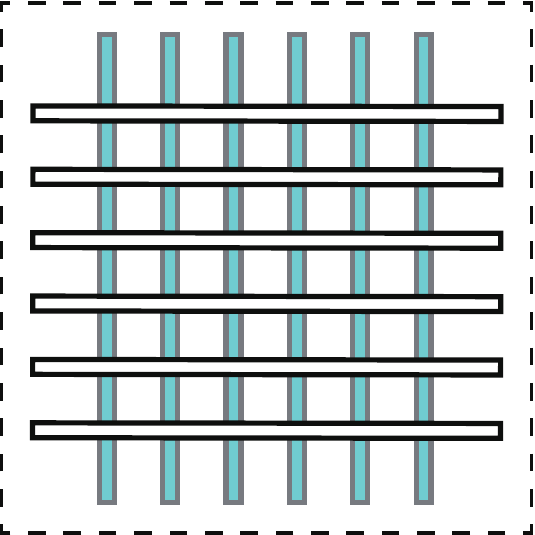}
    \caption{An instance with 6 cuboids where horizontal and vertical sets represent 
    start and goal poses, respectively. An arbitrary labeling of the objects may be 
    given in the labeled setting.}
    \label{fig:stick}
\end{figure}

When there is no restriction on object geometry, \mrb can easily reach the maximum 
possible $n -1$ for an $n$ object instance, even in the \urbm case. An example of when this happens is given in Fig.~\ref{fig:stick}, where $n = 6$ thin cuboids are aligned horizontally 
in $\mathcal A_1$, one above the other. The cuboids are vertically aligned in 
$\mathcal A_2$. Every pair of start pose and goal pose then induce a (unique) dependency.
Clearly, this yields a bidirectional $K_6$ labeled dependency graph in the \lrbm 
case and a $K_{6, 6}$ unlabeled dependency graph in the \urbm case. For both, 
$n - 1 = 5$ objects must be moved to buffers before the problem can be resolved. 
The example readily generalizes to an arbitrary number of objects. 

\begin{proposition}
\mrb lower bound is $n - 1$ for $n$ objects for both \lrbm and \urbm, which is the maximum possible, even for uniform convex objects. 
\end{proposition}

%\jy{The figure can be updated to be thin rectangles/cuboids.}

The lower bound on \mrb being $\Omega(n)$ is undesirable, but it seems to depend on having 
objects that are thin; everyday objects are not often like that. An ensuing question of 
high practical value is then: what happens when the footprint of the objects is ``fat''? 
Next, we establish that, the lower bound drops to $\Omega(\sqrt{n})$ for uniform cylinders, 
which approximate many real-world objects/products. Furthermore, we show that this lower 
bound is tight for \urbm (in Section~\ref{subsec:upper}).

\begin{figure}[h!]
    \centering
    \includegraphics[width=0.5\textwidth]{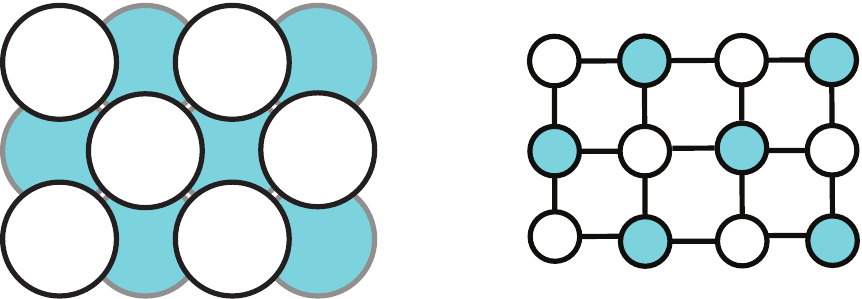}
    \caption{A \urbm instance (left) and its unlabeled dependency graph (right), a 
    $4\times 3$ dependency grid. Unshaded (resp., shaded) discs/vertices indicate start poses (resp., goal poses).}
    \label{fig:DependencyGrid}
\end{figure}

For convenience, assume $n$ is a perfect square, i.e., $n = m^2$ for some integer $m$. 
To establish the $\Omega(\sqrt{n})$ lower bound, a grid-like unlabeled dependency graph is used, which we call a \emph{dependency grid}, where $\mathcal A_1$ and $\mathcal A_2$ have similar 
patterns with $\mathcal A_2$ offset from $\mathcal A_1$ to the left (or right/above/below) by the length of one grid edge. An illustration of a portion of such a setup is given in Fig.~\ref{fig:DependencyGrid}. We use 
$\D(w, h)$ to denote a dependency grid with $w$ columns and $h$ rows. 

\begin{lemma}\label{l:urbm-lower}
Given a \urbm instance with $n = m^2$ objects and whose dependency graph is $\D(m, 2m)$, 
its \mrb is lower bounded by $\Omega(m) = \Omega(\sqrt{n})$
\end{lemma}

\b{Because the proof has limited relevance to the delivery of the main contributions of the paper, we highlight the main idea and refer the readers to the appendix for the complete proof. What we show is that, for certain unlabeled \toro instance with $n$ objects with such a dependency graph, at least $\Omega(\sqrt{n})$ objects must be moved to buffers simultaneous to solve the \toro instance.}  

Because \urbm  \  instances always have lower  or equal \mrb than  the \lrbm  \  instances with the same objects and 
goal placements, the conclusion of Lemma~\ref{l:urbm-lower} directly applies to 
\lrbm. Therefore, we have 

\begin{theorem}
For both \urbm and \lrbm with $n$ uniform cylinders, \mrb is lower bounded by 
$\Omega(\sqrt{n})$.
\end{theorem}

\begin{figure}[h!]
    \centering
    \includegraphics[height=1.4in]{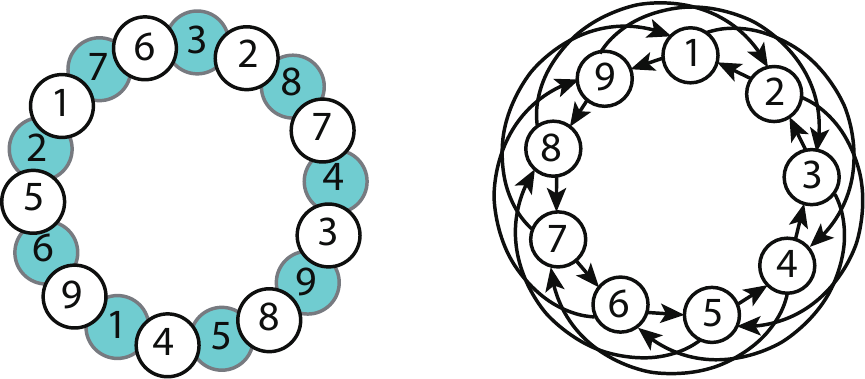}
    \caption{An example of a 9-object \lrbm yielding $\Omega(\sqrt{n})$ \mrb (left)
    and the corresponding dependency graph (right).}
    \label{fig:lrbm-cycle}
\end{figure}

For uniform cylinders, while the lower bound on \urbm is tight (as shown in
Section~\ref{subsec:upper}), we do not know whether the lower bound on \lrbm is tight;
our conjecture is that $\Omega(\sqrt{n})$ is not a tight lower bound for \lrbm. 
Indeed, the $\Omega(\sqrt{n})$ lower bound can be realized when uniform cylinders are 
simply arranged on a cycle, an illustration of which is given in Fig.~\ref{fig:lrbm-cycle}. 
For a general construction, for each object $o_i$, let $o_i$ depend on $o_{(i-1\mod n)}$ 
and $o_{(i+\sqrt{n}\mod n)}$, where $n$ is the number of objects in the instance. From the 
labeled dependency graph, we can construct the actual \lrbm instance where start and 
goal arrangements both form a cycle. 
It can be shown that when $n/2$ objects are at the goal poses, $\Omega(\sqrt{n})$ objects are 
at the buffer. We omit the proof, which is similar in spirit to that for Lemma~\ref{l:urbm-lower}. %We consider each $mod \sqrt{n}$ class of objects $M_i=\{o_j:j\equiv i mod n\}$. Denote the number of objects in $M_i$ at the goal poses by $g_i$. Suppose that there are $m$ $mod \sqrt{n}$ classes with $1\leq g_i \leq \sqrt{n}-1$. Similar to the proof of Thm. \ref{thm:unlabeled}, no matter $m=o(\sqrt{n})$ or $\Theta(\sqrt{n})$, the number of objects at the buffer is $\Omega(\sqrt{n})$.

\subsection{Upper Bounds on \mrb  \  for  \urbm}\label{subsec:upper}
We now establish, regardless of how $n$ uniform cylinders are to be 
rearranged, the corresponding \urbm instance admits a solution using 
$\mrb = O(\sqrt{n})$. Lower and upper bounds on \urbm agree;
therefore, the $O(\sqrt{n})$ bound is tight. 

To prove the upper bound, We propose an $O(n\log(n))$-time algorithm \spp for the 
setting based on a vertex separator of $\udg$.
\spp computes a sequence of goal vertices to be removed from the dependency graph.
Given a sequence of goal vertices to be removed, 
the running buffer size at each moment equals
$
\max(0,\|N(g,\udg)\|-\|g\|)
$,
where $g$ is the set of removed goal vertices at this moment,
and $N(g,\udg)$ is the set of neighbors of $g$ in $\udg$.
We prove that \spp can find a rearrangement plan 
with $ O(\sqrt{n})$ running buffers.

\begin{algorithm}
\begin{small}
    \SetKwInOut{Input}{Input}
    \SetKwInOut{Output}{Output}
    \SetKwComment{Comment}{\% }{}
    \caption{ \spp}
		\label{alg:spp}
    \SetAlgoLined
		\vspace{0.5mm}
    \Input{$\udg(V,E)$: unlabeled dependency graph}
    \Output{$\pi$: goal sequence}
		\vspace{0.5mm}
		$\pi,V,E\leftarrow$ RemovalTrivialGoals($\udg(V,E)$)\\
        \lIf{$V$ is $\emptyset$}{
        \Return $\pi$
        }
        $A, B, C \leftarrow$ Separator($\udg(V,E)$)\\
        $\pi \leftarrow \pi + g(C)$\\
        $A' \leftarrow A-N(g(C),\udg(V,E))$\\
        $B' \leftarrow $\ $B$$-N(g(C),\udg(V,E))$\\
		$\pi_{A'}\leftarrow $ \spp($\udg(A',E(A'))$)\\
		$\pi_{B'}\leftarrow $ \spp($\udg(B',E(B'))$)\\
		\lIf{ $\delta(A')\geq \delta(B')$}{
		    $\pi\leftarrow \pi + \pi_{A'} + \pi_{B'}$
		}
		\lElse{
		$\pi\leftarrow \pi + \pi_{B'} + \pi_{A'}$
		}
		\Return $\pi$\\
\end{small}
\end{algorithm}

\begin{figure}[h!]
    \centering
\begin{overpic}
[width=0.6\textwidth]{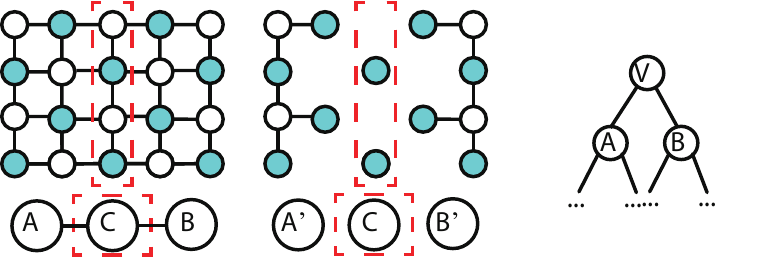}
\put(12,-5){(a)}
\put(46,-5){(b)}
\put(82,-5){(c)}
\end{overpic}
\vspace{4mm}
    \caption{The recursive solver \spp for \urbm. (a) A $O(\sqrt{|V|})$ vertex separator for the planar dependency graph. (b) By removing the start vertices in $C$ and the neighbors of the goal vertices in $C$, the remaining graph consists of two independent subgraphs and isolated goal poses in $C$. (c) The problem can be solved by recursively calling \spp.}
    \label{fig:separator}
\end{figure}

The algorithm is presented in Algo.~\ref{alg:spp}. 
\spp consumes a graph $\udg(V,E)$, which is a subgraph of $\udg$ induced by vertex set $V$.
To start with, the isolated goal vertices or those with only one dependency in $\udg(V,E)$ can be removed without using buffers (Line 1).
After that, $V$ can be partitioned into three disjointed subsets $A$, $B$ and $C$ \cite{lipton1979separator} (Line 3), 
such that there is no edge connecting vertices in $A$ and $B$, $|A|,|B|\leq 2|V|/3$, and $|C|\leq 2\sqrt{2|V|}$ (Fig.~\ref{fig:separator}(a)). 
For the start vertices in $C$ and the neighbors of the 
goal vertices in $C$, we remove them from $\udg$. 
Since there are at most $5$ neighbors for each goal vertex, there are at most $10\sqrt{2|V|}$ objects moved to the buffer in this operation. 
After that, we remove the goal vertices 
in $C$,
which should be isolated now (Line 4). 
Function $g(\cdot)$ obtains the goal vertices in a given vertex set.
Let $A'$, $B'$ be the remaining vertices in $A$ and $B$ (Line 5-6). 
 And correspondingly, let $C'$ be the removed vertices, i.e., $C':=(A\bigcup B \bigcup C)\backslash (A'\bigcup B')$.
Function $N(\cdot,\cdot)$ obtains the neighbors of a vertex set in a given dependency subgraph.
With the removal of  $C'$ from $\udg$, $A'$ and $B'$ form two independent subgraphs (Fig.~\ref{fig:separator}(b)). 
We can deal with the subgraphs one after the other by 
recursively calling \spp(Fig.~\ref{fig:separator}(c)) (Line 7-8). 
Let $\delta(V'):= |g(V')|-|s(V')|$ where $g(V')$ and $s(V')$ are the goal and start vertices in a vertex set $V'$ respectively. 
Between vertex subsets $A'$ and $B'$, we prioritize the one with larger $\delta(\cdot)$ value(Line 9-10).  That is because, after solving a rearrangement subproblem induced by a vertex set $V'$, there is $\delta(V')$ fewer objects in buffers or $\delta(V')$ more available goal poses.

As shown in detail in the appendix, \spp guarantees an \mrb upper bound of  $\dfrac{20}{1-\sqrt{2/3}}\sqrt{n}$.

\begin{theorem}\label{t:urbm-upper}
For \urbm with $n$ uniform cylinders, a polynomial time algorithm can compute a
plan with $O(\sqrt{n})$ \rb, which implies that \mrb is  bounded by $O(\sqrt{n})$.
\end{theorem}

%\begin{proof}See Sec.~\ref{sec:proofs}.\end{proof}

\section{Algorithms}\label{sec:algorithms}
In this section, \b{several algorithms for computing solutions for \toro tasks under various \mrb objectives are described. The hardness result (Theorem~\ref{t:mrb-toro-hard}) suggests that the challenge in computing \mrb solutions for \toro resides with filtering through an exponential number of vertex linear orderings of the dependency graph. This leads to our algorithms being largely search-based, which are enhanced with many effective heuristics.}

We first describe a dynamic programming-based method for 
\lrbm (Sec.~\ref{subsec:dp}).
Then, we propose a priority queue-based method in Sec.~\ref{subsec:pqs} for \urbm. 
Furthermore, a significantly faster 
depth-first modification of DP for computing \mrb is provided in 
Sec.~\ref{subsec:dfsdp}.
Finally, we also developed an integer
programming model, denoted $\ilpmrb$, for computing the minimum total number 
of buffers needed subject to the \mrb constraint, which is compared 
with the algorithm that computes the minimum total buffer without 
the \mrb constraint, denoted as $\ilpfvs$, from \cite{han2018complexity}.
A brief description of $\ilpmrb$ is given in Sec.~\ref{subsec:ilp}.

\subsection{Dynamic Programming (DP) for \lrbm}\label{subsec:dp}
%For \lrbm, a rearrangement plan can be represented with an ordering of the objects based on the time moving out of the start poses or the time moving into the goal poses. Due to the high degree of freedom in \lrbm, a class of rearrangement plans may be represented by the same object ordering. We propose strategies to find the optimal plan represented by a certain object ordering. In this way, we transform \lrbm into a permutation problem of objects. After that, the optimal ordering can be found with a dynamic programming(DP) algorithm.
As established in Sec.~\ref{subsec:logvrb}, a rearrangement plan in \lrbm can be represented as a linear ordering of object labels.
%where objects will be moved out of the start pose based on the order. 
That is, given an ordering of objects, $\pi$, we start with $o_{\pi(1)}$. 
If $x^g_{\pi(1)}$ is not occupied, then $o_{\pi(1)}$ is directly moved there. Otherwise, it is moved to a buffer location. We then continue with the second object in the order, and so on. After we work with each object in the given order, we always check whether objects in buffers can be moved to their goals, and do so if an opportunity is present. 
We now describe a \emph{dynamic programming} (DP) algorithm for computing such a linear ordering that yields the \mrb. 

%The strategies denoted by \StratStart and \StratGoal, are shown below. Given an object ordering $\pi$, \StratStart and \StratGoal interpret $\pi$ as an ordering of objects sorted by the increasing time of leaving the start poses and arriving at the goal poses respectively. Among the class of rearrangement plans represented by the same ordering, the one found by \StratStart follows the rule that objects in the buffer will go to the goal poses as soon as possible. Similarly, the plan found by \StratGoal follows the rule that objects will not go to the buffer until they have to.
%\begin{definition}[\StratStart]
%For each object $o_i$ in the order of $S$
%\begin{enumerate}
%    \item If $c^g_i$ is accessible, move $o_i$ to $c^g_i$. Otherwise, move $o_i$ to the buffer.
%    \item For each object $o_j$ in the buffer, move $o_j$ to $c^g_j$ if $c^g_j$ is accessible after the movement of $o_i$.
%\end{enumerate}
%\end{definition}

%\begin{definition}[\StratGoal]
%For each object $o_i$ in the order of $S$
%\begin{enumerate}
%    \item Move all the objects occupying $c^g_i$ to the buffer.
%    \item Move $o_i$ to $c^g_i$
%\end{enumerate}
%\end{definition}

%\begin{corollary}\label{cor:strategies}
%By enumerating all the object orderings, both \StratStart and \StratGoal can find the optimal %rearrangement plan.
%\end{corollary}
%The proof of Corollary~\ref{cor:strategies} is available in the extended version of this paper on arXiv. 

%\subsubsection{Dynamic Programming Approach}
The pseudo-code of the algorithm is given in Algo.~\ref{alg:dp}. The algorithm maintains 
a search tree $T$, each node of which represents an arrangement where a set of objects
$S$ have left the corresponding start poses. We record the objects currently at the 
buffer ($T[S]$.b) and the minimum running buffer from the start arrangement 
$\mathcal{A}_1$ to the current arrangement ($T[S].\mrb$). The DP starts with an empty 
$T$. We let the root node represent $\mathcal A_1$ (Line 1). At this moment, there is
no object in the buffer and the \mrb is 0(Line 2-3). And then we enumerate all the arrangements with $|S|=$ 1, 2, $\cdots$ and finally $n$(Line 4-5). For arbitrary $S$, 
the objects at the buffer are the objects in $S$ whose goal poses are still occupied by 
other objects (Line 6), i.e., $\{o\in S | \exists o' \in \mathcal{O} \backslash S, 
(o,o')\in A\}$, where $A$ is the set of arcs in $\ldg$. $T[S].\mrb$, the minimum running
buffer from the root node $T[\emptyset]$ to $T[S]$, depends on the last object $o_i$ 
added into $S$ and can be computed by enumerating $o_i$ (Line 7-20):
$$
\begin{array}{l}
    T[S].\mrb = \displaystyle\min_{o_i\in S}\max( T[S\backslash \{o_i\}].\mrb, \\ 
    \qquad\qquad\qquad\qquad\qquad |T[S\backslash \{o_i\}].b|\ +\text{TC}(S\backslash \{o_i\},S)),
\end{array}
$$
where the \emph{transition cost} TC is given as 
$$
\text{TC}(S\backslash \{o_i\},S)=\begin{cases}1,\  o_i \in T[S].b,\\ 
0,\  otherwise,\end{cases}
$$
with $x^g_i$ currently occupied (Line 10), the transition cost is due to objects 
dependent on $o_i$ cannot be moved out of the buffer before moving $o_i$ to the 
buffer (Line 11). 
 Specifically, $T[S\backslash \{o_i\}].\mrb$ is the previous $\mrb$ and $|T[S\backslash \{o_i\}].b|\ +\text{TC}(S\backslash \{o_i\},S)$ is the running buffer size in the new transition. 
If $T[S].\mrb$ is minimized with $o_i$ being the last object in 
$S$ from the starts, then $T[S\backslash \{o_i\}]$ is the parent node of $T[S]$ 
in $T$ (Line 16-19). Once $T[\mathcal O]$ is added into $T$, $T[\mathcal{O}].\mrb$ 
is the \mrb of the instance (Line 23) and the path in $T$ from $T[\emptyset]$ to
$T[\mathcal{O}]$ is the corresponding solution to the instance.

\begin{algorithm}[h!]
\begin{small}
    \SetKwInOut{Input}{Input}
    \SetKwInOut{Output}{Output}
    \SetKwComment{Comment}{\% }{}
    \caption{Dynamic Programming for \lrbm}
		\label{alg:dp}
    \SetAlgoLined
		\vspace{0.5mm}
    \Input{$ \ldgg(\mathcal{O},A)$: labeled dependency graph}
    \Output{$\mrb$: the minimum number of running buffers}
		\vspace{0.5mm}
		$T.root \leftarrow \emptyset$ \\
		\vspace{0.5mm}
		$T[\emptyset].b \leftarrow\emptyset$ \Comment{{\footnotesize objects currently at the buffer}}
		\vspace{0.5mm}
		$T[\emptyset].\mrb \leftarrow 0$ \Comment{{\footnotesize current minimum running buffer}}
		\vspace{0.5mm}
		\For{$1\leq k\leq |\mathcal{O}|$}{
		\Comment{\footnotesize{enumerate cases where $k$ objects have left the start poses}}
		\For{$S\in \text{k-combinations of } \mathcal{O}$}{
		$T[S].b\leftarrow \{o\in S \mid \exists o' \in \mathcal{O} \backslash S, (o,o')\in A\}$\\
		\Comment{{\footnotesize Find the \mrb from $T[\emptyset]$ to $T[S]$}}
		$T[S].\mrb  \leftarrow \infty$\\
		\For{$o_i \in S$}{
		$parent = S\backslash \{o_i\}$\\
		\If{$o_i\in T[S].b$}{
		$RB \leftarrow \max(T[parent].\mrb$, $|T[parent].b|+1$)
		}
		\Else{
		$RB \leftarrow  T[parent].\mrb$ 
  %$\hcancel[black]{\max(T[parent].\mrb, |T[S].b|)}$
		}
		\If{RB$<T[S].\mrb$}{
		$T[S].\mrb \leftarrow RB$\\
		$T[S].parent \leftarrow$  $parent$
		}
		}
		}
		}
		\vspace{0.5mm}
		\Return $T[\mathcal{O}]$.\mrb\\
\end{small}
\end{algorithm}

For the \lrbm instance in Fig.~\ref{fig:ex-prob}, Tab.~\ref{Tab:DP} shows $T[S].\mrb$ 
with different last-object options when $S=\{o_2, o_5, o_6\}$. If the last object $o_i$ is 
$o_5$, then we need to move $o_5$ into the buffer before moving $o_6$ out of the buffer. Therefore, 
even though the buffer size of the parent node and the current node are both 2, there is a 
moment when all of the three objects are at the buffer. However, when we choose $o_2$ or 
$o_6$ as the last object to add, the $T[S].\mrb$ becomes 2.
% \begin{center}
\begin{table}[h!]
    \caption{\label{Tab:DP}$T[S].\mrb$ for different last objects ($[p]= [parent]$)}
    \centering
    \resizebox{0.5\columnwidth}{!}{
    \begin{tabular}{p{4.0em}|p{4.0em}|p{3.4em}|p{3.4em}|p{3.8em}}
        \hline
        \small{Last object} & \small{$T[p].\mrb$} & \small{$T[p].b$} & \small{$T[S].b$} & \small{$T[S].\mrb$}\\
        \hline
        $o_2$ & 1 & \{$o_5$\} & \{$o_2$, $o_5$\} & 2\\
        $o_5$ & 2 & \{$o_2$, $o_6$\} & \{$o_2$, $o_5$\} & 3\\
        $o_6$ & 2 & \{$o_2$, $o_5$\} & \{$o_2$, $o_5$\} & 2\\
    \end{tabular}
    }
\end{table}
% \end{center}

\subsection{A Priority Queue-Based Method for \urbm}\label{subsec:pqs}
Similar to \lrbm, rearrangement plans for \urbm can be represented by a linear ordering of goal vertices in $\udg$. 
We can compute the ordering that yields \mrb by maintaining a search tree as we have done in Algo.~\ref{alg:dp}. 
Each node $T[g]$ in the tree represents an arrangement where a set of goal vertices $g$ have been removed from $\udg$. 
The remaining dependencies of $T[g]$ is an induced graph of $\udg$, 
formed from $V(\udg)\backslash (g\cup N(g,\udg))$ where $V(\udg)$ is the vertex set of $\udg$, $N(g,\udg)$ is the neighbors of $g$ in $\udg$. 
 When the goal vertices $g$ are removed from $\udgg$, the running buffer size is $max(0,|N(g)|-|g|)$,  i.e., the number of objects cleared away minus the number of available goal poses. Specifically, when $|N(g)|>|g|$, some objects are in buffers; when $|N(g)|<|g|$, some goal poses are available; when $|N(g)|=|g|$, the cleared objects happen to fill all the available goal poses. 
Given an induced graph $I(g)$, denote the goal vertices with no more than one neighbor in $I(g)$ as \emph{free goals}. 
We make two observations. 
First, given an induced graph $I(g)$, we can always prioritize the removal of free goals without optimality loss.
Second, multiple free goals may appear after a goal vertex is removed. For example, in the instance shown in Fig.~\ref{fig:ex-prob}(a), when the vertex representing $x^5_g$ is removed, $x^2_g$, $x^3_g$, and $x^4_g$ become free goals and can be added to the linear ordering in an arbitrary order.
 Denote nodes without free goals in the induced graph as \emph{key nodes}. 
In conclusion, the  key nodes
in the search tree may be sparse and enumerating  all possible nodes  with DP carries much overhead. 

As such, instead of exploring the search tree layer by layer like DP, we maintain 
a sparse tree with a priority queue $Q$. While each node still represents an 
arrangement, each edge in the tree represents  actions to clear out the next goal pose $x_i^g$. The corresponding child node of the edge represents the arrangement where $x_i^g$ and free goals are removed from the induced graph. 
%  We only add key nodes to $Q$, rather than all new nodes as A*. }
% We always pop out and develop the node with the smallest \mrb in $Q$. If a child node of the one that we develop already exists in the tree but is with a smaller \mrb than previously claimed, then we update the parent of the child node into the node we 
% are developing. 
% The \mrb of the node representing $\mathcal A_2$ sets an upper bound 
% of the solution and nodes in $Q$ with larger \mrb will be pruned away. 
% The algorithm terminates when $Q$ is empty.
%
 Similar to A*, we always pop out the key node with the smallest \mrb in $Q$. 
We denote this priority queue-based search method \PQS
% \b{, which can also be viewed as a variation of A*}
.

\subsection{Depth-First Dynamic Programming}\label{subsec:dfsdp}
Both \lrbm and \urbm can be viewed as solving a series of decision problems, i.e., 
asking whether we can find a rearrangement plan with $k= 1, 2, \ldots$ running buffers. 
As dynamic programming is applied to solve such decision problems, instead of 
performing the more standard breadth-first exploration of the search tree, we 
identified that a depth-first exploration is much more effective. 
We call this variation of dynamic programming \DFSDP, which is a fairly 
straightforward alteration of a standard DP procedure. 
%\DFSDP \cite{wang2021uniform}.
%
The high-level structure of \DFSDP is described in Algo.~\ref{alg:dfdp}. 
It consumes a dependency graph and returns the minimum running buffer size for the corresponding instance. 
Essentially, \DFSDP fixes the running buffer size RB and checks whether there is a plan requiring 
no more than RB running buffers. As the search tree (see Sec.~\ref{subsec:dp}) 
is explored, depth-first exploration is used instead of breadth-first (Line 4). 

\begin{algorithm}[ht!]
\begin{small}
    \SetKwInOut{Input}{Input}
    \SetKwInOut{Output}{Output}
    \SetKwComment{Comment}{\% }{}
    \caption{ Dynamic Programming with Depth-First Exploration}
		\label{alg:dfdp}
    \SetAlgoLined
		\vspace{0.5mm}
    \Input{$ \ldgg$: labeled dependency graph}
    \Output{RB: the minimum number of running buffers}
		\vspace{0.5mm}
		RB$\leftarrow 0$\\
		\While{not time exceeded}{
		 $T$.root $ \leftarrow \emptyset$ \\
		$T\leftarrow$ Depth-First-Search($T$,$\emptyset$, RB, $ \ldgg$)\\
		\lIf{ $\mathcal O \in T$}{\Return RB}
		\lElse{RB+=1}
		}
\end{small}
\end{algorithm}

The details of the depth-first exploration are shown in Algo.~\ref{alg:dfs}.
Similar to DP, each node in the tree represents an object state indicating whether an object is at the start pose, the goal pose, or external buffers.
And as described in Sec.~\ref{subsec:logvrb}, essential object states can be represented by the set of objects $S$ that have picked up from start poses.
If $S$ is $\mathcal O$, then all the objects are at the goal poses and we find a path on the search tree from the start state to the goal state (Line 1).
Given the set of objects away from the start poses $S$, we can get the sets of objects at start poses $S_S$, goal poses $S_G$, and external buffers $S_B$ (Line 2). Specifically, $S_S=\mathcal O\backslash S$, $S_G$ are the objects in $S$ that have no dependency in $S_S$, and $S_B$ are the objects in $S$ that have dependencies in $S_S$.
We further explore child nodes of the current state by moving one object away from the start pose (Line 3).
By checking the dependencies in $ \ldgg(\mathcal{O},A)$, we determine whether this object should be moved to buffers or its goal pose (Line 4).
The transition from $S$ to $S\bigsqcup \{o\}$ fails if the external buffers are overloaded or the child node has been explored (Line 5).
Otherwise, we can add $S\bigsqcup \{o\}$ into the tree (Line 7) and explore the new node (Line 8).
The algorithm returns a tree.

\begin{algorithm}[ht!]
\begin{small}
    \SetKwInOut{Input}{Input}
    \SetKwInOut{Output}{Output}
    \SetKwComment{Comment}{\% }{}
    \caption{Depth-First-Search}
		\label{alg:dfs}
    \SetAlgoLined
		\vspace{0.5mm}
    \Input{ $T$: Search Tree; $S$: The set of objects away from start poses; $RB$: Running buffer size; $ \ldgg(\mathcal{O},A)$: labeled dependency graph}
    \Output{$T$}
		\vspace{0.5mm}
		\lIf{$S$ is $\mathcal{O}$}{\Return $T$}
		$S_S$, $S_G$, $S_B$ = GetState($S$)\\
		\For{$o\in S_S$}{
		$ToBuffer$=CollisionCheck($S_S\backslash \{o\}$,$o$, $ \ldgg$)\\
		\lIf{(ToBuffer and $\|S_B\|+1>$RB) or ($S\bigsqcup \{o\}\in T$)}{\Return $T$}
		\Else{
		AddNode($T$, $S$, $S\bigsqcup \{o\}$)\\
		$T$ $\leftarrow$ Depth-First-Search($T$, $S\bigsqcup \{o\}$, RB, $ \ldgg$)\\
		\lIf{
		$\mathcal O\in T$}{\Return $T$}
		}
		}
		\Return $T$
\end{small}
\end{algorithm}

The intuition is that, when there are many rearrangement plans on the search tree that do not use more than $k$ running buffers, a depth-first search will quickly 
find such a solution, whereas a standard DP must grow the full search tree 
before returning a feasible solution. 
A similar depth-first exploration heuristic is used in \cite{wang2021uniform}. 
%With \DFSDP, we still develop a search tree as described in Sec.~\ref{subsec:dp}, 
%each node of which represents an arrangement with a certain set of objects moved 
%away from the start poses. When a search node is added to the tree, we only need 
%to check in linear time whether the transition from the parent arrangement to the 
%child arrangement can be done with $k$ running buffers. We develop the tree in a 
%depth first manner until either the node representing $\mathcal A_2$ is added into 
%the tree or all the leaf nodes in the tree are proven to be dead ends. T

\subsection{Minimizing Total Buffers Subject to \mrb Constraints}\label{subsec:ilp}

 We also construct a Mixed Integer Programming(MIP) model minimizing \mrb or total buffers.
Let binary variables $c_{i,j}$ represent  the dependency graph $\ldg$: $c_{i,j}=1$ if and only if $(i,j)$ is in the arc set of $\ldg$. Let $y_{i,j}(1\leq i<j\leq n)$ be the binary 
sequential variables: $y_{i,j}=1$ if and only if $o_i$ moves out of the start pose before $o_j$.
 $y_{i,j}$ are used to represent the ordering of actions.
% \mrb can be expressed based on three constraints: 
% First, \mrb is at least the size of the running buffer at any moment; 
% Second, an object $o_j\in \mathcal{O}$ is at the goal pose if and only 
% if all the objects $o_k\in \mathcal{O}$ with $c_{j,k}=1$ have left the start 
% poses. Third, an object $o_j\in \mathcal{O}$ is at the buffer if and only 
% if $o_j$ is neither at the start pose nor at the goal pose.
%
We further introduce two sets of binary variables $g_{i,j}$ and $b_{i,j}(1\leq i,j\leq n)$ to indicate object positions at each moment. 
$g_{i,j}=1$ indicates that $o_j$ has no dependency on other objects when moving $o_i$ from the start pose. In other words, the goal pose of $o_j$ is available at the moment. $b_{i,j}=1$ indicates that $o_j$ stays at the buffer after moving $o_i$ away from the start pose.
Finally, binary variables $B_i=1$ if and only if $o_i$ is moved to a buffer at some point.
The objective function consists of two terms: 
the total buffer term and running buffer term.
The total buffer term, scaled by $\alpha$, counts the number of objects that need buffer locations.
\mrb is represented with an integer variable $K$ and scaled by $\beta$.
To minimize total buffers subject to \mrb constraints,
we set $\alpha=1,\beta=n$.
The objective function is adaptable to different demands on rearrangement plans. 
Specifically, when $\alpha=0$,$\beta > 0$,
the MIP model minimizes \mrb.
When $\alpha > 0$, $\beta =0$, the MIP model minimizes total buffers, 
i.e. total actions in the rearrangement plan.
When $\alpha/\beta >n-1$, the MIP model first minimizes total buffers, 
and then minimizes running buffers.
When $\beta/\alpha >n-1$, the MIP model first minimizes running buffers, 
and then minimizes total buffers.

In the MIP model, Constraints~\ref{eq:c1} imply the rules for sequential variables  to make sure a valid ordering of indices $1,\dots,n$ can be encoded from $y_{i,j}$.
Constraints~\ref{eq:c2} imply that $B_j=1$ if $o_j$ has been to buffers in the plan.
Constraints~\ref{eq:c3} imply that \mrb $K$ is lower bounded by the maximum number of objects concurrently placed in buffers.
With Constraints~\ref{eq:c4} and \ref{eq:c5}, $g_{i,j}=0$ if and only if $o_j$ depends on an object $o_k$ which is still at the start pose when $o_i$ is moved.
With Constraints~\ref{eq:c6}-\ref{eq:c8}, 
$b_{i,j}=1$ if and only if $o_j$ is moved before $o_i$ and the goal pose is still unavailable when $o_i$ is moved from the start pose.
%\jy{``constraint x'' $\to$ ``Constraint x''}

\begin{equation}
    \arg \min \alpha [\sum_{i=1}^{n}B_i]+\beta K
\end{equation}
\begin{equation}\label{eq:c1}
    0\leq y_{i,j}+y_{j,k}-y_{i,k}\leq 1 \ \ \  \forall 1\leq i < j < k \leq n
\end{equation}
% \begin{equation}\label{eq:c2}
%     y_{i,i}=1 \ \ \  \forall 1 \leq i \leq n
% \end{equation}
\begin{equation}\label{eq:c2}
    B_j \geq \sum_{1\leq i\leq n} \dfrac{b_{i,j}}{n}\ \ \  \forall 1\leq j\leq n
\end{equation}
\begin{equation}\label{eq:c3}
    K \geq \sum_{1\leq j\leq n} b_{i,j}\ \ \  \forall 1\leq i\leq n
\end{equation}
\begin{equation}\label{eq:c4}
    \begin{split}
    \sum_{1\leq k < i} \dfrac{c_{j,k}(1-y_{k,i})}{n} + \sum_{i < k \leq n} \dfrac{c_{j,k}y_{i,k}}{n} \leq 1-g_{i,j} \\
    \forall 1\leq i,j \leq n
    \end{split}
\end{equation}
\begin{equation}\label{eq:c5}
    \begin{split}
    1-g_{i,j}\leq \sum_{1\leq k < i} c_{j,k}(1-y_{k,i}) + \sum_{i < k \leq n} c_{j,k}y_{i,k}\\ 
    \forall 1\leq i,j \leq n
    \end{split}
\end{equation}
\begin{equation}\label{eq:c6}
\begin{split}
    \dfrac{g_{i,j}+y_{i,j}}{2}\leq 1-b_{i,j} \leq g_{i,j}+y_{i,j}\\
    \forall 1\leq i<j \leq n
\end{split}
\end{equation}
\begin{equation}\label{eq:c7}
\begin{split}
    \dfrac{g_{j,i}+(1-y_{i,j})}{2}\leq 1-b_{j,i} \leq g_{j,i}+(1-y_{i,j})\\
    \forall 1\leq i<j \leq n
\end{split}
\end{equation}
\begin{equation}\label{eq:c8}
b_{i,i}=1-g_{i,i}
\end{equation}

% \vspace{2mm}
% The constraints can be added into the \fvs ILP formulation \cite{han2018complexity} to 
% find the minimum total buffer size with at most $k$ running buffers. We denote this method
% as $\ilptb$ when $k = \mrb$, and the \fvs method from \cite{han2018complexity} as $\ilpfvs$.

%Therefore, we have 
%two ILP solutions, $\ilpmrb$, which computes solutions optimizing \mrb, and $\ilpfvs$
%which computes the minimum number of total buffers subject to having minimum \mrb.
%Note that even though each vertex $o_i$ are split into $o^{in}_i$ and $o^{out}_i$ in that formulation, the object ordering is indicated by the ordering of $\{o^{in}_i\}_{i=1}^n$.
%\kg{Here I use the ordering of the in-vertices to represent the original object ordering}

\section{Applications to In-Place Rearrangement with Bounded Workspace}\label{sec:application}
Both \lrbm and \urbm allow external open space for temporary object displacements, 
computing rearrangement plans  for \toro with external buffers (\toroe).
However, there are many practical rearrangement scenarios where objects have to be displaced inside the workspace.
In this section, we apply the proposed algorithms to  \toro with internal buffers (\toroi).
\toroi seeks a feasible rearrangement plan minimizing the number of total actions.
While solving \toroe only deals with \emph{inherent} constraints defined by the start and goal poses,
some objects in \toroi may be temporarily displaced inside the workspace and induce further \emph{acquired} constraints.
For instance, to solve the problem in Fig.~\ref{fig:toro} with external buffers, 
we can move the Pepsi can to an external buffer to break the cycle, 
move the Coke first and then Fanta to their goal locations, and finally, bring back the Pepsi can into the workspace. 
In \toroi, we must find a temporary location for the Pepsi can in $\mathcal W$. 
If the buffer location overlaps with the goal of the Coke can (or the Fanta can),
then the coke can (or the Fanta can) depends on the Pepsi can again.
To avoid these acquired constraints,
the buffer location needs to avoid the goals of Coke and Fanta.
Due to acquired constraints arising from internal buffer selection, \toroi, 
the problem we study in this section, 
is more challenging than \toroe. 
Intuitively, selecting buffers inside the workspace (\toroi) is much more difficult and constrained than using buffers outside the workspace (\toroe) to store displaced objects. 
Since \toroe has been shown to be computationally intractable \cite{han2018complexity}, and is  equivalent to  a special case of \toroi  
  where the workspace is large enough that collision-free buffer locations are guaranteed,
\toroi is also NP-hard.
 Additionally, in \toroi, it may be challenging to allocate valid buffer locations so it is necessary to limit the running buffer size. With this observation, we compute \toroi solutions based on methods for \lrbm and \urbm. 

In this section, We propose Tabletop Rearrangement with Lazy Buffers (\trlb), an effective framework for solving \toroi based on algorithms mentioned in Sec.~\ref{sec:algorithms}. We first describe a rearrangement solver with lazy buffer allocation (Sec.~\ref{sec:LazyBufferGeneration}), where buffer allocation is delayed after DFDP computes a ``rough'' schedule of object movements. 
Finally, a preprocessing routine based on DFDP for \urbm helps with further speedups (Sec.~\ref{sec:Preprocess}).
To enhance scalability to larger and more cluttered instances, 
the \trlb framework (Sec.~\ref{sec:PartialPlan}) recovers from buffer allocation failures.

\subsection{Lazy Buffer Allocation}\label{sec:LazyBufferGeneration}
% \textcolor{red}{There should be a general introduction of the method at a high level, i.e., how is a problem solved in phases and the intermediate data structure examples using a running example.}
When an object stays at a buffer, it should avoid blocking the upcoming manipulation actions of other objects. Otherwise, either the object in the buffer or the manipulating object has to yield, which increases the number of necessary actions. In other words, we need to carefully choose acquired constraints. If we know the schedule of other objects in advance, a buffer can be selected to minimize unnecessary obstructions. This observation motivates solving the rearrangement problem in two steps: First, compute a \emph{primitive plan}, which is an incomplete schedule ignoring acquired constraints; second, given the incomplete schedule as a reference, generate buffers to optimize the selection of acquired constraints.

%Before moving an object to a goal pose, 
%other objects occupying the pose have to temporarily move to buffer locations if their own goal positions are inaccessible.
% Similar ideas are used in NAMO(Navigation Among Movable Obstacles) problems.
% Stilman et al. used reverse search methods to constrain prior object displacements with future manipulations\cite{stilman2007planning, stilman2008planning}.
% However, 
% these methods, solving combinatorial and geometric challenges at the same time,
% are computationally expensive.

\subsubsection{Primitive Plan}
To compute a \emph{primitive plan}, we assume enough free space is available so that no acquired constraints will be created. This transforms the problem into a \toroe problem, where each object is displaced at most once before it moves to the goal pose. Then, an object $o_i \in \mathcal O$ can have three \emph{primitive} actions:   
%large enough to contain all the objects, 
%for temporary object placements.
%With the large empty area,
%there is no reason to move an object from a buffer location to another.
%Therefore, an object $o_i \in \mathcal O$ can have three primitive actions: 
\begin{enumerate}
    \item $(o_i, s\rightarrow g)$: moving from  $x^s_i$ to  $x^g_i$;
    \item $(o_i, s\rightarrow b)$: moving from  $x^s_i$ to a buffer;
    \item $(o_i, b\rightarrow g)$: moving from a buffer to  $x^g_i$.
\end{enumerate}
A primitive plan is a sequence of primitive actions;
%, 
%moving objects from $\mathcal A_s$ to $\mathcal A_g$ without collision.
%Similar to \toroe, 
computing such a plan is similar to finding a linear vertex ordering \cite{adolphson1973optimal, shiloach1979minimum} of the dependency graph. 
Since we need to allocate buffer locations inside the workspace in the next phase of the algorithm, 
it is beneficial to limit the number of buffer positions that need to be concurrently allocated. 
%\kg{I add one sentence here in order to make the connection between RBM and TRLB stronger and also indicate the reason for choosing DFDP. }
We use DFDP to achieve this, 
which minimizes the number of running buffers. 

\subsubsection{Buffer Allocation}
%In many practical cases, 
Free space inside the workspace $\mathcal{W}$ is scarce in cluttered spaces (e.g.,  Fig.~\ref{fig:density}) and acquired constraints must be dealt with through  the careful allocation of buffers inside $\mathcal{W}$. We apply a greedy strategy to find feasible buffers based on a primitive plan (Algo. \ref{alg:buffer}). The general idea is to incrementally add constraints on the buffers until we find feasible buffers for the whole primitive plan or terminate at a step where there are no feasible buffers for the primitive plan. In Algo.~\ref{alg:buffer},~$\mathcal O_s, \mathcal O_g$,  and $\mathcal O_b$ are the sets of objects currently at start poses, goal poses and buffers respectively.

%
%We do so based on a primitive plan.
%
%Since primitive plans are by nature under a \emph{two-step assumption}: 
%each object in $\mathcal O$ moves at most twice,
%we follow the assumption in this subsection,
%and more general cases will be discussed in Sec. \ref{sec:PartialPlan}.
\begin{algorithm}
\begin{small}
    \SetKwInOut{Input}{Input}
    \SetKwInOut{Output}{Output}
    \SetKwComment{Comment}{\% }{}
    \caption{ Buffer Allocation}
		\label{alg:buffer}
    \SetAlgoLined
		\vspace{0.5mm}
    \Input{$\pi$: a primitive plan;  $\mathcal A_1=\{x^s_1,...,x^s_n\}$: start arrangement;  $\mathcal A_2=\{x^g_1,...,x^g_n\}$: goal arrangement}
    \Output{$B$: buffers; TerminatingStep: the action step where buffer generation fails, $\infty$ if Success.}
		\vspace{0.5mm}
		$\mathcal O_s$ $\leftarrow$ $\mathcal O$; 
		$\mathcal O_g$, $\mathcal O_b$ $\leftarrow$ $\emptyset$; 
		$B \leftarrow$ RandomPoses($\mathcal O$)\\
             Constraints $\leftarrow$ InitializeConstraints();\\
		\For{   $(o_i, m)\in \pi$}{
    		\If{$m$ is s $\rightarrow$ b}{
        		$\mathcal O_b$.add($o_i$)\\
        		Constraints[$o_i$]$\leftarrow$GetPoses($\mathcal O_s \bigcup \mathcal O_g - \{o_i\}$)\\
    		}
    		\ElseIf{$m$ is b $\rightarrow$ g}{
        		\lFor{$o\in \mathcal O_b\backslash\{o_i\}$}{
        		Constraints[$o$].add(  $x^g_i$})
    		}
    		}
    		\Else{
        		%\Comment{$m$ is s $\rightarrow$ g}
        		\lFor{$o\in \mathcal O_b$}{
        		    Constraints[$o$].add(  $x^g_i$})
        	
    		}
    		Success, $B'$ $\leftarrow$ BufferGeneration($\mathcal O_b$, Constraints, $B$)\\
    		\If{Success}{
    		$B$ $\leftarrow$ $B'$\\
    		$\mathcal O_s, \mathcal O_g, \mathcal O_b \leftarrow$ UpdateState($\mathcal O_s, \mathcal O_g, \mathcal O_b$)\\
    		}
    		\lElse{
    		\Return $B$, $\pi$.index(action)
    		}
		
		\vspace{0.5mm}
		\Return $B$, $\infty$\\
\end{small}
\end{algorithm}

We start with  $\mathcal A_1$ where all the objects are at start poses and the buffers are initialized at random poses (Line 1). 
 We then initialize a mapping from $\mathcal O$ to the power set of object poses, indicating the set of obstacles to avoid when allocating buffer poses for the object (Line 2). 
Each action in $\pi$ indicates an object $o_i$ that is manipulated and the action $m$ performed (Line  3). 
If $o_i$ is moved to a buffer (Line  4), then we add it into $\mathcal O_b$ (Line  5). 
The current poses of other objects in $O_s\bigcup O_g$ are seen as fixed obstacles for $o_i$ (Line  6). 
If $o_i$ is leaving the buffer (Line  8), then other objects in $\mathcal O_b$ should avoid the goal pose  $x^g_i$ of $o_i$ (Line  9). 
If $o_i$ is moving directly from  $x^s_i$ to  $x^g_i$ (Line  11, the ``else'' corresponds to $m$ being $s\to g$,  i.e., directly go from start to goal), then all buffers for objects in the current $\mathcal O_b$ need to avoid  $x^g_i$ (Line  12). 
After setting up acquired constraints, we generate new buffers for objects in $O_b$ to satisfy these constraints by either sampling or solving an optimization problem (Line  14). 
Old buffers in $B$ satisfying new constraints will be directly adopted. If feasible buffers are found (Line  15), then buffers and object states will be updated (Line  16-17). 
Otherwise, we return the feasible buffers computed and record the terminating step of the algorithm (Line  19). 
In the case of a failure, the returned buffers provide a \emph{partial plan}.
%, so the 
%break the two-step assumption and develop a robust planner which does not rely too much on the quality of primitive plans. 
%The details of this part is disclosed in Sec. \ref{sec:PartialPlan}.

\begin{figure}
    \vspace{2mm}
    \centering
    \includegraphics[width=0.6\columnwidth]{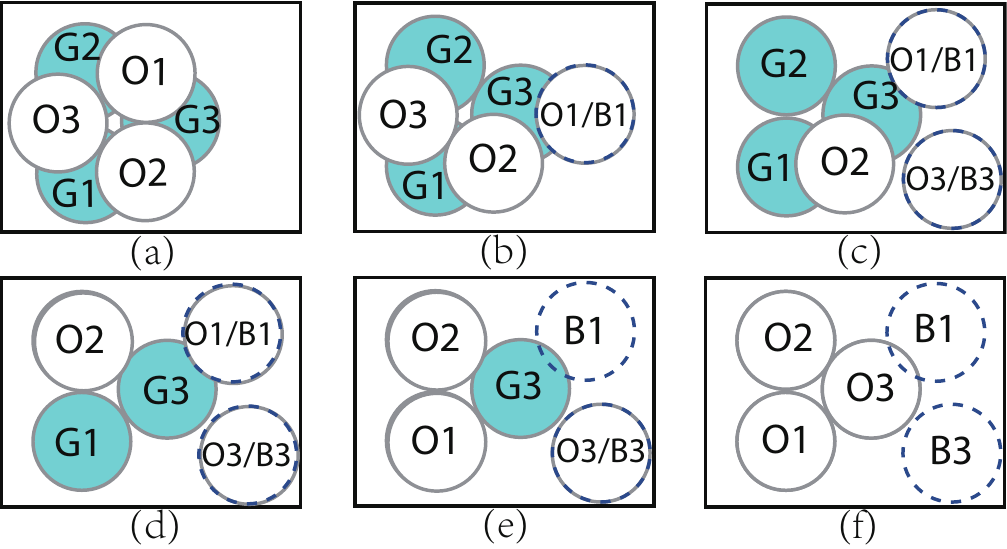}
    \vspace{-.1in}
    \caption{A working example with three objects defined in (a).
    The primitive plan is [($o_1$, s $\rightarrow$ b), ($o_3$, s $\rightarrow$ b), ($o_2$, s $\rightarrow$ g), ($o_1$, b $\rightarrow$ g), ($o_3$, b $\rightarrow$ g)].  Figures (b)-(f) show the steps of Alg.~\ref{alg:buffer} after each action. The unshaded discs with dashed line boundaries ($B_i$) represent the buffers satisfying constraints up to each step. For each object $o_i$, the unshaded discs with solid line boundaries ($O_i$) represent the current poses and the shaded ($G_i$) discs represent goal poses.}
    \label{fig:AlgoExample}
\end{figure}

Fig.~\ref{fig:AlgoExample} illustrates the buffer allocation process via an example. 
The unshaded  discs and shaded discs with solid line boundaries represent the current poses  and goal poses respectively. 
The discs with dashed line boundaries represent the allocated buffers. 
When we move $o_1$ to a buffer $B_1$ (Fig. \ref{fig:AlgoExample}(b)),  it only needs to avoid collision with  $x^s_2$ and  $x^s_3$.  
But as we move $o_3$ to a buffer,  $B_1$ needs to avoid $o_3$'s buffer $B_3$ as well. 
To satisfy the added constraint, $B_1$ will be reallocated. Since the new buffers $B_1$ and $B_3$ (Fig.~\ref{fig:AlgoExample}(c)) satisfy the constraints added in the following steps, they need not to be relocated. 
Note that the buffer originally selected for $o_1$ but then replaced will not appear in the resulting plan, i.e., $o_1$ will move directly to the new buffer (Fig.~\ref{fig:AlgoExample}(c)-(f)). Algo. \ref{alg:buffer} works with one strongly connected component of the dependency graph at a time, treating objects in other components as fixed obstacles.
%whose start poses are the same as goal poses.

%Based on the topological order of the strongly connected components of the dependency graph, 
%the rearrangement task can be decoupled into independent subproblems;
%The input primitive plan for each call of Algo. \ref{alg:buffer} only move objects in one strongly connected component in $G$. 
%The decomposition speeds up the method, especially in large scale problems.

Once the feasible buffers are found, all the primitive actions can be transformed into feasible pick-n-place actions inside the workspace. 
And therefore, the primitive plan can be transformed into a rearrangement plan moving objects from  $\mathcal A_1$ to  $\mathcal A_2$.
The function BufferGeneration is implemented by either sampling or solving an optimization problem, both of which are discussed below.

\paragraph{Sampling}\label{sec:Sampling}
Given the object poses that buffers need to avoid so far, feasible buffers can be generated by sampling poses inside the free space. When objects stay in buffers at the same time, we sample buffers one by one and previously sampled buffers will be seen as obstacles for the latter ones.

\paragraph{Optimization}\label{sec:Optimization}
For cylindrical objects $o_i$ at $(x_i,y_i)$ with radius $r_i$, and $o_j$ at $(x_j,y_j)$ with radius $r_j$, they are collision-free when
$(x_i-x_j)^2+(y_i-y_j)^2 \geq (r_i+r_j)^2$ holds. By further restricting the range of buffer centroids to assure they are in the workspace, the buffer allocation problem can be transformed into a quadratic optimization problem. For objects with general shapes, collision avoidance cannot be presented by inequalities of object centroids. We can construct the optimization problem with $\phi$ functions of the objects \cite{chernov2010mathematical} and solve the problem with gradients.

\subsection{Preprocessing}\label{sec:Preprocess}
\begin{figure}[h]
    \centering
    \includegraphics[width=0.6\columnwidth]{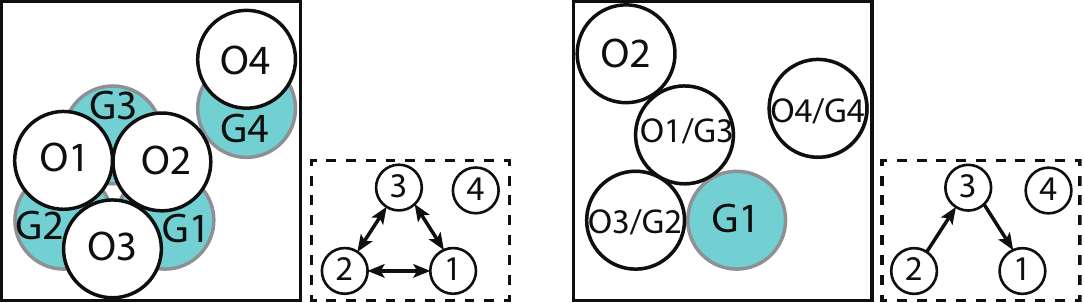}
    \vspace{-.1in}
    \caption{A four-object example of preprocessing. The unshaded and shaded discs represent current and goal arrangements respectively. Before preprocessing (left), two buffers need to be allocated synchronously. After  preprocessing (right), the problem becomes monotone.}
    \label{fig:preprocessing}
    \vspace{-.1in}
\end{figure}

In dense environments, allocating buffers is hard, motivating minimizing the number of running buffers, which is generally low even in high-density settings in \urbm. 
Based on this, for each component of the dependency graph that is not  isolated vertex or simple cycle, \b{we first solve it as if it is an unlabeled instance. After this \emph{preprocessing} step, all the components in the resulting labeled instance are either simple cycles ($\mrb=1$) or isolated vertices($\mrb=0$).}
%
%After preprocessing, we obtain a \toroi requiring at most one running buffer. 
Fig.~\ref{fig:preprocessing} shows an example of preprocessing.  
$o_1$, $o_2$, and $o_3$ form a complete graph, where at least two objects need to be placed at buffers simultaneously. 
We conduct preprocessing of the three-vertex component by moving $o_2$ to a buffer position, $o_1$ to  $x^g_3$ and $o_3$ to  $x^g_2$. 
$o_2$ will not move to  $x^g_1$ since it does not occupy other goal poses. The preprocessing step needs one buffer and the resulting rearrangement problem is monotone.

%Preprocessing may be further enhanced by only executing actions where the manipulating objects occupy at least two goal poses. 
%That is because the aim of preprocessing is to simplify cycles, not moving objects to others' goal poses.
%

\subsection{Solving  \toroi with Lazy Buffers (\trlb)}\label{sec:PartialPlan}
The \trlb framework builds on the insight that a new \toroi instance is generated when lazy buffer allocation fails. The new instance has the same goal  $\mathcal A_2$ as the original one but some progress has been made in solving the \toroi task. There are two straightforward implementations of \trlb: forward search and bidirectional search. In the first case, by accepting partial solutions, a rearrangement plan can be computed by developing a search tree $T$ rooted at  $\mathcal A_1$.  In the search tree $T$, nodes are feasible arrangements and edges are partial plans containing a sequence of collision-free actions. When buffer allocation fails, we add the resulting arrangement into the tree and resume the rearrangement task from a random node in $T$.  This randomness and the randomness in primitive plan computation and buffer allocation allows \trlb to recover from failures. 
\vspace{0.05in}

%In the case where the lazy buffer allocation fails, 
%planning can be resumed from resulting arrangements of the partial plans.
%With the search tree, the algorithm can find more general plans breaking the two-step assumption and relies less on the quality of primitive plans.

\begin{algorithm}[h]
\begin{small}
    \SetKwInOut{Input}{Input}
    \SetKwInOut{Output}{Output}
    \SetKwComment{Comment}{\% }{}
    \caption{\trlb with Bidirectional Search}
    \label{alg:BST}
    \SetAlgoLined
		\vspace{0.5mm}
    \Input{ $\mathcal A_1$, $\mathcal A_2$, $max\_time$}
    \Output{ Search trees: $T_1$, $T_2$}
		\vspace{0.5mm}
		$T_1$.root, $T_2$.root$\leftarrow$  $\mathcal A_1$,  $\mathcal A_2$\\
		\While{not exceeding $max\_time$}{
		$\mathcal A_{rand}\leftarrow$ RandomNode($T_1$)\\
		$\mathcal A_{new1} \leftarrow$ LazyBufferAllocation($\mathcal A_{rand}$, $T_2$.root)\\
		$T_1$.add($\mathcal A_{new1}$)\\
		\lIf{$\mathcal A_{new1}$ is $T_2$.root}{\Return $T_1$, $T_2$}
		$\mathcal A_{near}\leftarrow$ NearestNode($\mathcal A_{new1}$, $T_2$)\\
		$\mathcal A_{new2} \leftarrow$ LazyBufferAllocation($\mathcal A_{near}$, $\mathcal A_{new1}$)\\
		$T_2$.add($\mathcal A_{new2}$)\\
		\lIf{$\mathcal A_{new2}$ is $\mathcal A_{new1}$}{\Return $T_1$, $T_2$}
		$T_1, T_2 \leftarrow T_2, T_1$\\
		}
\end{small}
\end{algorithm}

%Besides conducting an one-directional search from $\mathcal A_s$ to $\mathcal A_g$, 
In bidirectional search, two search trees rooted at  $\mathcal A_1$ and  $\mathcal A_2$ are developed. This more involved procedure is shown in Algo. \ref{alg:BST}, which computes two search trees that connect  $\mathcal A_1$ and  $\mathcal A_2$.  In Line 1, the trees are initialized. For each iteration, we first rearrange between a random node $\mathcal A_{rand}$ on $T_1$ to the root node of $T_2$ (Line 3-5). The function LazyBufferAllocation refers to the overall algorithm developed in Sec. \ref{sec:LazyBufferGeneration}. A found path yields a feasible plan for \toroi (Line 6). Otherwise, we rearrange between the new arrangement $\mathcal A_{new1}$ and its nearest neighbor in $T_2$ (Line 7-9). If a path is found, then we find a feasible rearrangement plan for \toroi (Line 10). Otherwise, we switch the trees and attempt rearrangement from the opposite side (Line 11).

\begin{comment}
\begin{figure}[h!]
    \vspace{2mm}
\centering
    \begin{overpic}[width=1\columnwidth]{figures/BST-eps-converted-to.pdf}
    \end{overpic}
    \caption{Bidirectional search tree}
    \label{fig:BST}
\end{figure}
\end{comment}

% \subsection{Buffer Generation}\label{sec:BufferGeneration}
% \input{04_02_general_buffer_generation}

% \subsection{Robust Buffer Generation}\label{sec:GeneralBufferGeneration}
% \input{04_02_general_buffer_generation}

\section{Performance Evaluation of Simulations}\label{sec:experiments}
Our evaluation focuses on uniform cylinders, given their prevalence in 
practical applications. 
For simulation studies, instances with different object densities are created, as measured 
by \emph{density level} $\rho := n\pi r^2/(h*w)$, where $n$ is the number of objects and 
$r$ is the base radius. $h$ and $w$ are the height and width of the workspace.
%\jy{Kai: please update $D$ to $\rho$, which is a more standard symbol for density.}
In other words, $\rho$ is the proportion of the tabletop surface occupied by objects. 
%We notice that instances with a fixed $D$ have roughly the same number of dependencies on average for each object, regardless of the number of objects in the environment. 
%For example, when $D=0.3$, each object has averaging 1.25 dependencies. 

\begin{figure}[h!]
    \centering
    \includegraphics[width=0.5\columnwidth]{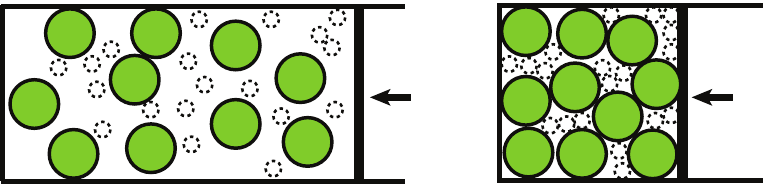}
    \vspace{1mm}
    \caption{Generating dense instances using a physics-engine based simulator
    through     compression of the left scene to the right scene.}
    \label{fig:compression}
\end{figure}

\begin{figure}[h!]
    \centering
    \includegraphics[width=0.5\columnwidth]{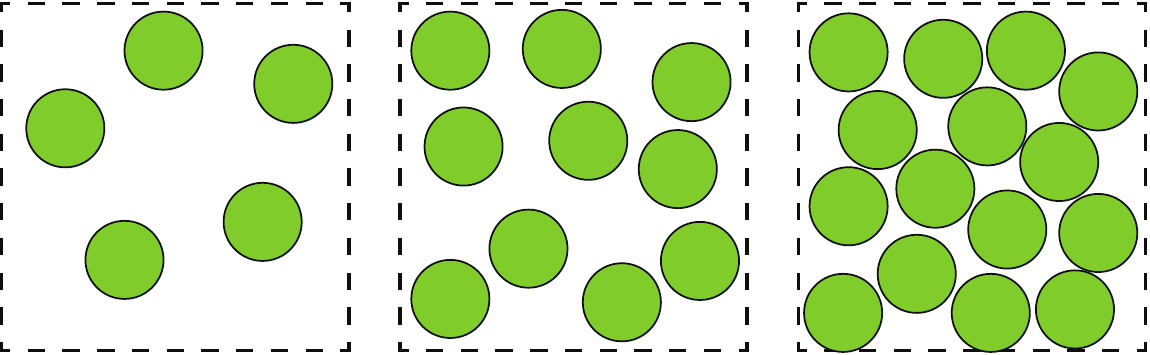}
    \caption{Unlabeled arrangements with $\rho=0.2, 0.4, 0.6$ respectively.}
    \label{fig:density}
\end{figure}

The evaluation is conducted on both random object placements and manually 
constructed difficult setups (e.g., dependency grids with $MRB = \Omega(\sqrt{n}$)).
For generating test cases with high $\rho$ value, we invented a physic engine 
(we used Gazebo \cite{koenig2004design}) based approach for doing so. Within a rectangular box, we sample 
placements of cylinders at lower density and then also sample locations for some 
smaller ``filler'' objects (see Fig.~\ref{fig:compression}, left). From here, 
one side of the box is pushed to reach a high density setting 
(Fig.~\ref{fig:compression}, right), which is very difficult to generate via random 
sampling. By controlling the ratio of the two types of objects, different density 
levels can be readily obtained. Fig.~\ref{fig:density} shows three random object placements 
for $\rho = 0.2, 0.4$ and $0.6$.

From two randomly generated object placements with the same $\rho$ and $n$ values, a 
\urbm instance can be readily created by superimposing one over the other. \lrbm
instances can be generated from \urbm instances by assigning each object a random 
label in $[n]$ for both start and goal configurations.

%In random scenario, we consider cylindrical objects, whose bottoms are unit discs, placed in a square environment without collisions. 
%The size of the square environment are determined by the density level $D$ and the number of objects $n$. 
%Unlabeled arrangements are generated under uniform distribution of objects: Object centroids are assigned one by one following the uniform distribution in the workspace.
%An object is placed if the assigned location is out of collision with previous-placed objects. Otherwise, other randomly generated locations will be assigned to it until a collision-free location is found. 
%For each instance, $\mathcal A_1$ and $\mathcal A_2$ are randomly chosen from the generated unlabeled arrangements and in \lrbm, the labels of the each object are also randomly assigned.
%
%When the desired density level is too high, the generation process becomes slow: collision-free positions are hard to find for objects at the bottom of the list. In this case, we generate the arrangements by compressing sparse instances(Fig.~\ref{fig:conpression}) in Gazebo, a simulation software\cite{koenig2004design}. To enrich the randomness of the object positions, smaller cylindrical obstacles are temporarily added into the environment so that the density is balanced all over the environment.
%
%In the special scenario, we evaluate the special cases discussed in the previous sections whose $\mrb=\Theta(\sqrt{n})$.

 For tabletop object rearrangement with external buffers (\toroe), evaluated methods in different scenarios(labeled or unlabeled) are presented in Tab.~\ref{tab:toroe}.
For tabletop object rearrangement with internal buffers (\toroi),
we present experiments comparing lazy buffer generation algorithms given different options(Tab.~\ref{tab:toroi}), including: (1) Primitive plan computation: running buffer minimization with \DFSDP (RBM), total buffer minimization with DP (TBM), random order (RO); (2) Buffer allocation methods: optimization (OPT), sampling (SP); (3) High-level planners: one-shot (OS), forward search tree (ST), bidirectional search tree (BST); and (4) With or without preprocessing (PP).
Here, the one-shot (OS) planner is using primitive plans and buffer allocation (Sec.~\ref{sec:LazyBufferGeneration}) without tree search (Sec.~\ref{sec:PartialPlan}). In OS, we attempt to compute a feasible rearrangement plan up to $30|\mathcal O|$ times before announcing a failure.
% Notice that at least $|\mathcal O|$ actions are required for solving any instance. 
A full \trlb algorithm is a combination of components, e.g.,  RBM-SP-BST stands for using the primitive plans that minimize running buffer size,  performing buffer allocation by sampling, maintaining a bidirectional search tree, and doing so without preprocessing.

To highlight the application of RBM in \toroi, we only compare methods in primitive plan computation and evaluate the effect of our preprocessing routine. A complete ablation study is presented in \cite{gao2021fast}.

\begin{table}[h]
\caption{Evaluated methods for \toroe. }
\label{tab:toroe}
% \resizebox{\columnwidth}{!}{
\begin{tabular}{ |p{0.25\columnwidth}|p{0.65\columnwidth}| } 
 \hline
 Problems & Methods \\ 
 \hline
 \lrbm & \DFSDP, DP, $\ilpfvs$, $\ilptb$ \\ 
 \hline
 \urbm & \DFSDP, \PQS \\ 
 \hline
\end{tabular}
% }
\end{table}

\begin{table}[h]
    \caption{Evaluated methods for \toroi. }
    \label{tab:toroi}
    \begin{tabular}{ |p{0.5\columnwidth}|p{0.4\columnwidth}| } 
     \hline
     Components & Methods \\ 
     \hline
     Primitive plan computation & RBM, TBM, RO\\ 
     \hline
     Buffer allocation methods & OPT, SP\\ 
     \hline
     High level planners & OS, ST, BST\\
     \hline
     Preprocessing & PP\\ 
     \hline
    \end{tabular}
\end{table}

The proposed algorithms are implemented in Python and all experiments are executed 
on an Intel$^\circledR$ Xeon$^\circledR$ CPU at 3.00GHz. For solving  MIP, Gurobi 9.16.0 \cite{gurobi} is used.

\vspace{-1mm}
\subsection{\lrbm over Random Instances }
\begin{figure}[h!]
    \vspace{4mm}
\centering
    \begin{overpic}[width=0.6\columnwidth]{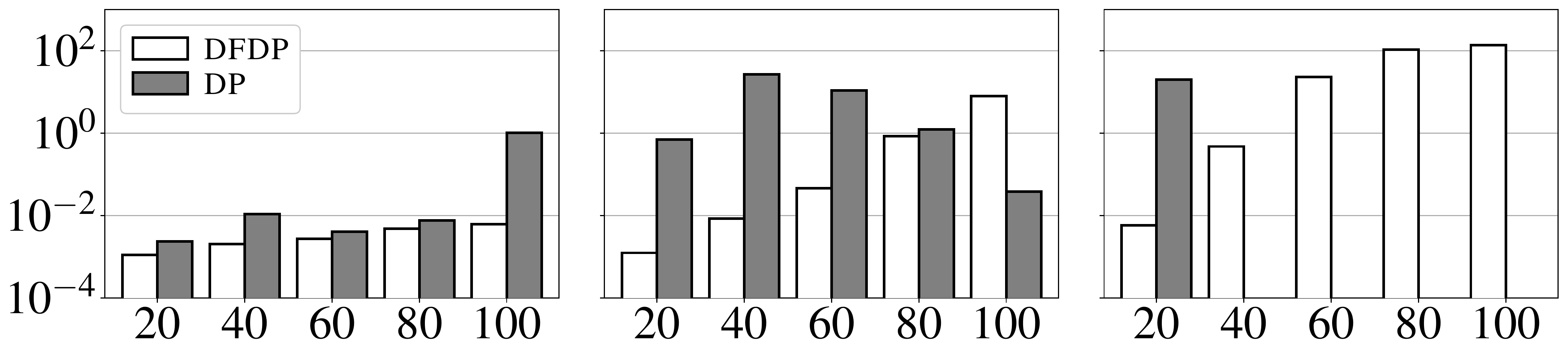}
    \put(0.5,-22.5){ \includegraphics[width=0.6\columnwidth]{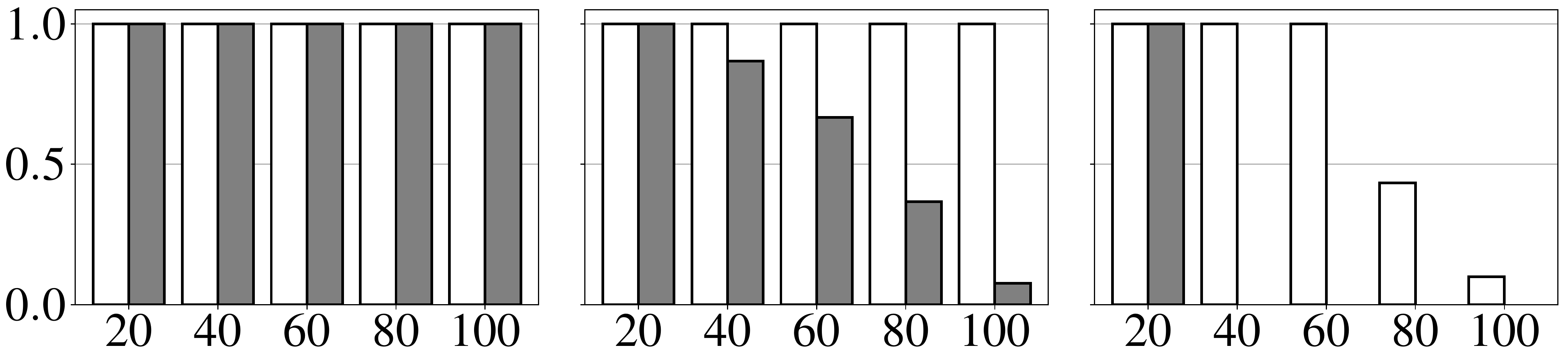}}
    \end{overpic}
    \vspace{20mm}
    \caption{Performance of \DFSDP and DP over \lrbm. The top row shows the average
    computation time (s) and the bottom row the success rate, for density levels
    $\rho=0.2, 0.3$, $0.4$, from left to right. The $x$-axis denotes the number of 
    objects involved in a test case.}
    \label{fig:LabeledAlgorithms}
\end{figure}

In Fig.~\ref{fig:LabeledAlgorithms}, we compare the effectiveness of DP and \DFSDP, in terms of computation time and success rate, for different densities. 
Each data point is the average of $30$ test cases minus the unfinished ones, if any, 
subject to a time limit of $300$ seconds per test case. For \lrbm, we are able to 
push to $\rho = 0.4$, which is fairly dense. 
The results clearly demonstrate that \DFSDP significantly outperforms the baseline
DP.
Based on the evaluation, both methods can be used to tackle practical-sized problems 
(e.g., tens of objects), with \DFSDP demonstrating superior efficiency and robustness. 

\begin{figure}[h!]
    \centering
\begin{overpic}
[width=0.33\textwidth]{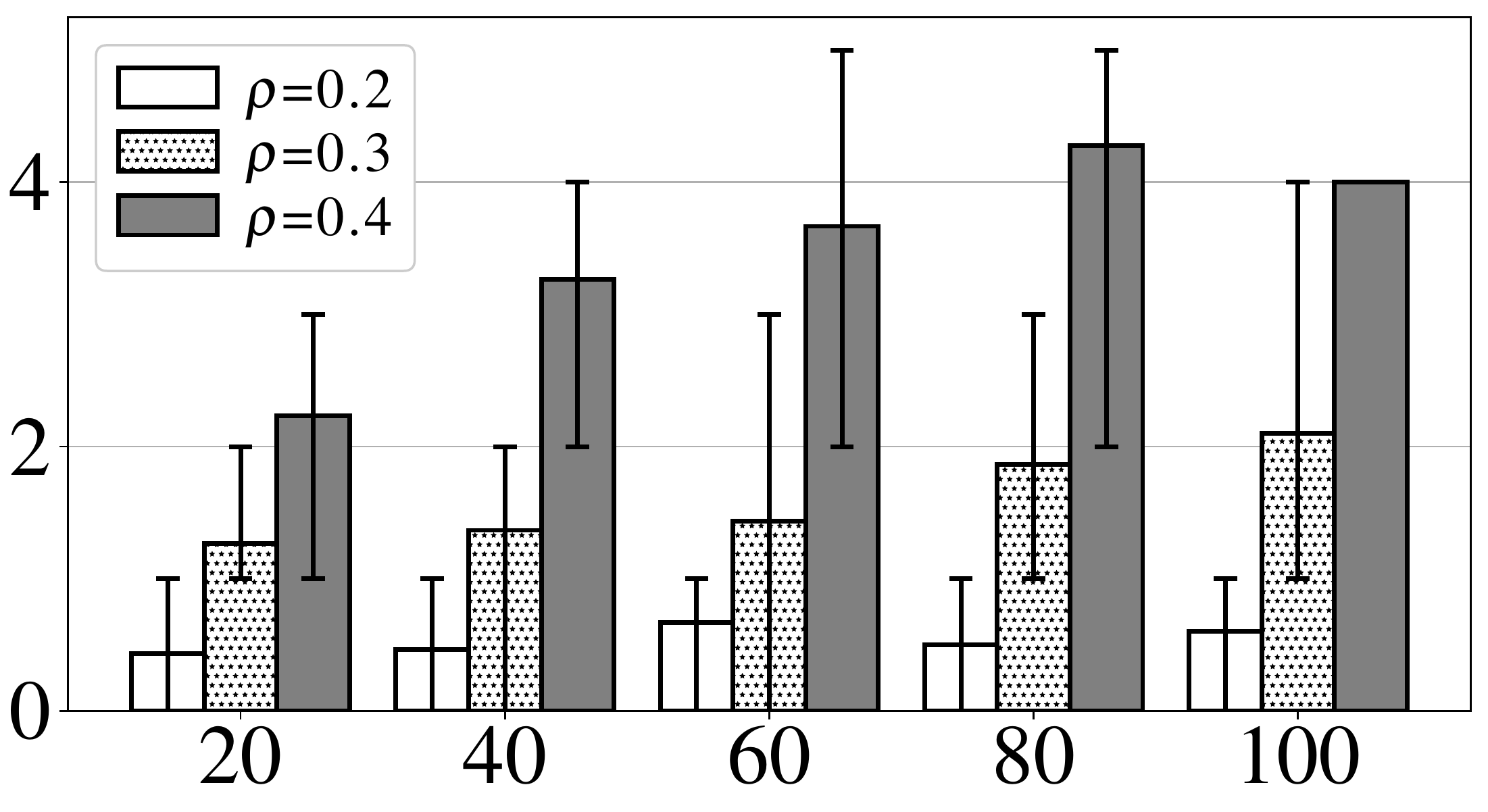}
\end{overpic}
\begin{overpic}
[width=0.33\textwidth]{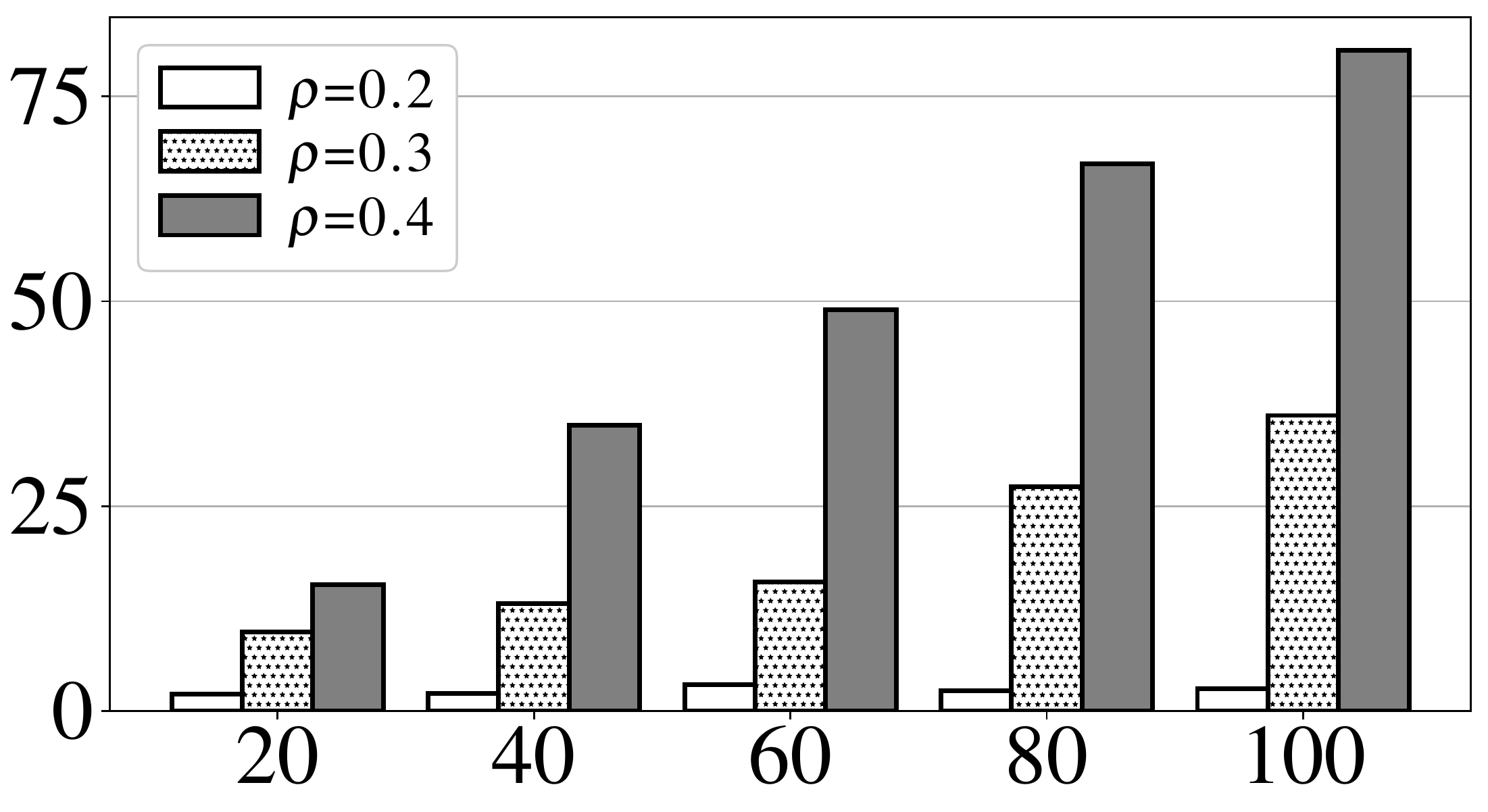}
\end{overpic}
    \caption{For \lrbm instances with $\rho = 0.2$-$0.4$ and $n=20$-$100$, the 
    left figure shows average \mrb size and range. The right figure shows
    the size of the largest connected component of the dependency graph.}
    \label{fig:LabeledResults}
\end{figure}

The actual \mrb sizes for the same test cases from Fig.~\ref{fig:LabeledAlgorithms} 
are shown in Fig.~\ref{fig:LabeledResults}
on the left. We observe that \mrb is rarely very large even for fairly large \lrbm
instances. The size of \mrb appears correlated to the size of the largest connected
component of the underlying dependency graph, shown in Fig.~\ref{fig:LabeledResults}
on the right.
% \jy{We should probably get rid of the range for the second 
% figure in Fig.~\ref{fig:LabeledResults}; otherwise we probably 
% need to add to all other figures...}

\begin{figure}[h!]
    \centering
    \begin{overpic}[width=0.53\textwidth]{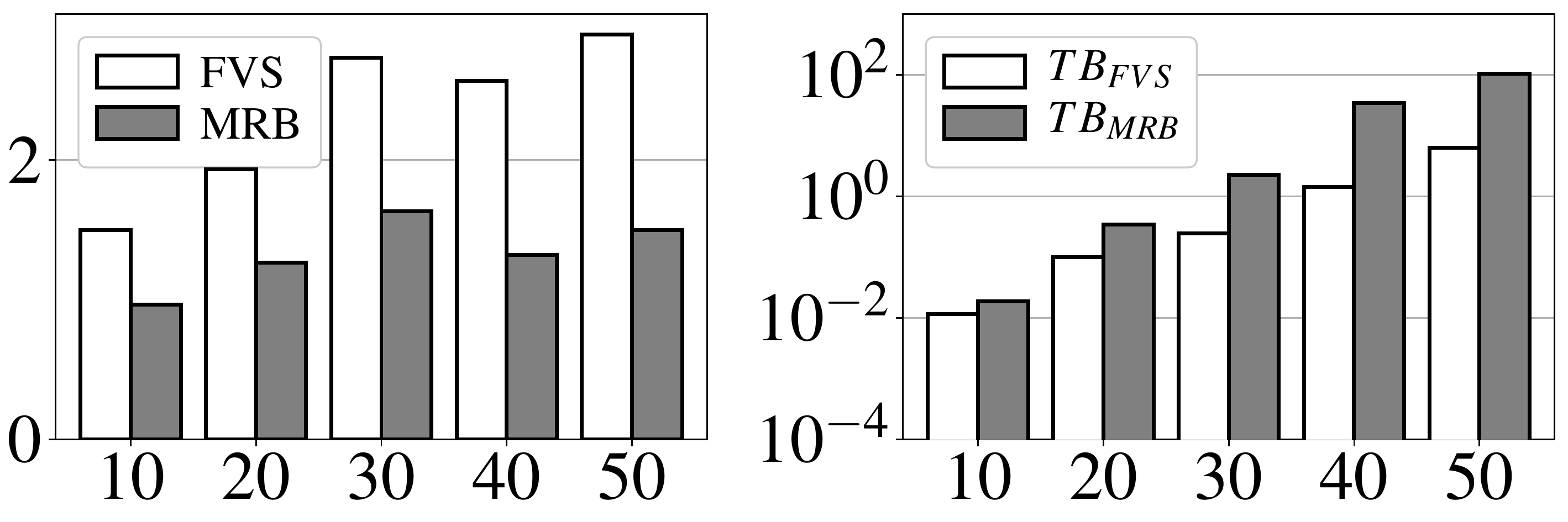}
    \put(22,-3.5){{\small (a)}}
    \put(76,-3.5){{\small (b)}}
    \end{overpic}
    \vspace{2mm}
    \caption{(a) Comparison between size of \mrb and \fvs. (b)  Computation time 
    comparison between $\ilpfvs$ and $\ilptb$. }
    \label{fig:MRBFVS}
    \vspace{-1mm}
\end{figure}

For $\lrbm$ with $\rho = 0.3$ and $n$ up to $50$, we computed the \fvs sizes 
using $\ilpfvs$ (which does not scale to higher $\rho$ and $n$) and compared that 
with the \mrb sizes, as shown in Fig.~\ref{fig:MRBFVS} (a). We observe that the
\fvs is about twice as large as \mrb, suggesting that \mrb provides more reliable 
information for estimating the design parameters of pick-n-place systems. For 
these instances, we also computed the total number of buffers needed subject to 
the \mrb constraint using $\ilptb$. Out of about $150$ instances, only $1$ showed 
a difference as compared with \fvs (therefore, this information is not shown in 
the figure). In Fig.~\ref{fig:MRBFVS} (b), we provided computation time 
comparison between $\ilpfvs$ and $\ilptb$, showing that $\ilptb$ is practical,
if it is desirable to minimize the total buffers after guaranteeing the 
minimum number of running buffers. 

Considering our theoretical findings and the evaluation results, an important 
conclusion can be drawn here is that \mrb is effectively a small constant for
random instances, even when the instances are very large. Also, 
minimizing the total number of buffers used subject to \mrb constraint can 
be done quickly for practical-sized problems. 

\subsection{\urbm over Random Instances}
\begin{figure}[h!]
    \centering
    \begin{overpic}[width=0.6\columnwidth]{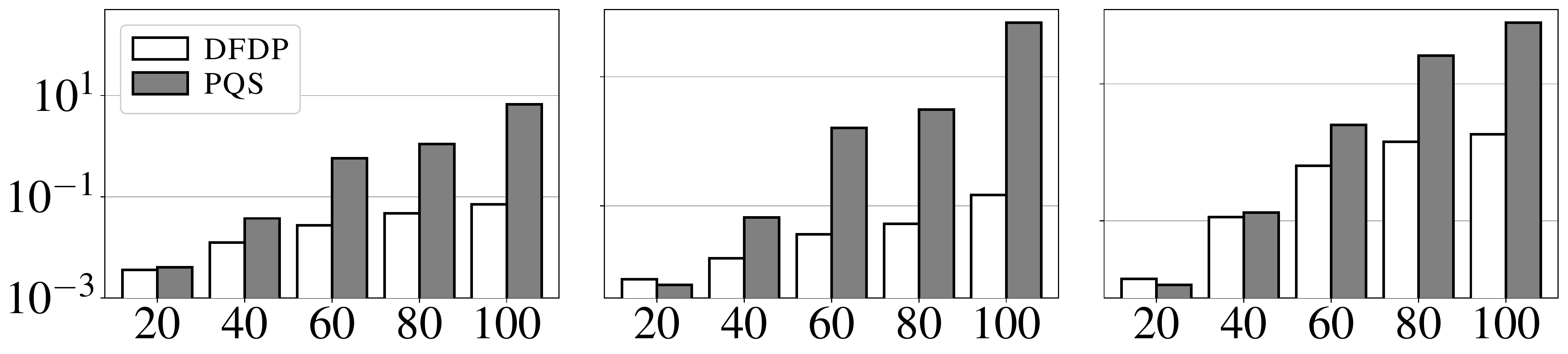}
    \put(0.5,-22.5){ \includegraphics[width=0.6\columnwidth]{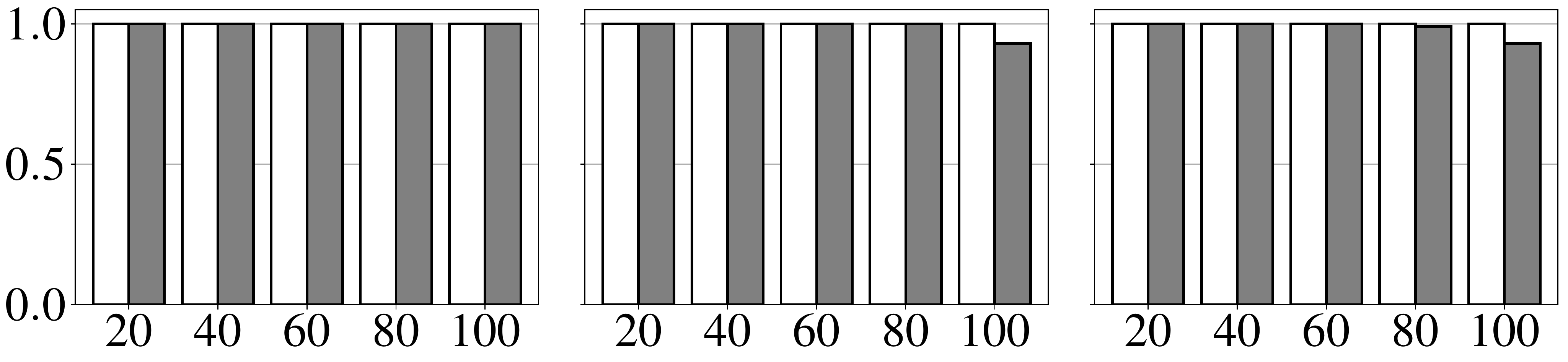}}
    \end{overpic}
        \vspace{20mm}
\caption{Performance of \DFSDP and \PQS over \urbm. The top row shows the average
    computation time and the bottom row shows the success rate, for density levels
    $\rho=0.4, 0.5, 0.6$, from left to right.}
    \label{fig:UnlabeledAlgorithms}
\end{figure}

\begin{figure}[h!]
    \centering
\includegraphics[width=0.4\textwidth]{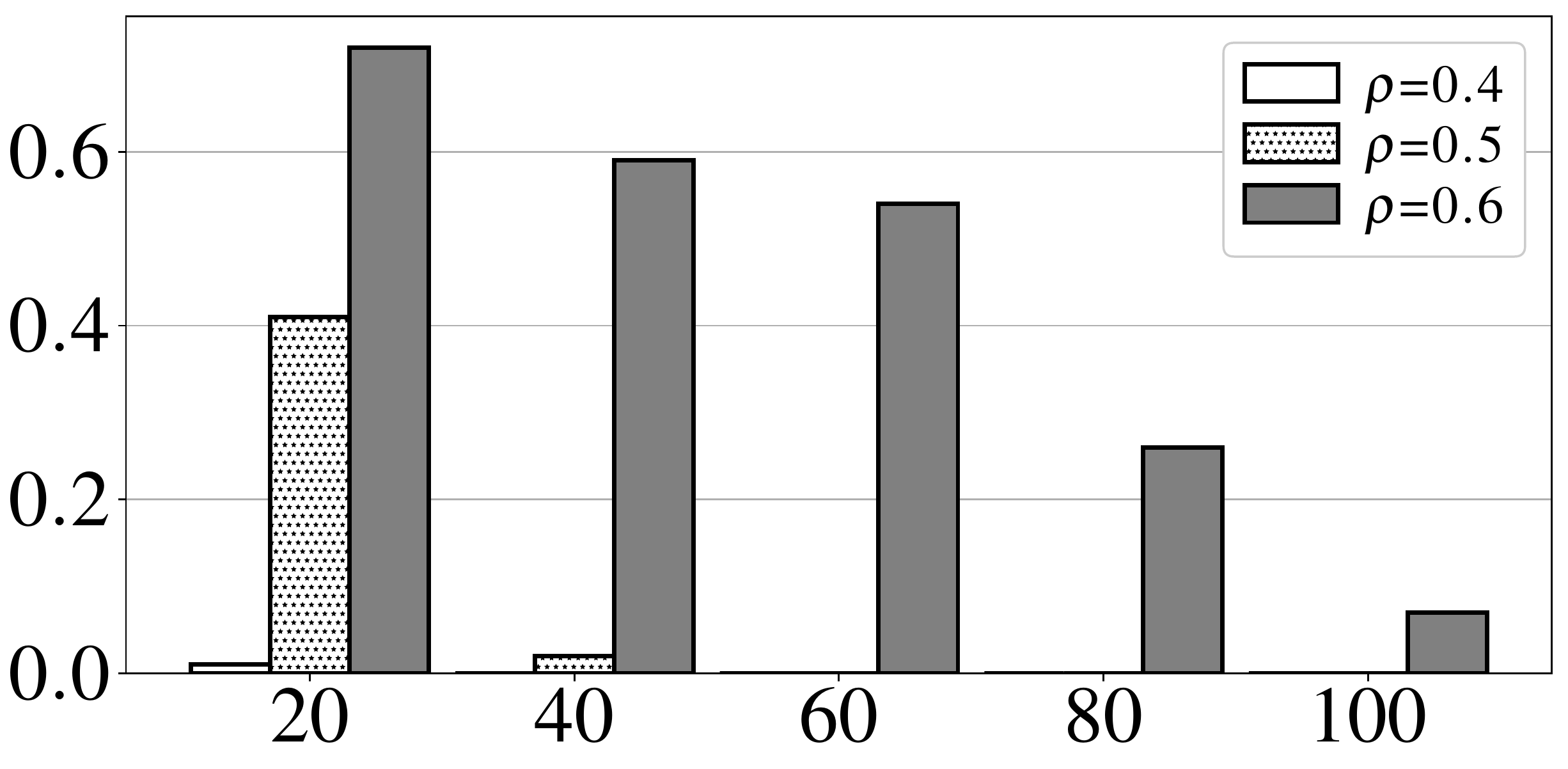}
    \caption{Average \mrb size for \urbm instances with $\rho=0.4-0.6$ and $n=20-100$.
    For $\rho = 0.4$ and $0.5$, the $\mrb$ sizes are near zero as the number of 
    objects goes beyond $20$.}
    \label{fig:UnlabeledResults}
    \vspace{-1mm}
\end{figure}

For \urbm, we carry out a similar performance evaluation as we have done for \lrbm. 
Here, \PQS and \DFSDP are compared. For each combination of $\rho$ and $n$, $100$ 
random test cases are evaluated. Notably, we can reach $\rho = 0.6$ with relative
ease. From Fig.~\ref{fig:UnlabeledAlgorithms}, we observe that \DFSDP is more 
efficient than \PQS, especially for large-scale dense settings. In terms of the \mrb 
size, all instances tested have an average \mrb size between $0$ and $0.7$, which is 
fairly small (Fig.~\ref{fig:UnlabeledResults}). Interestingly, we witness a decrease of 
\mrb as the number of objects increases, which could be due to the lessening 
``border effect'' of the larger instances. That is, for instances with fewer 
objects, the bounding square puts more restriction on the placement of the 
objects inside. For larger instances, such restricting effects become smaller.
We mention that the total number of buffers for random \urbm cases subject to 
\mrb constraints are generally very small. 
%The right figure in Fig.~\ref{fig:UnlabeledResults} shows the 
%number of running buffers we can get "for free" before we do the first branching.

%\begin{comment}
%\jy{Make Fig.~\ref{fig:UnlabeledResults} shorter, similar to Fig. 14.}
%\end{comment}

\subsection{Manually Constructed Difficult Cases of \toroe}
\begin{figure}[h!]
    \centering
    \begin{overpic}[width=0.6\textwidth]{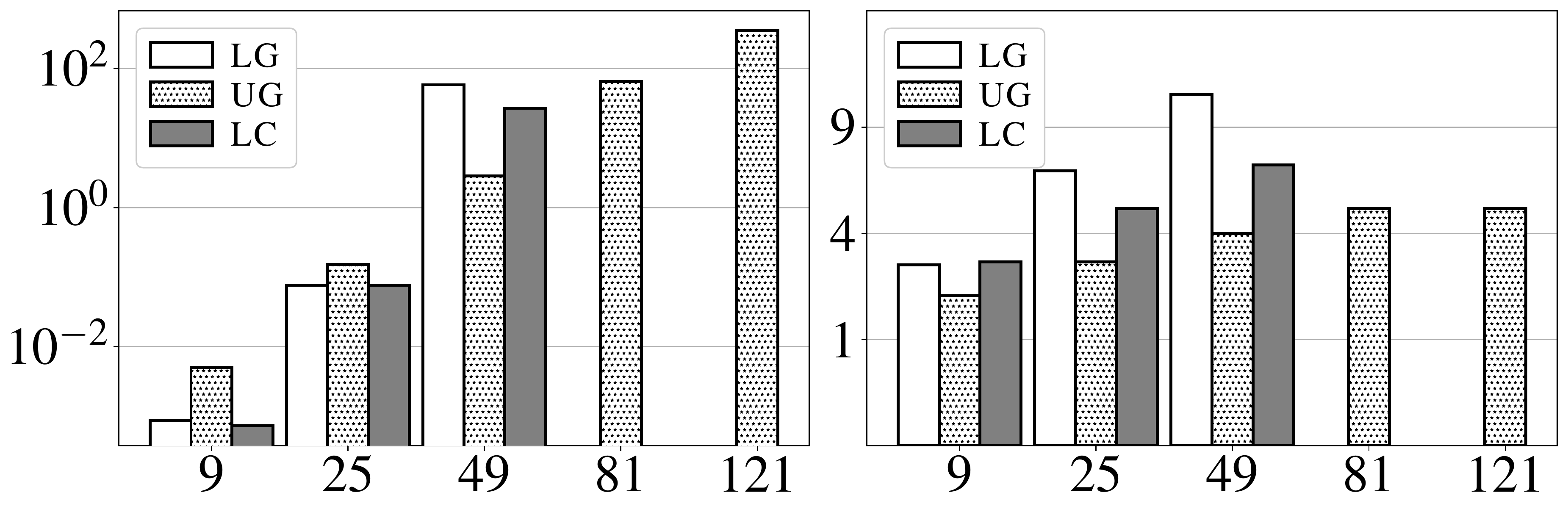}
    \end{overpic}
    %\vspace{-2mm}
    \caption{ For handcrafted cases and different numbers of objects, the left 
    figure shows the computation time by \DFSDP and the right figure the resulting
    \mrb size.}
    \label{fig:SpecialResults}
\end{figure}

In the random scenario, the running buffer size is limited. In particular, for 
\lrbm, the dependency graph tends to consist of multiple strongly connected 
components that can be dealt with independently. We further show the performance 
of \DFSDP on the instances with $\mrb=\Theta(\sqrt{n})$. We evaluate three kinds 
of instances: (1) \textsc{UG}: $m^2$-object \urbm instances whose $\udg$ are 
dependency grid $\mathcal D(m,2m)$ (e.g., Fig.~\ref{fig:DependencyGrid}); (2) 
\textsc{LG}: $m^2$-object \lrbm instances whose start and goal arrangements are 
the same as the instances in (1). (3) \textsc{LC}: $m^2$-object \lrbm instances 
with objects placed on a cycle (Fig.~\ref{fig:lrbm-cycle}). The computation time and 
the corresponding \mrb are shown in Fig.~\ref{fig:SpecialResults}. For 
\textsc{LG} instances, the labels are randomly assigned. We try 30 test cases and 
then plot out the average. We observe that the \mrb is much larger for these 
handcrafted instances as compared with random instances with similar density and
number of objects.

\subsection{Evaluation on \toroi Instances}
% We also implemented the algorithms of the \trlb framework in Python. 

\subsubsection{Ablation Study for Cylindrical Objects.}
\begin{figure}[h!]
    \vspace{2mm}
\centering
    \begin{overpic}[width=0.6\columnwidth]{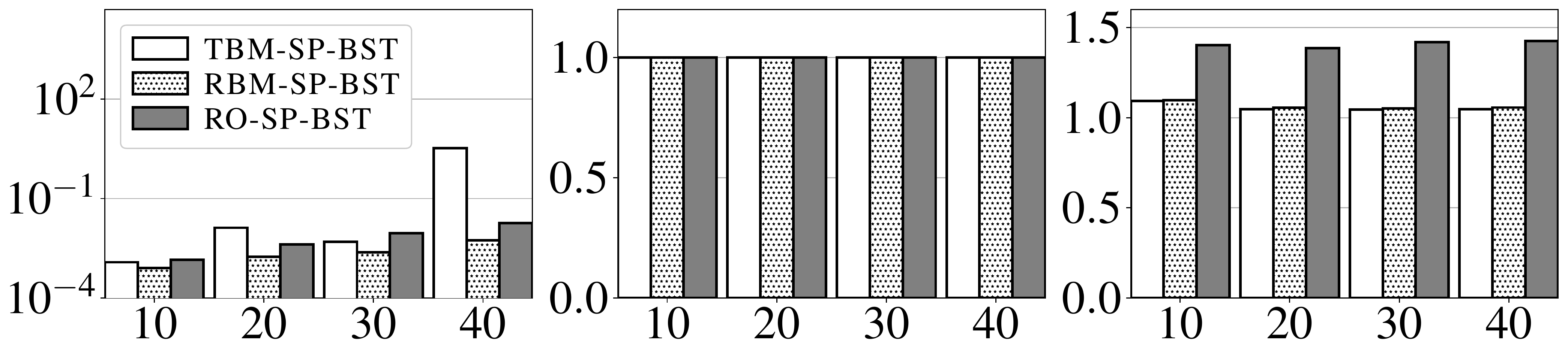}
    \put(-1.5,-22.5){ \includegraphics[width=0.6\columnwidth]{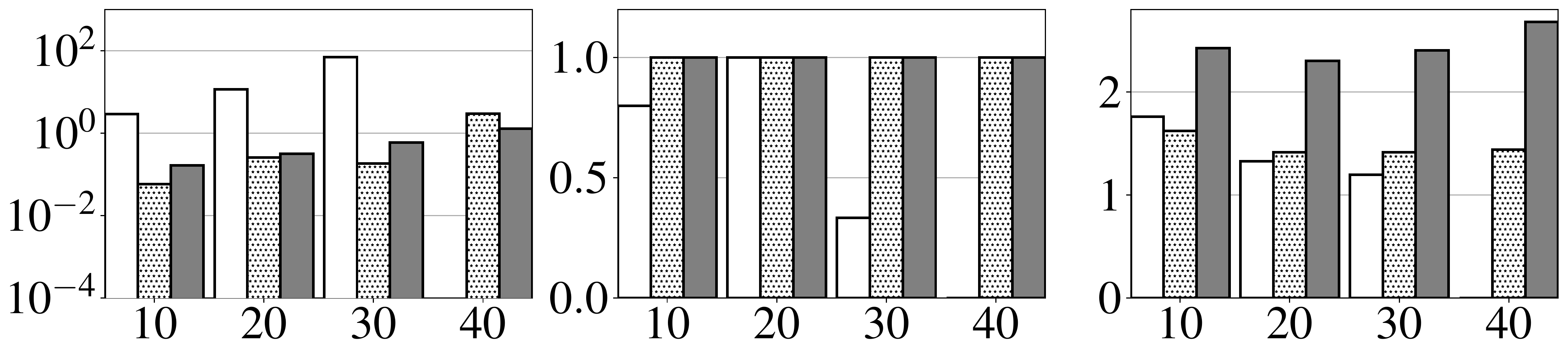}}
    \end{overpic}
    \vspace{20mm}
    \caption{Comparison of primitive planners with $10$-$40$ cylinders and density levels $\rho=0.3$ (top), 0.5 (bottom) (left: computation time in seconds; middle: success rate; right: number of actions as multiples of $|\mathcal O|$).}
    \label{fig:Primitive}
\end{figure}

For evaluation, we present experiments with cylindrical objects to compare lazy buffer generation algorithms. 
A more detailed version of the ablation study is presented in \cite{gao2021fast}.
We first compare the primitive plan computation options, 
using sampling-based buffer allocation (SP) and bidirectional tree search (BST). 
% Primitive planner TBM computes plans minimizing the number of total buffers using a dynamic programming-based algorithm.
% When computing primitive plans, minimizing the total buffer size is equivalent to minimizing the number of total actions.
% Another primitive planner RBM computes plans minimizing the number of running buffers using DFDP.
% We also compare a baseline primitive planner RO, which returns a random ordering of objects.
The results are shown in Fig.~\ref{fig:Primitive}. 
Even though plans generated by TBM-SP-BST are slightly shorter than RBM-SP-BST, TBM is less scalable as either the density level or the number of objects in the workspace increases. 
Compared to RBM plans, individual RO plans can be generated almost instantaneously but we don't see much benefit in computation time for the overall algorithm. 
The results indicate that minimizing \mrb in primitive plan computation is beneficial as it results in efficient and high-quality \toroi solutions.

\begin{figure}[h!]
    \vspace{1mm}
\centering
    \begin{overpic}[width=0.6\columnwidth]{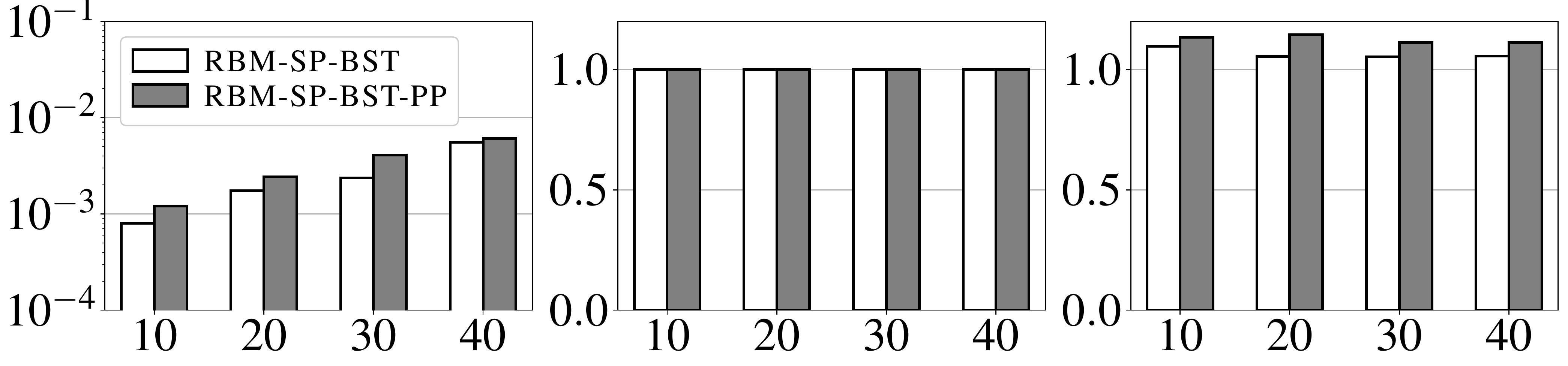}
    \put(-1.5,-22.5){ \includegraphics[width=0.6\columnwidth]{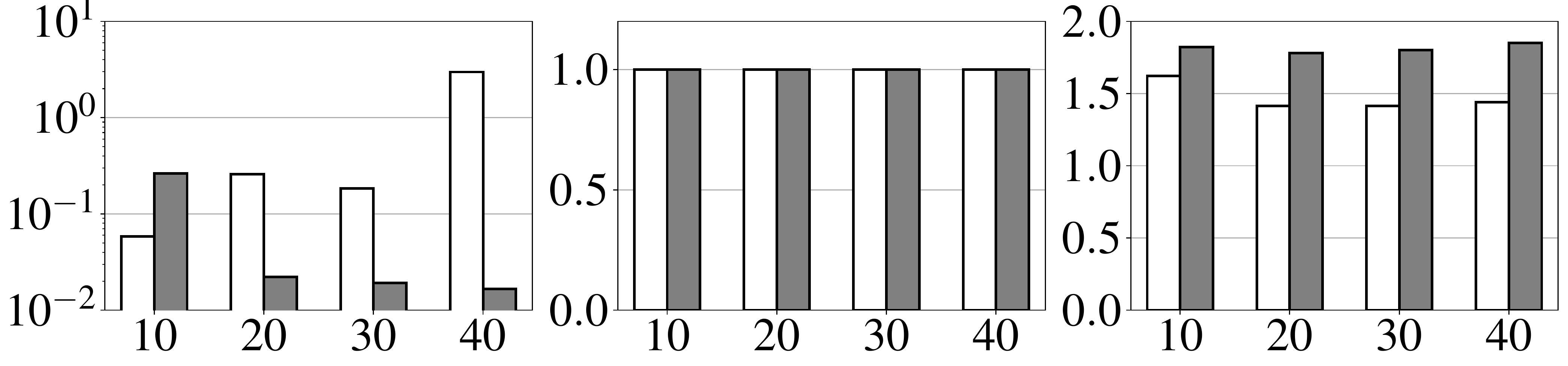}}
    \end{overpic}
    \vspace{20mm}
    \caption{Comparison between lazy buffer allocation algorithms with and without preprocessing. 
    There are 10-40 cylinders in the workspace at density levels $\rho=0.3$ (top), $0.5$ (bottom) (left: computation time in seconds; middle: success rate; right: number of actions as multiples of $|\mathcal O|$).}
    \label{fig:PP}
\end{figure}

We also integrate the \ urbm-based preprocessing routine into the bi-directional search tree framework (RBM-SP-BST-PP) and compare it with RBM-SP-BST.
The results (Fig.~\ref{fig:PP}) suggest that preprocessing is effective in increasing the success rate in dense environments.
In addition, preprocessing significantly speeds up computation in large-scale dense cases at the price of extra actions to execute preprocessing. 
By simplifying the dependency graph with preprocessing, 
less time is needed to compute a primitive plan.

\subsubsection{Comparison with Alternatives for Cylindrical Objects.}
\begin{figure}[h!]
    \vspace{2mm}
\centering
    \begin{overpic}[width=0.6\columnwidth]{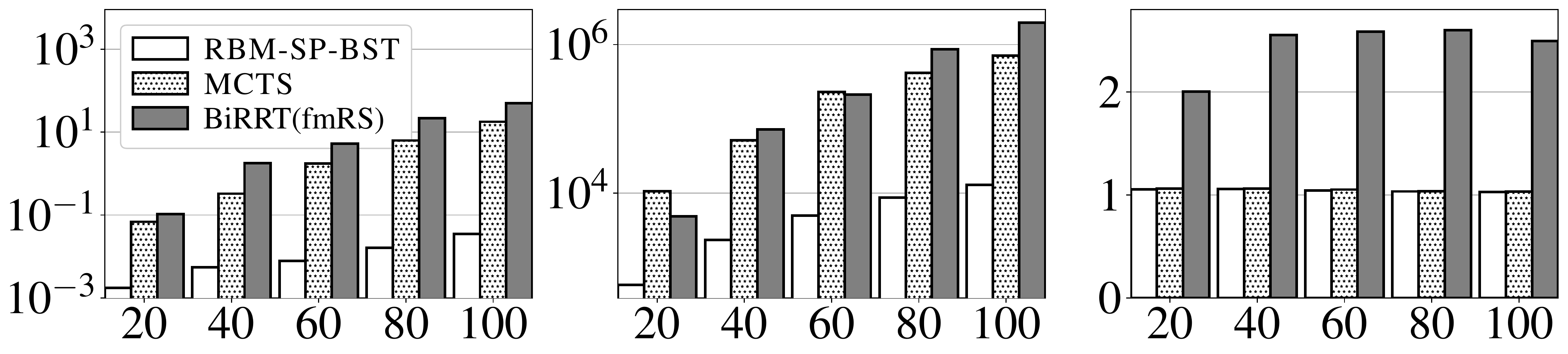}
    \end{overpic}
    \vspace{-.1in}
    \caption{Comparison of algorithms with 20-100 cylinders at density level $\rho=0.3$ (left: computation time in seconds; middle: number of collision checks; right: number of actions as multiples of $|\mathcal O|$).}
    \label{fig:LargeScale}
\end{figure}

We compare the proposed method RBM-SP-BST with BiRRT(fmRS) \cite{krontiris2016efficiently} and an MCTS planner \cite{labbe2020monte}, which, to the best of our knowledge, are state-of-the-art planners for \toroi. The MCTS planner is a C++ solver, while the other two methods are implemented in Python. Besides success rate, solution quality, and computation time, 
we also compare the number of collision checks which are time-consuming in most planning tasks.  In Fig.~\ref{fig:LargeScale}, we compare the methods in large-scale problems with $\rho=0.3$. The success rate is $100\%$ for all.
Our method, RBM-SP-BST, avoids repeated collision checks due to the use of the dependency graph. BiRRT(fmRS), which only uses dependency graphs locally, spends a lot of time and conducts a lot of collision checks to generate random arrangements. MCTS generates solutions with similar optimality but does so also with a lot of collision checking, which slows down the computation.  We note that a value of $1$ in the right figure (number of actions) is the minimum possible, so both RBM-SP-BST and MCTS compute high-quality solutions, while RBM-SP-BST does slightly better. To sum up, in sparse large-scale instances, RBM-SP-BST is two magnitudes faster and conducts much fewer collision checks than the alternatives.

\begin{figure}[h!]
    \vspace{2mm}
\centering
    \includegraphics[width=0.6\columnwidth]{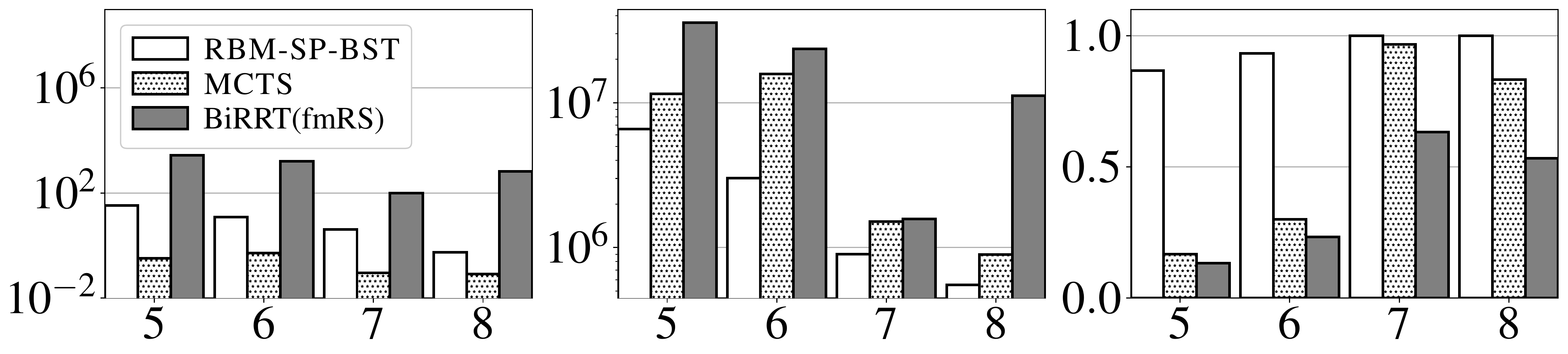}
    \caption{Comparison of methods on ``dense-small'' instances where 5-8 objects are packed in an environment with $\rho=0.5$ (left: computation time in seconds; middle: number of collisions; right: success rate).}
    \label{fig:DenseSmallComaprison}
\end{figure}

\begin{figure}[h!]
    \centering
    \includegraphics[width=0.32\textwidth]{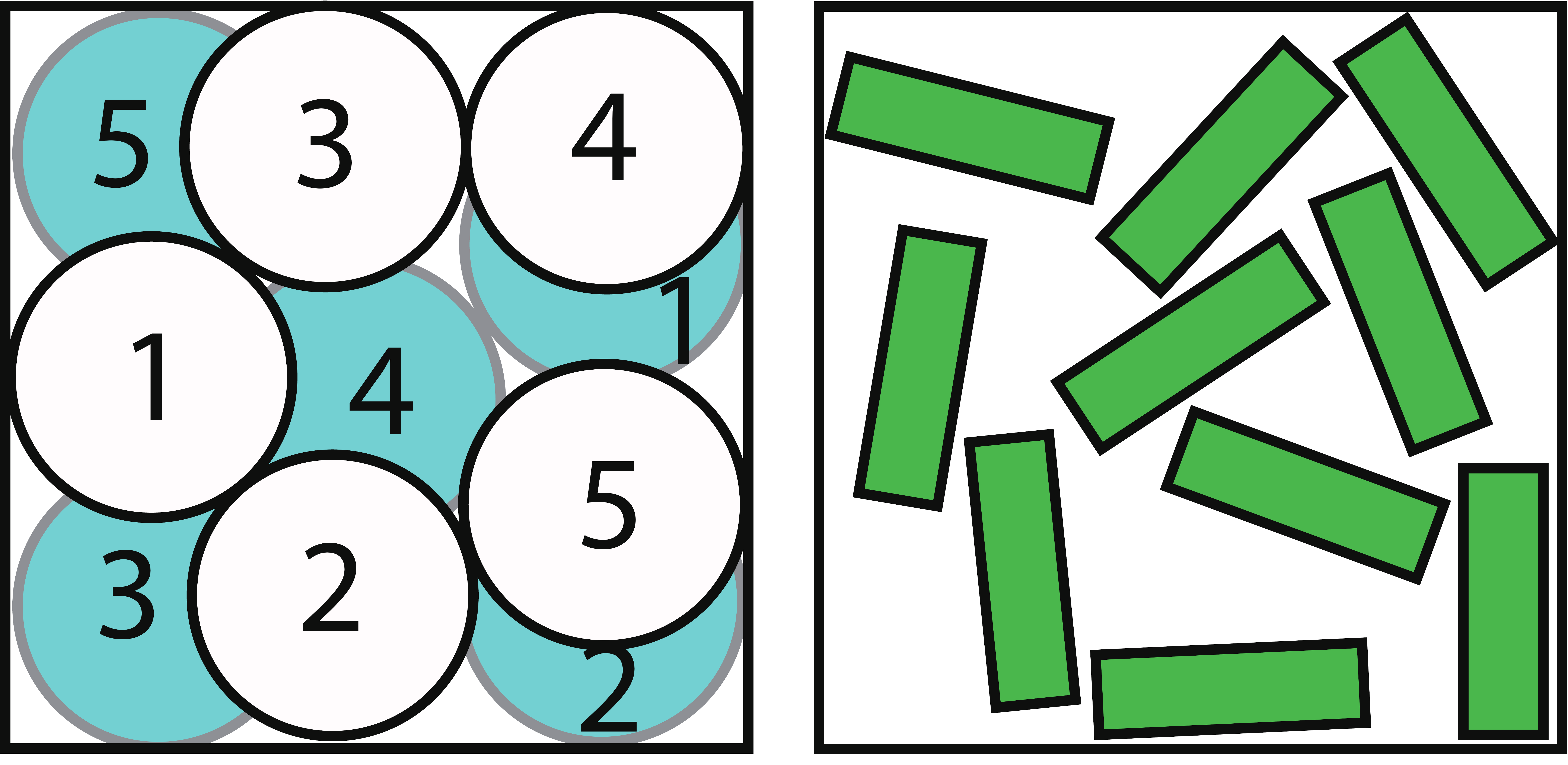}
    \caption{[Left] ``dense-small'' instances where 5-8 cylinders packed in the environment and density level $\rho=0.5$. [Right] 10 cuboids with $\rho=0.4$.}
    \label{fig:dense-n-stick}
\end{figure}

Next, in Fig.~\ref{fig:DenseSmallComaprison}, 
we compare the methods in ``dense-small'' instances, where a few objects are packed densely (Fig.~\ref{fig:dense-n-stick}[Left]). Here, RBM-SP-BST is the only method that maintains a high success rate in these difficult cases.

\begin{figure}[h!]
    \vspace{2mm}
\centering
    \includegraphics[width=0.35\textwidth]{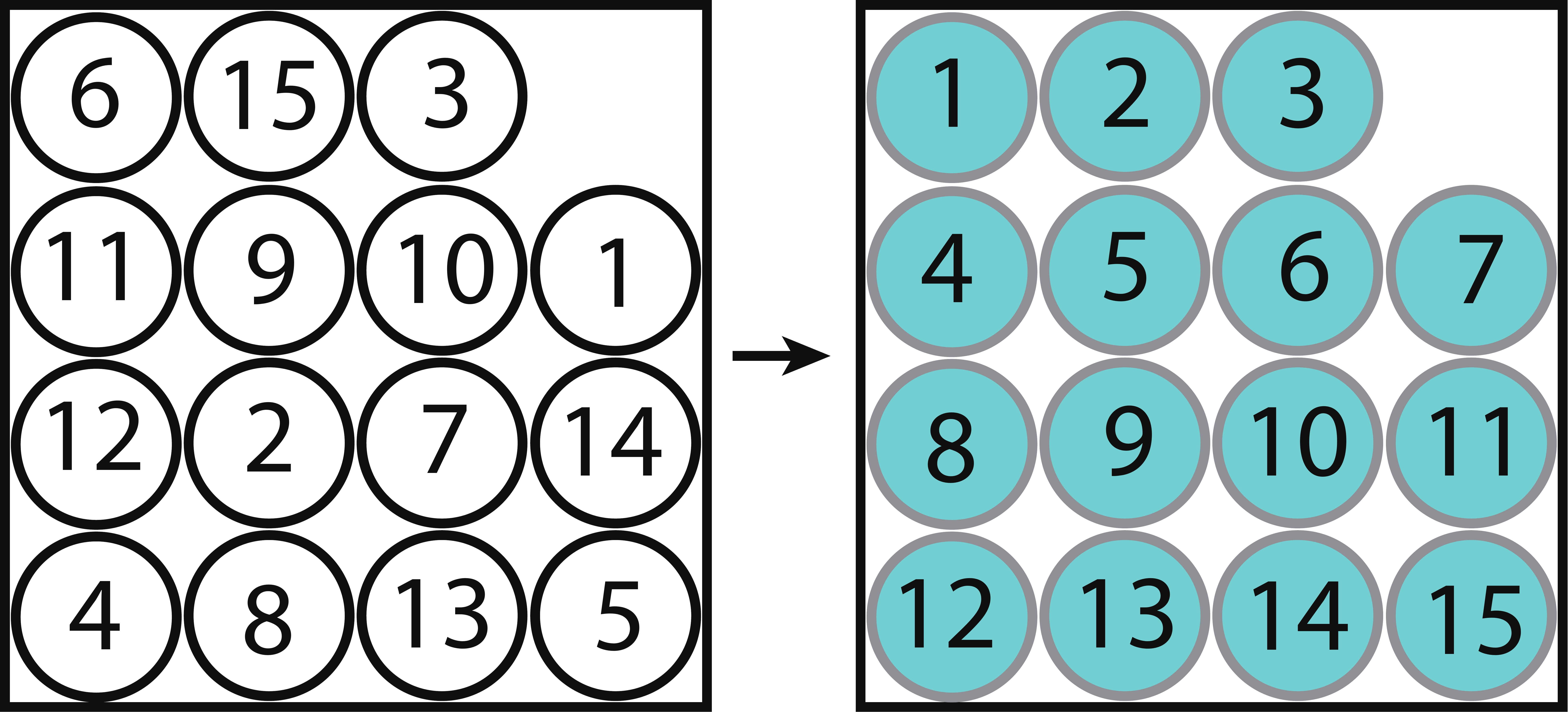}\hspace{2mm}
    \includegraphics[width=0.2\columnwidth]{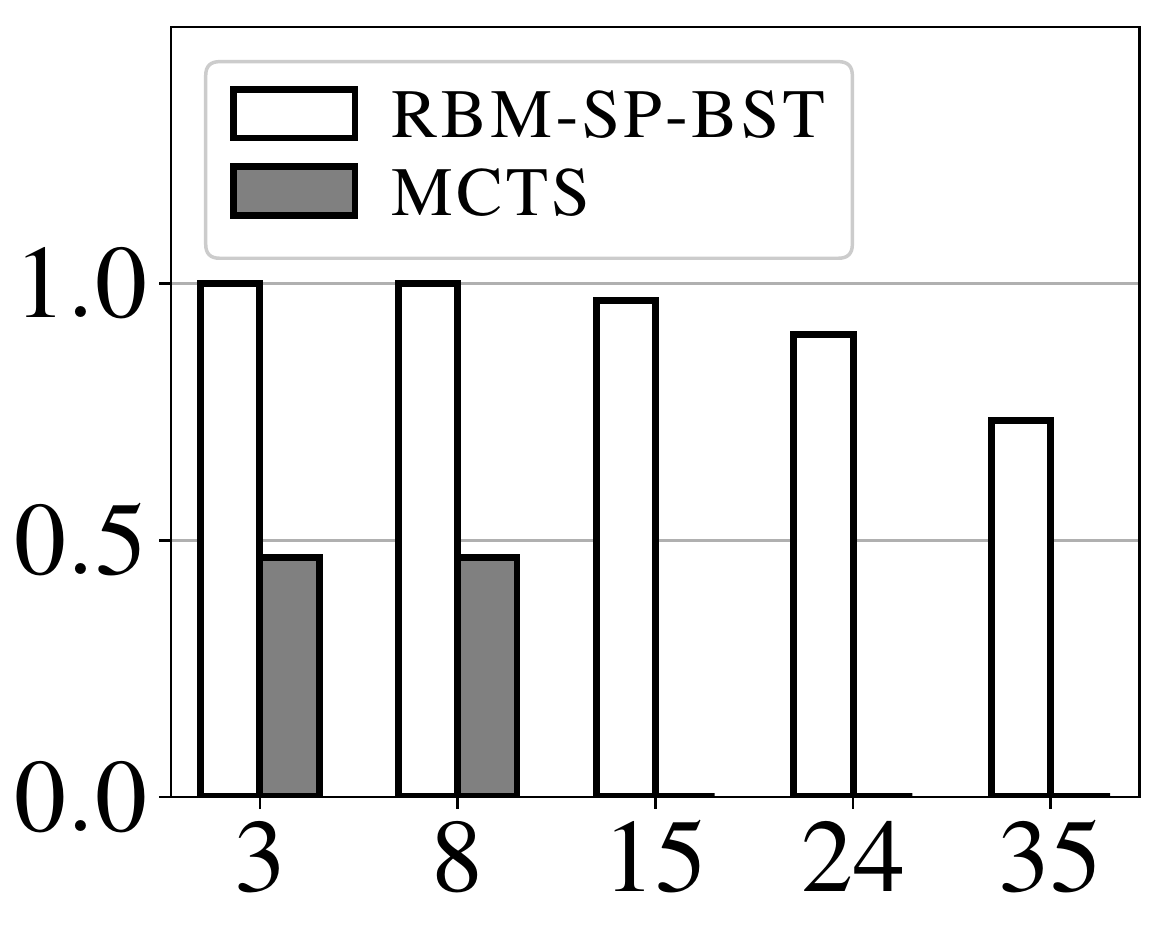}
    \caption{Comparison among methods in lattice instances with 3-35 objects. [left] lattice example; [right] success rate}
    \label{fig:lattice}
\end{figure}

We further compare the performance of RBM-SP-BST and MCTS in lattice rearrangement problems, which are recently studied in the literature \cite{yurearrangement}. 
An example with 15 objects is shown in Fig.~\ref{fig:lattice}[left]. 
In the start and goal arrangements, 
gaps between adjacent objects are set to be 0.01 object radius,
and thus buffer allocation is challenging for sampling-based methods. 
While MCTS tries all the actions on each node, RBM-SP-BST is able to detect the embedded combinatorial object relationship via the dependency graph and therefore needs fewer buffer allocation calls. As shown in Fig.~\ref{fig:lattice}[right], RBM-SP-BST has a much higher success rate in lattice rearrangement tasks.

\subsubsection{Cuboid Objects}
Because the MCTS solver only supports cylindrical objects, we only compare RBM-SP-BST and BiRRT(fmRS) in the cuboid setup (Fig.~\ref{fig:dense-n-stick}[right]).
When $\rho=0.3$, RBM-SP-BST computes high-quality solutions efficiently, 
while BiRRT(fmRS) can only solve instances with up to 20 cuboids.
We mention that, when $\rho=0.4$, BiRRT(fmRS) cannot solve any instance, 
but RBM-SP-BST can solve 50-object rearrangement problems in 28.6 secs on average.

\begin{figure}[h!]
    \vspace{2mm}
\centering
    \includegraphics[width=0.6\columnwidth]{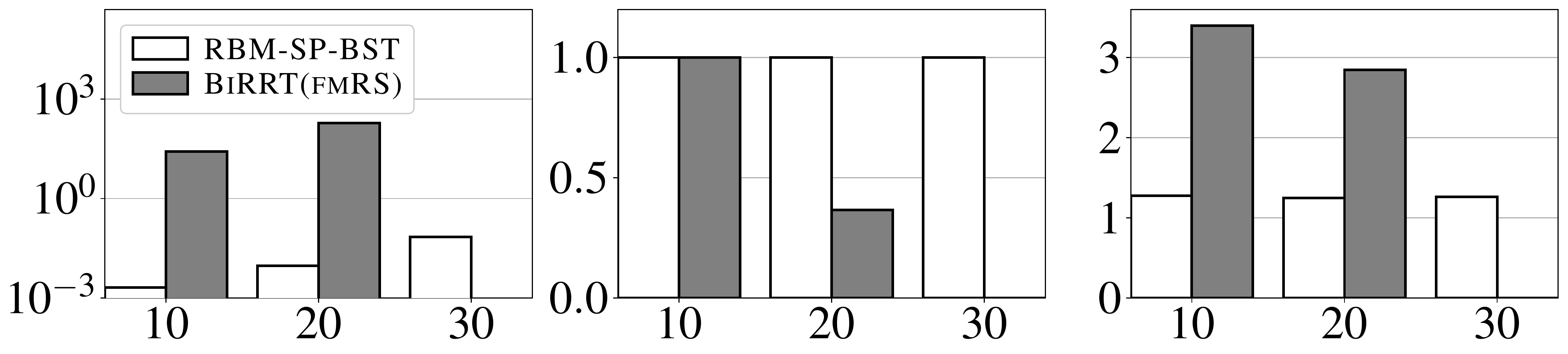}
    \caption{Comparison between methods in cuboid instances with $10$-$30$ cuboids and $\rho=0.3$ (left: computation time in seconds; middle: success rate; right: number of actions as multiples of $|\mathcal O|$).}
    \label{fig:StickComaprison}
\end{figure}

\section{Physical Experiments}\label{sec:hardware}
In this section, we demonstrate that the plans computed by proposed algorithms can be readily executed on real robots in a complete vision-planning-control pipeline.
We first introduce the hardware setup and the pipeline during the execution of the rearrangement plans.
After that, we present experimental results based on the execution of computed rearrangement plans.

\subsection{Hardware Setup}\label{subsec:setup}
\begin{figure}[ht!]
    \centering
    \includegraphics[width=0.3\textwidth]{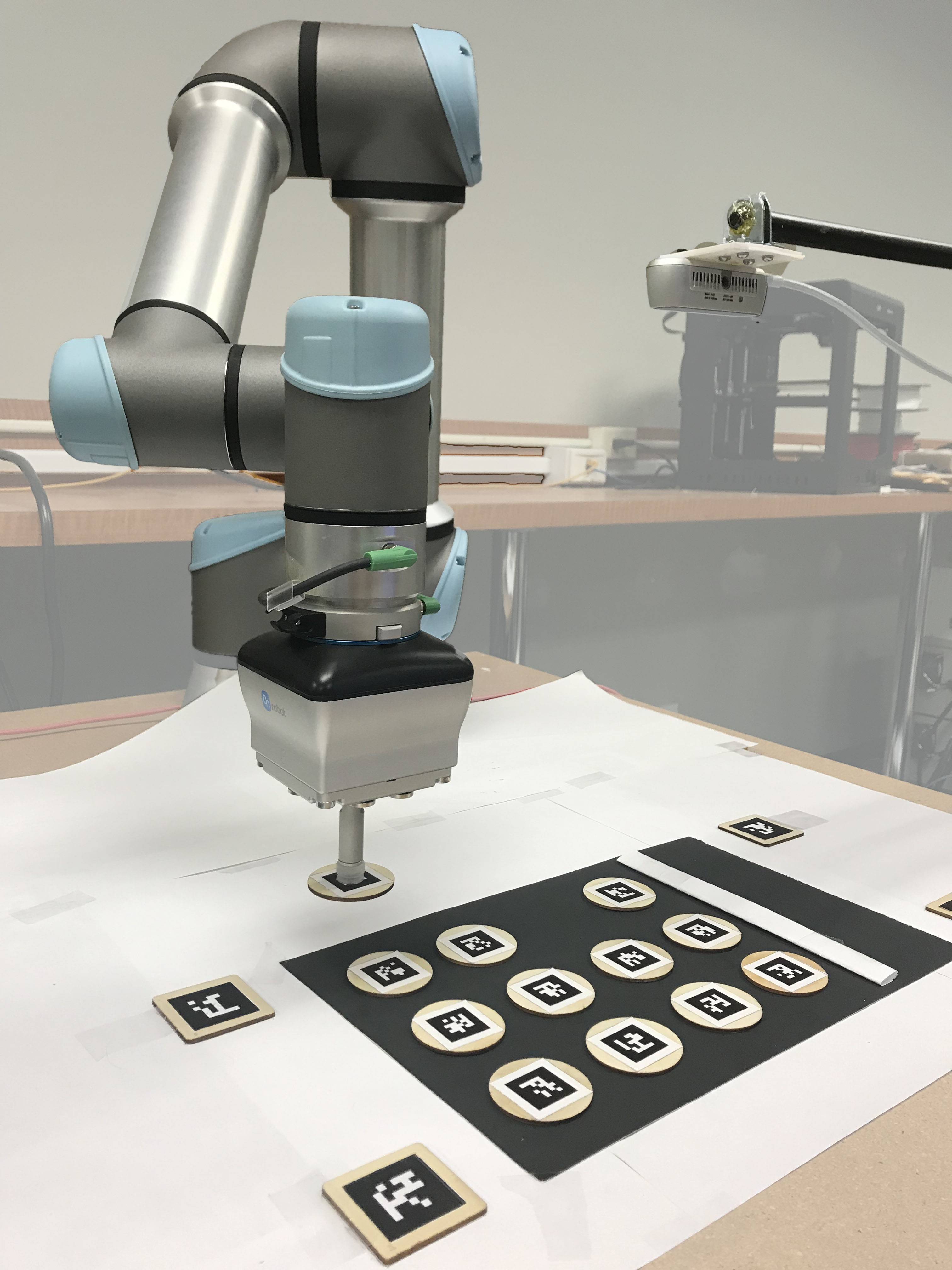}
    \caption{Our hardware setup for executing rearrangement plans computed by proposed algorithms.}
    \label{fig:workspace}
\end{figure}

\begin{figure}[ht!]
    \centering
    \includegraphics[width=0.3\textwidth]{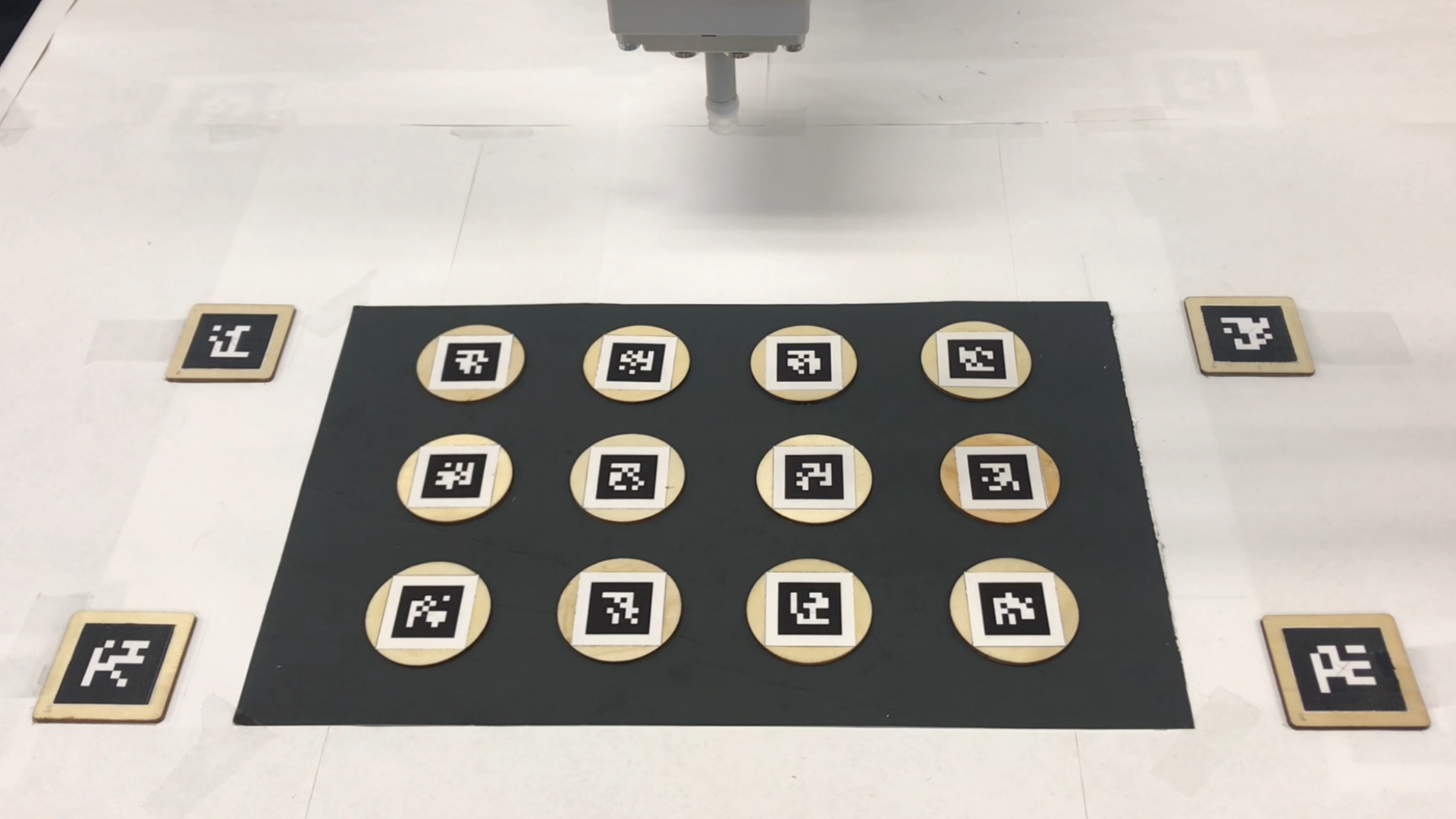}
    \includegraphics[width=0.3\textwidth]{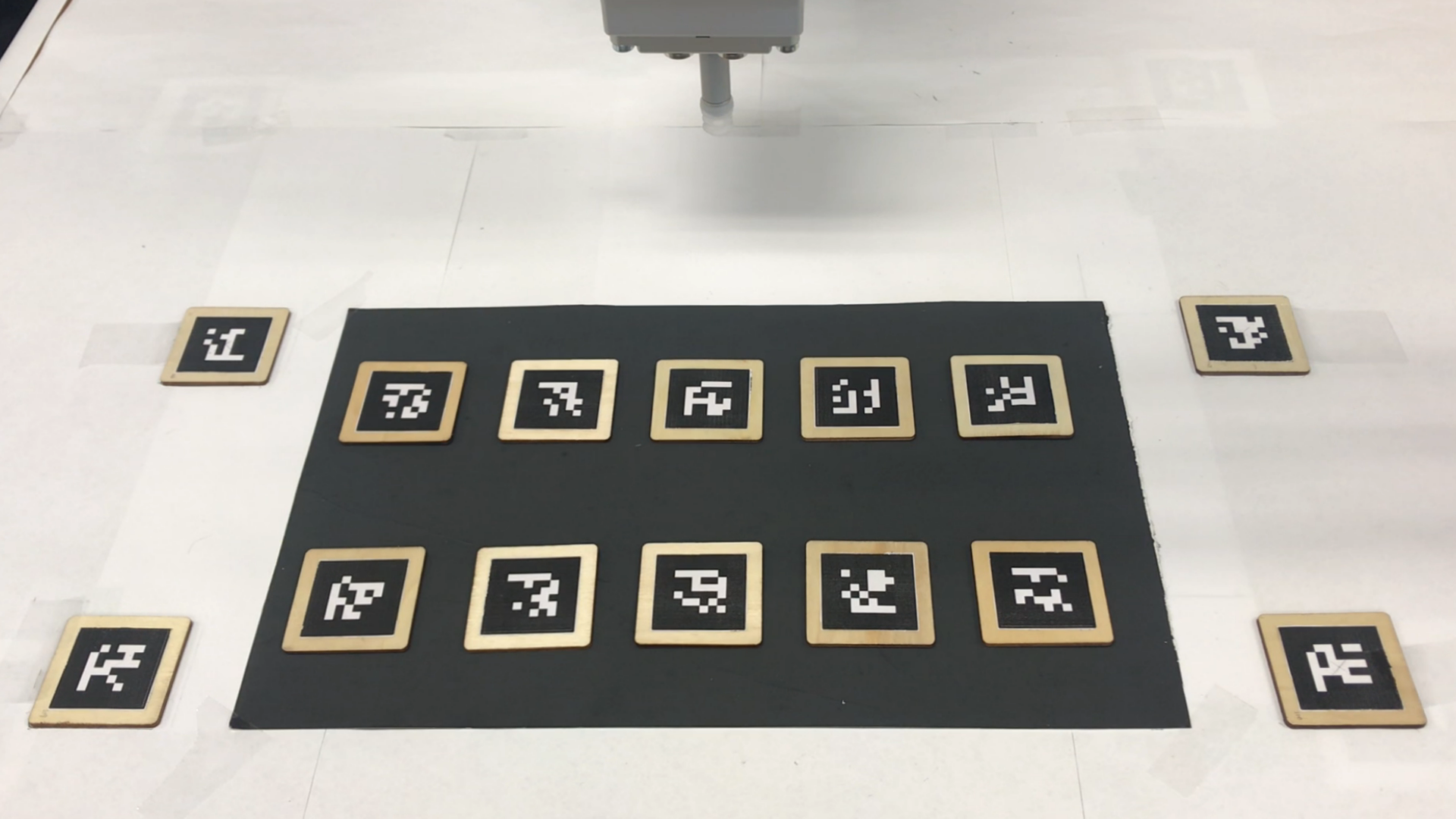}
    \includegraphics[width=0.3\textwidth]{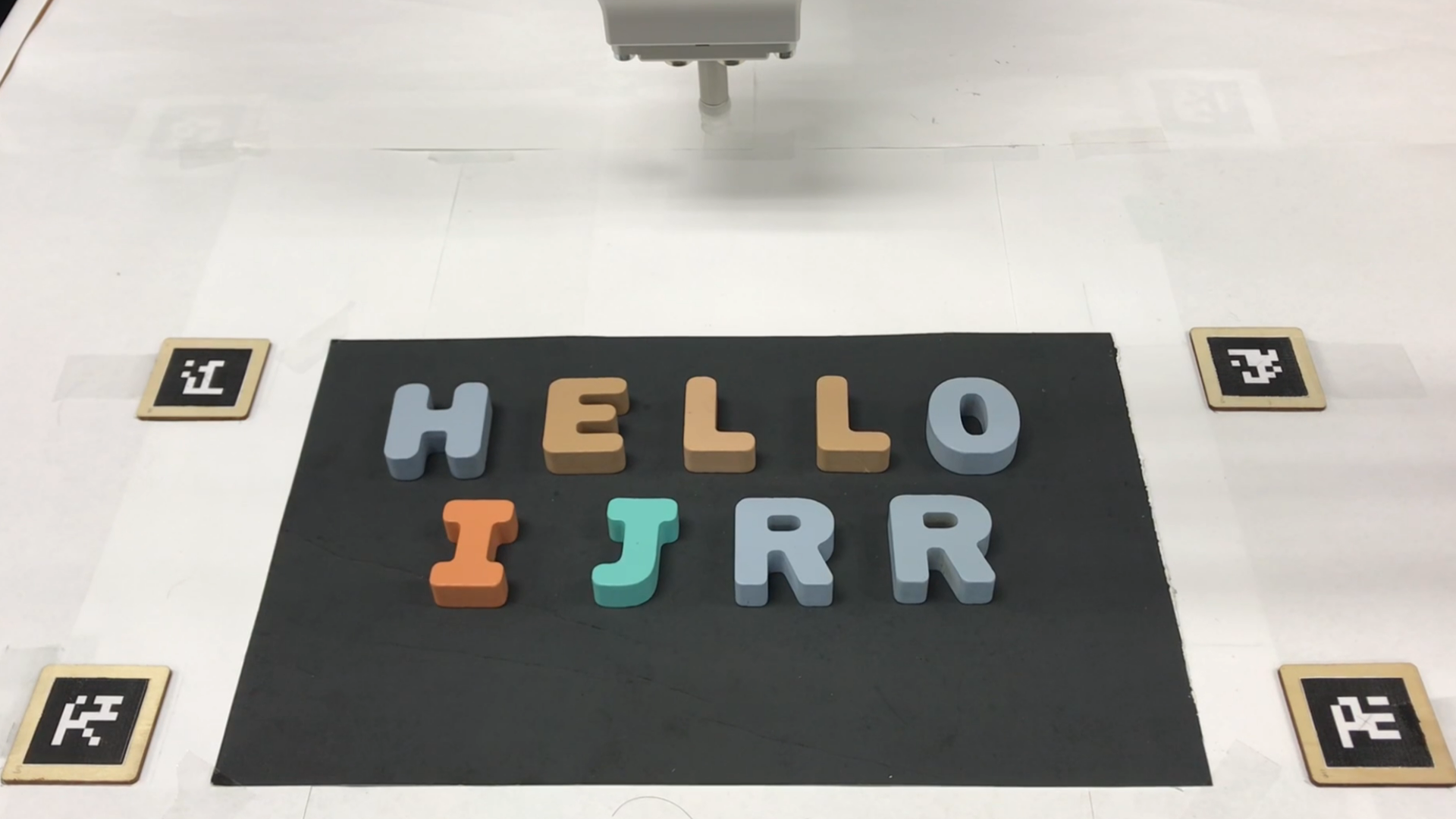}
    \caption{Three attempted scenarios in the hardware platform: cylindrical scenario (Left), cuboid scenario (Middle), letters (Right)}
    \label{fig:hardware_exp}
\end{figure}

In our hardware setup (Fig.\ref{fig:workspace}), we use a UR-5e robot arm with an OnRobot VGC 10 vacuum gripper to execute pick-n-places. 
An Intel RealSense D435 RGB-D camera is set up above the environment to provide an overview of the entire workspace.
The camera calibration is done with four 2D fiducial markers  at the corners of the environment using a C++ cross-platform software library Chilitags \cite{chilitags}.
In the real robot evaluation, we attempt three scenarios, which are presented in Fig.~\ref{fig:hardware_exp}. 
For both cylindrical objects and cuboid objects, 
pose estimation is conducted with the aid of Chilitags. 
For letter objects,
the poses are estimated with a deep learning model Mask R-CNN \cite{he2017mask}:
Given an overview image of the workspace, 
a pre-trained Mask R-CNN model provides masks of the workspace objects.
The 2d pose of each object is obtained by computing a rigid transformation between the detected 2d point cloud and the model point cloud of the object.

\begin{algorithm}[h]
\begin{small}
    \SetKwInOut{Input}{Input}
    \SetKwInOut{Output}{Output}
    \SetKwComment{Comment}{\% }{}
    \caption{Rearrangement Pipeline}
    \label{alg:ur5e}
    \SetAlgoLined
		\vspace{0.5mm}
    \Input{$\mathcal A_g$: goal arrangement}
		\vspace{0.5mm}
		$\mathcal A_c \leftarrow PoseEstimation()$\\
		$\pi \leftarrow$ RearrangementSolver($\mathcal A_c$, $\mathcal A_g$)\\
		\For{($o$, $p_c$, $p_t$) in $\pi$}
		{
		$p_c \leftarrow $ UpdatePose($o$)\\
		ExecuteAction(($o$, $p_c$, $p_t$))\\
		}
		
\end{small}
\end{algorithm}

The rearrangement pipeline is shown in Algo.~\ref{alg:ur5e}.
The system first estimates the current poses of workspace objects $\mathcal A_c$ (Line 1).
Given the current arrangement $
\mathcal A_c$ and goal arrangement $\mathcal A_g$, 
the rearrangement solver computes a rearrangement plan $\pi$ (Line 2).
Each action in $\pi$ consists of three components: manipulating object $o$, current pose $p_c$, and target pose $p_t$ (Line 3).
To improve the accuracy of pick-n-places, 
we update the grasping pose before each grasp (Line 4-5).

\subsection{Experimental Validation}\label{subsec:hardware_exp}
\begin{figure}
    \centering
    \includegraphics[width=0.8\columnwidth]{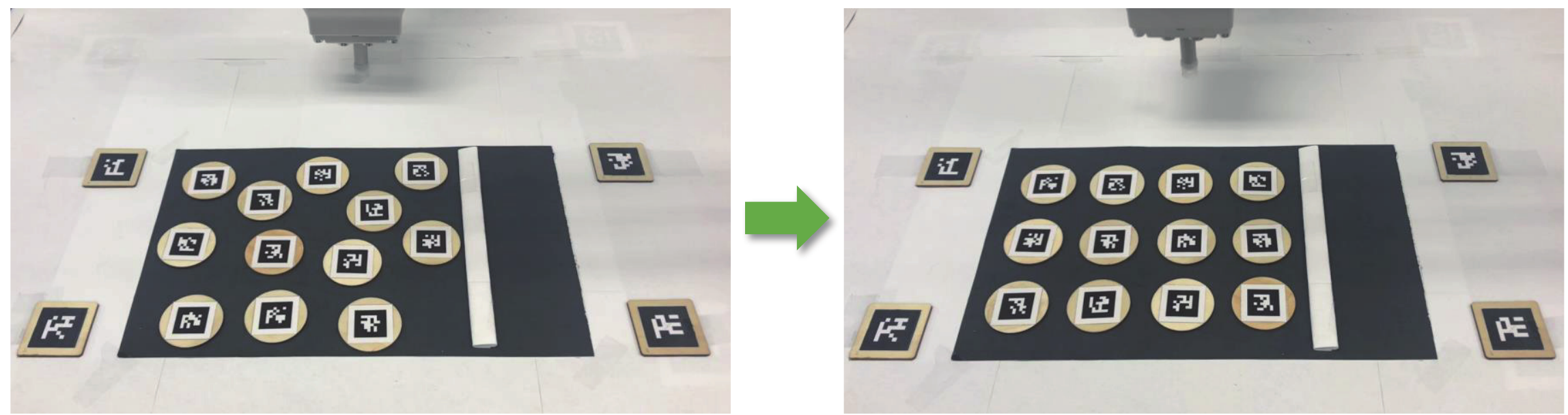}
    \caption{An example instance of our experiment.
The right side of the pad works as an external space for buffer placements.}
    \label{fig:tore-hardware-exp}
\end{figure}

We conduct hardware experiments on \toroe, 
comparing the execution time of RBM plans and TBM plans. With the same notations as those in Sec.~\ref{sec:experiments}, RBM plans minimize running buffer size and TBM plans minimize total buffer size.
In \toroe, minimizing the total buffer size is equivalent to minimizing the number of total actions.
An example instance of our experiment is shown in Fig.~\ref{fig:tore-hardware-exp}.
The right side of the pad works as an external space for buffer placements.
We tried 10 instances with 12 cylindrical objects.
The results are shown in Tab.~\ref{Tab:TB_RB}.
While minimizing running buffer size,
RBM plans are only 5$\%$ longer than TBM plans on average.

\begin{center}
\begin{table}[h!]
\caption{\label{Tab:TB_RB} Comparison between RBM plans and TBM plans in execution time and the number of actions.}
\centering
\resizebox{0.5\textwidth}{!}{
\begin{tabular}{|l|ll|ll|}
\hline
\multirow{2}{*}{instance} & \multicolumn{2}{l|}{RBM}                                                                    & \multicolumn{2}{l|}{TBM}                                                                   \\ \cline{2-5} 
                          & \multicolumn{1}{l|}{\begin{tabular}[c]{@{}l@{}}execution \\ time (secs)\end{tabular}} & \# actions & \multicolumn{1}{l|}{\begin{tabular}[c]{@{}l@{}}execution\\ time (secs)\end{tabular}} & \# actions \\ \hline
1 & \multicolumn{1}{l|}{211.35}&15& \multicolumn{1}{l|}{194.52}&14\\
2 & \multicolumn{1}{l|}{197.76}&14& \multicolumn{1}{l|}{197.66}&14\\
3 & \multicolumn{1}{l|}{184.49}&13& \multicolumn{1}{l|}{185.82}&13\\
4 & \multicolumn{1}{l|}{201.32}&14& \multicolumn{1}{l|}{205.96}&14\\
5 & \multicolumn{1}{l|}{227.03}&16& \multicolumn{1}{l|}{195.44}&14\\
6 & \multicolumn{1}{l|}{193.84}&14& \multicolumn{1}{l|}{192.61}&14\\
7 & \multicolumn{1}{l|}{216.49}&15& \multicolumn{1}{l|}{198.90}&14\\
8 & \multicolumn{1}{l|}{250.13}&17& \multicolumn{1}{l|}{211.15}&15\\
9 & \multicolumn{1}{l|}{176.75}&13& \multicolumn{1}{l|}{179.83}&13\\
10 & \multicolumn{1}{l|}{179.07}&12& \multicolumn{1}{l|}{160.96}&12\\ \hline
Average & \multicolumn{1}{l|}{203.82}&14.3& \multicolumn{1}{l|}{192.28}&13.7\\ \hline

\end{tabular}
}
\end{table}
\vspace{-3mm}
\end{center}

Besides the comparison in \toroe, 
we also compare \toroi algorithms in the hardware system.
As shown in the accompanying video, \trlb solves all attempted instances, which involve concave objects, in an apparently natural and efficient manner.

\section{Conclusions}\label{sec:conclusions}
In this work, we investigate the problem of minimizing the number of running 
buffers (\mrb) for solving labeled and unlabeled tabletop rearrangement problems 
with overhand grasps (\toro), which translates to finding a best linear ordering 
of vertices of the associated underlying dependency graph. 
For \toro, \mrb is an important quantity to understand as it determines the 
problem's feasibility if only external buffers are to be used, which is 
the case in some real-world applications \cite{han2018complexity}.
Despite the provably high computational complexity that is involved, we provide 
effective dynamic programming-based algorithms capable of quickly computing \mrb 
for large and dense labeled/unlabeled \toro instances.
In addition, we also provide methods for minimizing the total number of buffers 
subject to \mrb constraints.
Whereas we prove that \mrb can grow unbounded for both labeled and unlabeled 
settings for special cases for uniform cylinders, empirical evaluations 
suggest that real-world random \toro instances are likely to have much smaller
\mrb values. 
We demonstrate \mrb's high utility in solving real tabletop rearrangement 
problems in a bounded, constrained workspace. 

We conclude by leaving the readers with some interesting open problems. On the 
structural side, while \lrbm, in general, is proven to be NP-Hard, the computational 
the intractability of either \lrbm with uniform cylinders or \urbm in general remains
unresolved. 
As for bounds, the lower and upper bounds of \mrb for \lrbm for uniform cylinders
do not yet agree; can the bound gap be narrowed further? 
Objects may have different sizes. An interesting question here is if we can move a large object to buffer or a small object to buffer, which is more beneficial? Moving larger objects are more challenging but we may need to do this fewer times. 

%Open problems: 
%Objects can be partially labeled (or colored) - seems more complex (we can also study this!) 

\section*{Acknowledgments}\label{sec:ack}
This work is supported in part by NSF awards IIS-1845888, CCF-1934924 and
IIS-2132972. We sincerely thank the anonymous reviewers for bringing 
up many insightful suggestions and intriguing questions, which have helped 
improve the quality and depth of the study. 

\bibliographystyle{ieeetr}
\bibliography{bib}

\section*{Appendix}\label{sec:app}
\subsection*{Properties of Unlabeled Dependency Graphs for Special \toro Settings}
\begin{figure}[ht!]
    \centering
    \includegraphics[width=0.49\textwidth]{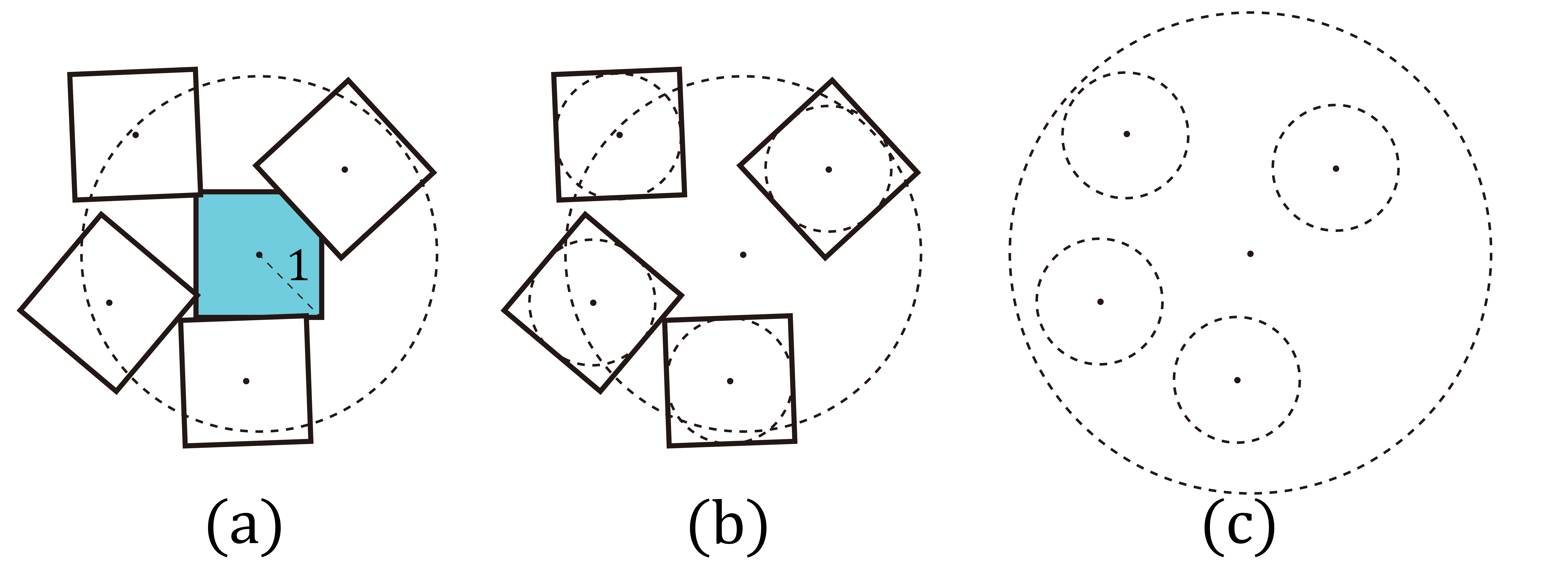}
    \caption{(a) The number of disjoint d-sided regular polygons that overlap with one d-sided regular polygon is upper bounded by that whose centers are inside a 2-circle. (b) The number of disjoint d-sided regular polygons whose centers are inside a 2-circle is upper bounded by the number of disjoint $\cos(\dfrac{\pi}{d})$-circles whose centers are inside a 2-circle. (c) The number of disjoint $\cos(\dfrac{\pi}{d})$-circles whose centers are inside a 2-circle is equal to the number of $\cos(\dfrac{\pi}{d})$-circles packed inside a $(2+\cos(\dfrac{\pi}{d}))$-circle.}
    \label{fig:prop_2}
\end{figure}

\begin{proof}[Proof of Proposition~\ref{p:udg-polygon}]
 The maximum degree of the dependency graph is equal to the maximum number of disjoint objects that one object can overlap with. 
In this proof, we evaluate the upper bound of the maximum overlaps with a disc packing problem.
We first discuss the range of distance between two overlapping regular polygons when they are overlapping and disjoint. After that, we relate the problem to a disc packing problem and derive the upper bound. Without loss of generality, 
we assume the radius of the regular polygons to be 1.
When two $d$-sided regular polygons overlap, the distance between the polygon centers is upper bounded by two times the radius of their circumscribed circles, i.e. 2.
When two $d$-sided regular polygons are disjoint, the distance between the polygon centers is lower bounded by two times the radius of their inscribed circles, i.e.  $2\cos(\dfrac{\pi}{d})$.
 Due to the mentioned upper bound, the number of disjoint d-sided regular polygons that overlap with one d-sided regular polygon is upper bounded by that whose centers are inside a 2-circle (Fig.~\ref{fig:prop_2}(a)). Additionally, due to the mentioned distance lower bound, this number is upper bounded by the number of disjoint $\cos(\dfrac{\pi}{d})$-circles whose centers are inside a 2-circle (Fig.~\ref{fig:prop_2}(b)), which is equal to the number of $\cos(\dfrac{\pi}{d})$-circles packed inside a $(2+\cos(\dfrac{\pi}{d}))$-circle (Fig.~\ref{fig:prop_2}(c)).
Since the radius ratio of circles $\dfrac{2+\cos(\dfrac{\pi}{d})}{\cos(\dfrac{\pi}{d})}$ is upper bounded by 5 when $d\geq 3$, there are at most 19 small circles packed in the large circle \cite{fodor1999densest}.
Therefore, the maximum degree of the unlabeled dependency graph is upper bounded by 19.
% \qed
\end{proof}

\begin{figure}[ht!]
    \centering
    \includegraphics[width=0.4\textwidth]{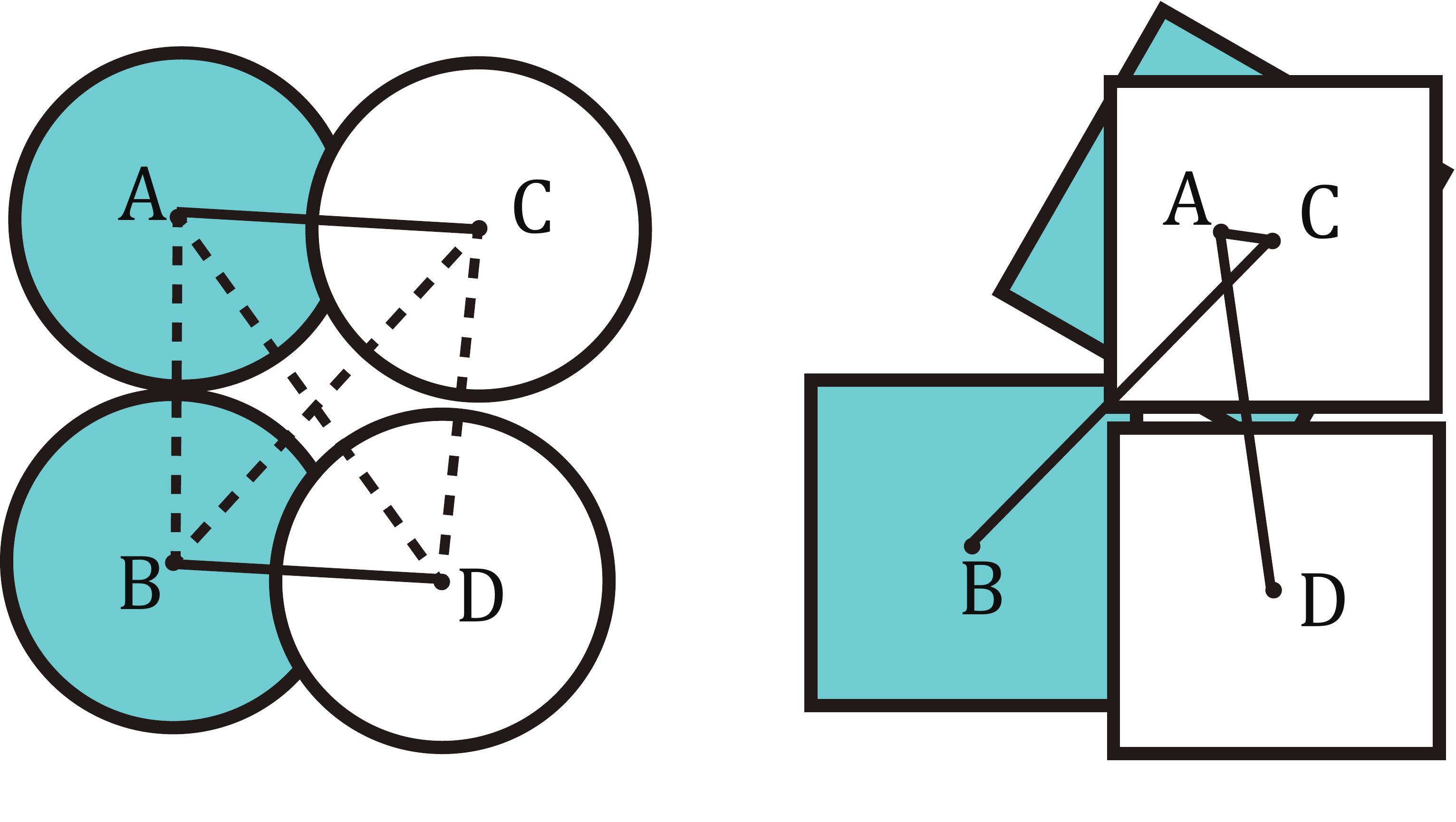}
    \caption{[Left] A case discussed in Prop.\ref{p:udg} [Right] A counterexample of the proof of Prop.\ref{p:udg} when the base of workspace objects is square.}
    \label{fig:planarity}
\end{figure}

\begin{proof}[Proof of Proposition~\ref{p:udg}]
Constructing the dependency graph based on the start and goal positions in the workspace, the planarity can be proven if no two edges cross each other.
Assuming the contrary, there are two edges crossing each other, e.g. $AD$ and $BC$ in Fig.~\ref{fig:planarity}[Left].
Since each edge comes from a pair of intersecting start and goal objects, without loss of generality,
we can assume $C$, $D$ are start positions and $A$, $B$ are goal positions.
Therefore, we have $|AB|,|CD|\geq 2r$,
where $r$ is the base radius of the cylinders.
Since $AD$ and $BC$ are both connected in the dependency graph, 
we have $|AD|,|BC| < 2r$, which leads to the conclusion that $|AD|+|BC|< 4r \leq |AB|+|CD|$, and contradicts the triangle inequality.
Therefore, the planarity is proven with contradiction.

Since uniform cylinders have uniform disc bases, one cylinder may only 
touch six non-overlapping cylinders and non-trivially intersect at most five 
non-overlapping cylinders. 
Therefore, the maximum degree of the unlabeled dependency graph is upper bounded by 5.
% \qed
\end{proof}

\begin{remark}
As shown in Fig.~\ref{fig:planarity}[Right],
the proof of the planarity in Proposition~\ref{p:udg} does not hold for objects with regular polygon bases.
\end{remark}

\subsection*{$\Omega(\sqrt{n})$ Lower Bound for \urbm}
Let $(x, y)$ 
be the coordinate of a vertex $v_{x,y}$ on $\D(w, h)$ with the top left being $(1,
1)$. The parity of $x + y$ determines the partite set of the vertex (recall that 
unlabeled dependency graph for uniform cylinders is always a planar bipartite 
graph, by Proposition~\ref{p:udg}), which may correspond to a start pose or 
a goal pose. With this in mind, we simply call vertices of $\D(w, h)$ start and 
goal vertices; without loss of generality, let $v_{1,1}$ be a start vertex. 
%
%Assuming the cylinders' footprints have unit radius, the distance between adjacent 
%vertices on $\D(w, h)$ is $\sqrt{2}$.
%

%For the unlabeled setting, we show that the \mrb of the worst cases is $\Omega(\sqrt{n})$ where $n$ is the number of objects in the instance. This property is proved with a instance whose dependency graph is a \emph{dependency grid} whose definition is as follows.

%\begin{definition}[Dependency Grid]
%A $W\times H$ dependency grid $\mathcal{D}$ is a square grid graph whose vertices correspond to points in the plane with integer coordinates, $x$-coordinates being in the range 1, ..., $W$, $y$-coordinates being in the range 1, ..., $H$ (Fig. \ref{fig:DependencyGrid}). Each vertex of $\mathcal{D}$ is colored blue or black but the neighboring vertices cannot be with the same color. Denote the vertex at the $i$th column and $j$th row in $\mathcal{D}$ as $v_{i,j}$. Without loss of generality, we can assume that $v_{1,1}$ is black.
%\end{definition}

%When the poses in $\mathcal{A}_1$ and $\mathcal{A}_2$ are placed at the coordinates of black and blue vertices in $\mathcal{D}$ respectively and the unit length of the grid is $\sqrt{2}$ units, $\mathcal{D}$ is the dependency graph of the rearrangement instance.

We use $\D(m, 2m)$ for establishing the lower bound on \mrb. We use a \emph{vertex 
pair} $p_{i, j}$ to refer to two adjacent vertices $v_{i, 2j-1}$ and $v_{i, 2j}$ in 
$\D(m, 2m)$. It is clear that a vertex pair contains a start and a goal vertex. 
We say that a goal vertex is \emph{filled} if an object is placed at the corresponding goal pose. 
We say that a start vertex (which belongs to a vertex pair) is \emph{cleared} if the 
corresponding object at the vertex is picked (either put at a goal or at a buffer)
but the corresponding goal in the vertex pair is unfilled. 
At any moment when the robot is not holding an object, the number of objects in the 
buffer is the same as the number of cleared vertices. For each column $i, 1\leq i 
\leq m$, let $f_i$ (resp., $c_i$) be the number of goals (resp., start) vertices in 
the column that are filled (resp., cleared). Notice that a goal cannot be filled 
until the object at the corresponding start vertex is removed.

%In the rest of the subsection, we prove that for a rearrangement instance whose dependency graph is a $\sqrt{n}\times (2\sqrt{n})$ dependency grid $\mathcal{D}$, the minimum \mrb is $\Omega (\sqrt{n})$. The general idea is to evaluate the minimum needed buffer space when $\lfloor n/3\rfloor$ objects are at the goal poses.
%
%Without loss of generality, we can assume that $n$ is a square number. Note that in each column, there are $m$ start vertices and goal vertices respectively. A \emph{vertex pair} $P(i,j)$ is defined to be a pair of adjacent vertices in the same column ($v_{i,2j-1}$, $v_{i,2j}) (1\leq i,j \leq m)$. One of them is a start vertex and the other one is a goal vertex. A goal vertex is \emph{filled\ up} if there is an object at the corresponding pose. A \emph{cleared vertex} is a start vertex whose corresponding object has been moved away but the goal vertex in the same vertex pair is not filled up. Therefore, when $\lfloor n/3\rfloor$ goal vertices are filled up, the number of objects in the buffer is the same as the number of cleared vertices at the moment. For each column $i(1\leq i \leq m)$, denote the number of filled-up goal vertices and the number of cleared vertices as $g_i$ and $p_i$ respectively.

\begin{lemma}\label{lemma:adjacent}
On a dependency grid $\D(m, 2m)$, for two adjacent columns $i$ and $i+1$, $1 \le i < m$, if
$f_i+f_{i+1}\neq 0$ or $2m$, then $c_i+c_{i+1}\geq 1$. In other words, there is at least one 
cleared vertex in the two adjacent columns unless $f_i=f_{i+1}=0$ or $f_i=f_{i+1}=m$.
\end{lemma}
%\begin{proof}See Sec.~\ref{sec:proofs}.\end{proof}
\begin{proof}
If there is a $j, 1\leq j \leq m$, such that only one of the goal vertices in vertex pairs 
$p_{i,j}$ and $p_{i+1, j}$ is filled (Fig. \ref{fig:LemmaProof}(a)), then the start 
vertex in the other vertex pair must be cleared. Therefore, $c_i+c_{i+1}\geq 1$.

On the other hand, if, for each $j, 1\leq j \leq m$, both or neither of the goal vertices 
in $p_{i,j}$ and $p_{i+1,j}$ is filled, then there is a $j, 1 \leq j \leq m-1$, such that both 
goal vertices in $p_{i,j}$ and $p_{i+1,j}$ are filled but neither of those in $p_{i,j+1}$ and $p_{i+1,j+1}$ is filled (Fig.~\ref{fig:LemmaProof}(b)) or the opposite 
(Fig.~\ref{fig:LemmaProof}(c)). Then, for the vertex pairs whose goal vertices are not filled, 
say $p_{i,j+1}$ and $p_{i+1, j+1}$, one of their start vertices is a neighbor of the filled 
goal in $p_{i,j}$ and $p_{i,j+1}$. Therefore, at least one of the start vertices in $p_{i, j+1}$ 
and $p_{i+1,j+1}$ is a cleared vertex. And thus, $c_i+c_{i+1}\geq 1$.

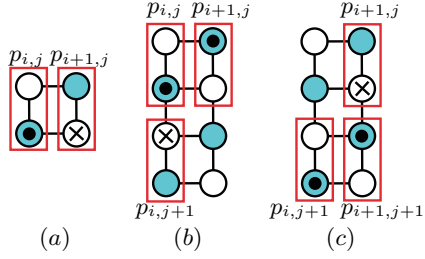
\begin{figure}[h!]
    \centering
    \vspace{2mm}
    \begin{overpic}
    [width = 0.30\textwidth]{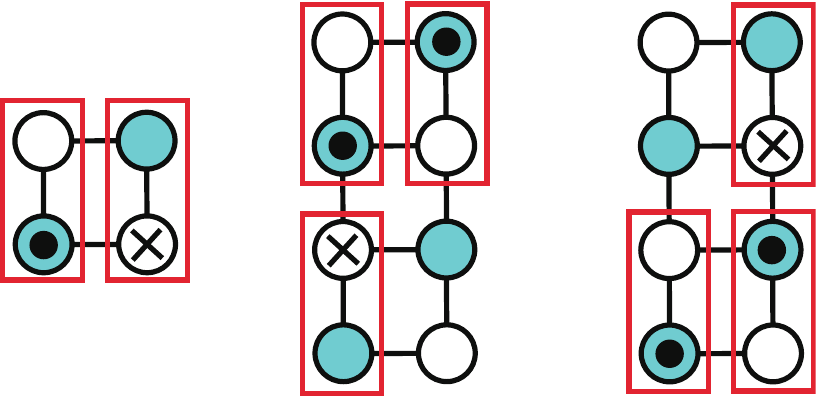}
    \put(0, 38){{\small $p_{i,j}$}}
    \put(11, 38){{\small $p_{i+1,j}$}}
    
    \put(37, 51){{\small $p_{i,j}$}}
    \put(49, 51){{\small $p_{i+1,j}$}}
    \put(34, -3){{\small $p_{i,j+1}$}}
    
    \put(89, 51){{\small $p_{i+1,j}$}}
    \put(70, -3){{\small $p_{i,j+1}$}}
    \put(89, -3){{\small $p_{i+1,j+1}$}}
    
    \put(8, -11.5){{\small $(a)$}}
    \put(44, -11.5){{\small $(b)$}}
    \put(85, -11.5){{\small $(c)$}}
    \end{overpic}
    \vspace{6mm}
    \caption{Some cases discussed in the lemma~\ref{lemma:adjacent}. The unshaded and shaded nodes represent the start and goal vertices in the dependency graph. Specifically, the shaded nodes with a dot inside represent the filled vertices and the unshaded nodes with a cross inside represent the cleared vertices. (a) When only one goal vertex in $p_{i,j}$ and $p_{i+1, j}$ is filled up, the start vertex in the other vertex pair is a cleared vertex. (b) When both goal vertices in $p_{i,j}$ and $p_{i+1,j}$ are filled but neither of those in $p_{i,j+1}$ and $p_{i+1,j+1}$ is filled, one of the start vertices $p_{i,j+1}$ and $p_{i+1,j+1}$ is a cleared vertex. (c) The opposite case of (b).}
    \label{fig:LemmaProof}
\end{figure}
% \qed
\end{proof}

%\begin{proof}See Sec.~\ref{sec:proofs}.\end{proof}
\begin{proof}[Proof of Lemma~\ref{l:urbm-lower}]
We show that there are $\Omega(m)$ cleared vertices when $\lfloor n/3 \rfloor$ goal vertices 
are filled. Suppose there are $q$ columns in $\mathcal{D}$ with $1\leq f_i\leq m-1$. According 
to the definition of $f_i$, for each of these $q$ columns, there is at least one goal vertex 
that is filled and at least one goal vertex that is not.

If $q< \dfrac{\lfloor n/3 \rfloor}{3(m-1)}$, then there are two columns $i$ and $j$, such that $f_i=m$ and $f_j=0$. That is 
because $\sum_{1\leq i \leq m} f_i= \lfloor n/3 \rfloor$ and $0\leq f_i \leq m$ for all $1\leq i \leq m$. 
Therefore, for the vertex pairs in each row $j$, at least one goal vertex is filled 
but at least one is not. And thus, for each $j$, there are two adjacent columns $i, i+1, 1\leq 
i < m$, such that \b{in vertex pairs $p_{i,j}$ and $p_{i+1,j}$, one goal vertex is filled while the
start vertex in the other vertex pair is cleared}
(Fig. \ref{fig:LemmaProof}(a)). 
Therefore, there are at least $m$ cleared vertices in this case.

If $q\geq \dfrac{\lfloor n/3 \rfloor}{3(m-1)}$, then we partition all the columns in $\mathcal{D}$ into $\lfloor m/2\rfloor$ disjoint pairs: 
(1,2), (3,4), ... The $q$ columns belong to at least $\lfloor q/2\rfloor$ pairs of 
adjacent columns. Therefore, according to Lemma \ref{lemma:adjacent}, we have $\Theta(m)$ 
cleared vertices.

In conclusion, there are $\Omega(m)$ cleared vertices when there are $\lfloor n/3 \rfloor$ 
filled goal. Therefore, the minimum \mrb of this instance is $\Omega(m)$.
% \qed
\end{proof}

\subsection*{$O(\sqrt{n})$ Algorithmic Upper Bound for \urbm}
To investigate the running buffer size of the plan computed by \spp,
we construct a binary tree $T$ based on \spp(Fig.~\ref{fig:separator}(c)).
Each node represents a recursive call consuming a subgraph $\udg(V,E)$ and the left and right children of the node are induced from subgraphs $\udg(A',E(A'))$ and $\udg(B',E(B'))$ of $\udg(V,E)$.
\spp computes the plan by visiting the binary tree in a depth-first manner.
For each node on $T$, given the input dependency graph $\udg(V,E)$, 
denote the sequence before we deal with $\udg(V,E)$ as $\pi_0$. 
Denote the vertices pruned in RemovalTrivialGoals( $\udg(V,E)$) as $P$.
Without loss of generality,
assume that \spp recurses into $A'$ before $B'$.
Let $\pi_{P}$, $\pi_{C'}$, $\pi_{A'}$, and $\pi_{B'}$ be the goal removal sequence after we remove vertices in $P$, $C'$, $A'$ and $B'$ from $\udg(V,E)$ respectively.
We define function $RB(\pi)$ to represent ``generalized'' current running buffer size of a removal sequence $\pi$:
When vertices in $\pi$ are removed from the dependency graph, 
$RB(\pi)$ equals either the number of running buffers; 
or the negation of the number of empty goal poses.
In the depth-first recursion, each node on the binary tree $T$ is visited at most three times:

\begin{enumerate}
    \item Before exploring child nodes. The peak may be reached during the removal of $C'$. Let $\pi_{C^*}$ be the sequence when the running buffer size reaches the peak. 
    $RB(\pi_{C^*}) \leq RB(\pi_0) + 10\sqrt{2} \sqrt{|V|}$.
    \item After we deal with $A'$ and before we deal with $B'$.
    $RB(\pi_{A'}) = RB(\pi_{0})$$-\delta(C')-\delta(A')$. 
    \item After we deal with $B'$.
    $RB(\pi_{B'}) = RB(\pi_0)$$-\delta(V)$.
\end{enumerate}

\begin{lemma}\label{lemma:average}
$RB(\pi_{A'})\leq (RB(\pi_{C^*})+RB(\pi_{B'}))/2$.
\end{lemma}

\begin{proof}
\begin{equation*}
    \begin{split}
        & RB(\pi_{A'}) \\
        & = RB(\pi_{0})-\delta{P}-\delta{C'}-\delta{A'}\\
        %&\ \ \ \ \hcancel[black]{+ (s(P)-g(P)) + (s(C')-g(C')) + (s(A')-g(A'))}\\ 
%     \end{split}
% \end{equation*}
% \begin{equation*}
%     \begin{split}
        &= RB(\pi_0)  -\delta(P)-\delta(C')-\max[\delta(A'),\delta(B')]\\
        %&\ \ \ \ \hcancel[black]{+ (s(P)-g(P))+ (s(C')-g(C'))}\\
        %&\ \ \ \ \hcancel[black]{+ \min[s(A')-g(A'), s(B')-g(B')]}\\
%     \end{split}
% \end{equation*}
% \begin{equation*}
%     \begin{split}
        &\leq RB(\pi_0) -\delta(P)-\delta(C')+\dfrac{1}{2}[-\delta(V)+\delta(P)+\delta(C')]\\
       % &\ \ \ \  \hcancel[black]{+ (s(P)-g(P)) + (s(C')-g(C'))}\\ 
       % &\ \ \ \  \hcancel[black]{+\dfrac{1}{2}[(s(V)-g(V)) - (s(P)-g(P))}\\
        %&\ \ \ \ \hcancel[black]{- (s(C')-g(C'))]}\\
%     \end{split}
% \end{equation*}
% \begin{equation*}
%     \begin{split}
        &= \dfrac{1}{2} \{[RB(\pi_0)-\delta(V)]+[RB(\pi_0)-\delta(P)-\delta(C')]\}\\ 
        %&\ \ \ \ \hcancel[black]{+ s(V)-g(V)]}\\
        %&\ \ \ \ \hcancel[black]{+[RB(\pi_0) + (s(P)-g(P)) + (s(C')-g(C'))]\}}\\
%     \end{split}
% \end{equation*}
% \begin{equation*}
%     \begin{split}
        &= \dfrac{1}{2} [RB(\pi_{B'})+RB(\pi_{C'})]\\
        &\leq \dfrac{1}{2} [RB(\pi_{B'})+RB(\pi_{C^*})]
    \end{split}
\end{equation*}
% \qed
\end{proof}

Lemma~\ref{lemma:average} establishes the relationship among $\pi_{C^*}$, $\pi_{A'}$, and $\pi_{B'}$ of a node on $T$.
With this lemma, we obtain an upper bound for each node:

% \begin{proposition}
% A positive \mrb of the plan must be $RB(\pi_{C^*})$ of a node on $T$. 
% \end{proposition}
%\jy{This proposition statement is confusing to me.}

\begin{lemma}\label{lemma:induction}
Given a node $N$ with depth $d$ in the binary tree $T$, let $\pi_{C^*}(N)$, $\pi_{A'}(N)$, and $\pi_{B'}(N)$ be $\pi_{C^*}$, $\pi_{A'}$, and $\pi_{B'}$ of $N$ respectively. 
$RB(\pi_{C^*}(N))$, $RB(\pi_{A'}(N))$, and $RB(\pi_{B'}(N))$ are all upper bounded by 
\begin{equation}
    \dfrac{[1-(\sqrt{\dfrac{2}{3}})^{d+1}]}{1-\sqrt{\dfrac{2}{3}}}20\sqrt{n}
\end{equation}
where $n$ is the number of objects in the instance. 
\end{lemma}

\begin{proof}
The conclusion can be proven by induction.\\

When $d=0$, the dependency graph at the root node $r$ has $n$ start vertices and $n$ goal vertices. 
We have $RB(\pi_{C^*})\leq 20\sqrt{n}, RB(\pi_{B'})=0$. 
According to Lemma \ref{lemma:average}, 
$RB(\pi_{A'})\leq 10 \sqrt{n}$.
The conclusion holds.

Assume that the conclusion holds for all the nodes with depth less than or equal to $k$.
Given an arbitrary node $N$ in the depth $k$, 
let the left and right children of $N$ be $L$ and $R$, which are the nodes with depth $k+1$. 
The corresponding dependency graphs have at most $(2/3)^{k+1}\cdot 2n$ vertices respectively. 
$$RB(\pi_{C^*}(L))\leq RB(\pi_{C^*}(N)) + \sqrt{(2/3)^{k+1}\cdot 2n}\cdot 10\sqrt{2}$$
$$RB(\pi_{C^*}(R))\leq RB(\pi_{A'}(N)) + \sqrt{(2/3)^{k+1}\cdot 2n}\cdot 10\sqrt{2}$$
Since $\pi_{B'}(L)=\pi_{A'}(N)$ and $\pi_{B'}(R)=\pi_{B'}(N)$, upper bound for nodes with depth $k$ holds for $\pi_{B'}(L)$ and $\pi_{B'}(R)$. Therefore, the running buffer size for the depth $k+1$ nodes has an upper bound
\begin{equation}
    \begin{split}
        & \dfrac{[1-(\sqrt{2/3})^{k+1}]}{1-\sqrt{2/3}}20\sqrt{n} + \sqrt{(2/3)^{k+1}\cdot 2n}\cdot 10\sqrt{2}\\ 
        = &\dfrac{[1-(\sqrt{2/3})^{k+2}]}{1-\sqrt{2/3}}20\sqrt{n}        
    \end{split}
\end{equation}

Therefore, the upper bound holds for all the nodes of depth $k+1$. 
With induction, the lemma holds.
% \qed
\end{proof}

With Lemma~\ref{lemma:induction}, it is straightforward to establish that \mrb is bounded by $\dfrac{20}{1-\sqrt{2/3}}\sqrt{n}$, yielding Theorem~\ref{t:urbm-upper}.

\end{document}